\documentclass[11pt,a4paper]{article}
\usepackage[margin=2.5cm]{geometry}
\usepackage[T1]{fontenc}
\usepackage{lmodern}
\usepackage{microtype}
\usepackage{graphicx}
\usepackage[font=small,labelfont=bf]{caption}
\usepackage[round,authoryear]{natbib}

\usepackage{multirow}
\usepackage{amsmath,amssymb}
\usepackage{amsthm}
\usepackage{xcolor}
\usepackage{booktabs}
\usepackage{array}
\usepackage{placeins}
\usepackage{adjustbox}
\usepackage{xurl}

\usepackage{amsmath,amsfonts,bm}

\def\eqref#1{equation~\ref{#1}}

\def\1{\bm{1}}

\DeclareMathAlphabet{\mathsfit}{\encodingdefault}{\sfdefault}{m}{sl}
\SetMathAlphabet{\mathsfit}{bold}{\encodingdefault}{\sfdefault}{bx}{n}

\newcommand{\probeLineMin}{4.3}
\newcommand{\probeLineMax}{11.0}
\newcommand{\probeFullMax}{42.6}
\newcommand{\gapQwenMax}{2.0}
\newcommand{\gapAqua}{4.2}
\newcommand{\thinkGsmRQ}{189}
\newcommand{\gapGsmRQ}{6.0}
\newcommand{\thinkGsmRL}{153}
\newcommand{\gapGsmRL}{13.0}

\newcommand{\svGsmQttLS}{50.6}

\newcommand{\svGsmQttFix}{33.6}

\newcommand{\dGsmQttabs}{17.0}

\newcommand{\dAqua}{+10.4}

\newcommand{\dMmluQttabs}{14.3}

\newcommand{\dMathRLabs}{6.2}

\newcommand{\dMathRQabs}{4.1}

\newcommand{\svGpqaQeLS}{11.0}

\newcommand{\svGpqaQeFix}{4.0}

\newcommand{\dGpqaQe}{+7.0}

\newcommand{\svGpqaQttLS}{19.2}

\newcommand{\svGpqaQttFix}{4.7}

\newcommand{\dGpqaQtt}{+14.6}

\newcommand{\dQmathAbsMin}{10.4}
\newcommand{\dQmathAbsMax}{17.0}
\newcommand{\dMmluAbsMin}{4.4}
\newcommand{\dMmluAbsMax}{14.3}
\newcommand{\mmluClearList}{Qwen3-32B and R1-Llama-8B}

\newcommand{\overheadMin}{0.8}
\newcommand{\overheadMax}{7.2}

\newcommand{\dropQmathMin}{2.4}
\newcommand{\dropQmathMax}{2.9}

\newcommand{\nBSbelowFix}{13}
\newcommand{\nSettings}{15}
\newcommand{\nConvBelowFix}{14}

\newcommand{\violMax}{10}
\newcommand{\violMean}{3.6}

\newcommand{\capGsmQeLS}{23.8}
\newcommand{\capGsmQeFix}{33.1}
\newcommand{\lineGsmQeLS}{45.7}
\newcommand{\lineGsmQeFix}{34.1}
\newcommand{\atGsmsave}{36.4}

\newcommand{\atMathsave}{56.3}

\newcommand{\atBestText}{\LS{} on GSM8K (19.6\%, accuracy 0.860) and the static budget{} on MATH-500 (24.9\%, accuracy 0.868)}
\newcommand{\natQmathDmin}{+12.3}
\newcommand{\natQmathDmax}{+17.6}
\newcommand{\poolWithin}{10}
\newcommand{\lotoWithin}{7}
\newcommand{\poolLossText}{GPQA-D with Qwen3-32B, GSM8K with Qwen3-32B and MATH-500 with Qwen3-32B ($-$16.3 to $-$4.5 points)}
\newcommand{\poolGainText}{MMLU-Pro with Qwen3-32B and GSM8K with R1-Qwen-7B (+8.9 and +4.3)}

\newcommand{\nCatStatic}{three of the four}
\newcommand{\settleGpqaMin}{5,835}
\newcommand{\settleGpqaMax}{7,524}
\newcommand{\settleMmluQMin}{2,048}
\newcommand{\settleMmluQMax}{3,072}

\newcommand{\looWinners}{static 8, LearnStop 7}
\newcommand{\bbhNat}{0.737}
\newcommand{\bbhEnd}{0.660}
\newcommand{\bbhCapHit}{24}
\newcommand{\nAlphaLS}{4}
\newcommand{\nAlphaFix}{3}

\newcommand{\alphaLSList}{GSM8K with Qwen3-8B, GSM8K with Qwen3-32B, MATH-500 with Qwen3-8B and MATH-500 with Qwen3-32B}
\newcommand{\alphaFixList}{MATH-500 with R1-Llama-8B, MMLU-Pro with R1-Qwen-7B and MMLU-Pro with R1-Llama-8B}
\newcommand{\flipToStaticList}{AQuA with Qwen3-8B, GPQA-D with Qwen3-8B and GPQA-D with Qwen3-32B}
\newcommand{\flipToLSList}{GSM8K with R1-Qwen-7B, GSM8K with R1-Llama-8B and MMLU-Pro with Qwen3-32B}

\newcommand{\flipGainFix}{51.1}
\newcommand{\flipGainLS}{36.6}
\newcommand{\flipLSGainFix}{12.8}
\newcommand{\flipLSGainLS}{29.3}

\newcommand{\alphaLSMinTwenty}{4.4}
\newcommand{\nSwing}{6}
\newcommand{\swingMin}{13.1}
\newcommand{\swingMax}{19.1}
\newcommand{\nSmallFlip}{2}
\newcommand{\smallFlipMax}{7.7}
\newcommand{\nSignChange}{8}
\newcommand{\nSignEnds}{6}
\newcommand{\swingExText}{from $-$8.2 to +10.9 points on GSM8K with R1-Qwen-7B, from +14.6 to $-$3.6 on GPQA-D with Qwen3-32B and from $-$14.3 to +2.0 on MMLU-Pro with Qwen3-32B}

\newcommand{\nRefFlips}{2}

\newcommand{\slGapMean}{2.3}

\newcommand{\uniFixA}{20.5}
\newcommand{\uniLSA}{22.7}
\newcommand{\uniSLB}{51.2}
\newcommand{\uniFixB}{48.3}
\newcommand{\uniLSB}{50.6}
\newcommand{\uniGapPC}{0.2}
\newcommand{\uniGapBB}{1.3}
\newcommand{\rMaxFix}{3.4}
\newcommand{\rMaxLS}{3.6}
\newcommand{\gpqaNcal}{99}

\newcommand{\gpqaLossText}{a candidate certified at $\delta/2$ may have no calibration loss, against one calibration loss at $\delta$}

\newcommand{\svSLGpqaQtt}{7.6}
\newcommand{\svLSGpqaQtt}{19.2}

\newcommand{\slGapWorst}{GPQA-D with Qwen3-32B}

\newcommand{\capGapMean}{5.5}
\newcommand{\capGapMax}{11.1}

\newcommand{\decCoverage}{6.8}
\newcommand{\decPower}{3.8}

\newcommand{\zeroTol}{0.5}

\newcommand{\nBEselPC}{8}
\newcommand{\nBEselBB}{2}

\newcommand{\selVsFourteenMin}{\ensuremath{-}2.6}
\newcommand{\selVsFourteenMax}{\ensuremath{-}1.3}

\newcommand{\beQmathLo}{0.026}
\newcommand{\beQmathHi}{0.079}

\newcommand{\beGsmQttBest}{0.34}

\newcommand{\beGsmQeTwo}{0.77}

\newcommand{\bootHalfMin}{0.6}
\newcommand{\bootHalfMax}{8.2}
\newcommand{\bootSdMin}{0.2}
\newcommand{\bootSdMax}{8.7}

\newcommand{\dRealQmathMin}{8.2}
\newcommand{\dRealQmathMax}{17.0}

\newcommand{\alphaGridMin}{0.01}
\newcommand{\alphaGridMax}{0.25}
\newcommand{\nTrainWinners}{7}
\newcommand{\paybackFiftyMin}{330}
\newcommand{\paybackFiftyMax}{1,480}
\newcommand{\paybackAllMin}{2,170}
\newcommand{\paybackAllMax}{5,210}
\newcommand{\paybackProbeMin}{20}
\newcommand{\paybackProbeMax}{140}
\newcommand{\trainMostN}{2}
\newcommand{\trainLessN}{4}

\newcommand{\trainGsmQttFifty}{+15.0}
\newcommand{\trainGsmQttAll}{+17.0}

\newcommand{\trainMathQeFifty}{+4.3}
\newcommand{\trainMathQeAll}{+11.5}

\newcommand{\trainNoFlipText}{a smaller training set never gives \LS{} the lead}

\newcommand{\labOldMmluMinAbs}{22.6}
\newcommand{\labOldMmluMaxAbs}{11.3}

\newcommand{\labWorseList}{MATH-500 with R1-Qwen-7B, MATH-500 with R1-Llama-8B and GPQA-D with Qwen3-8B}

\newcommand{\labGainQmathMin}{3.6}
\newcommand{\labGainQmathMax}{9.7}
\newcommand{\fZeroQmathMin}{3}
\newcommand{\fZeroQmathMax}{8}
\newcommand{\fZeroMmluMin}{18}
\newcommand{\fZeroMmluMax}{41}
\newcommand{\labFzeroShareMin}{62}
\newcommand{\labFzeroShareMax}{81}
\newcommand{\stFzMmluMin}{12.3}
\newcommand{\stFzMmluMax}{19.8}
\newcommand{\lsFzMmluMin}{8.0}
\newcommand{\lsFzMmluMax}{16.2}
\newcommand{\fZeroGpqaMin}{37}
\newcommand{\fZeroGpqaMax}{38}
\newcommand{\fineAgreeMin}{98.7}
\newcommand{\fineAgreeMax}{100.0}

\newcommand{\gridGainMax}{7.0}
\newcommand{\gridGainMean}{2.7}
\newcommand{\gridShrinkQmathMin}{1.4}
\newcommand{\gridShrinkQmathMax}{3.3}

\newcommand{\featDropBudget}{5.6}
\newcommand{\featDropConf}{9.7}

\newcommand{\featAloneMax}{22.2}
\newcommand{\featAloneBeatsText}{confidence features alone beat it only on GSM8K with Qwen3-32B and on MATH-500 with Qwen3-32B}
\newcommand{\featFixQm}{24.9}
\newcommand{\featConfMmluMin}{1.1}
\newcommand{\featConfMmluMax}{7.2}
\newcommand{\nBudgetOnlyMmlu}{three of the four}
\newcommand{\certLSBS}{22.7}

\newcommand{\certLSBU}{13.8}

\newcommand{\certLSHS}{0.9}

\newcommand{\certLSHU}{0.0}

\newcommand{\seedGsmDmin}{+15.7}
\newcommand{\seedGsmDmax}{+19.4}
\newcommand{\seedMmluDmin}{\ensuremath{-}5.2}
\newcommand{\seedMmluDmax}{+3.4}
\newcommand{\seedN}{3}

\newcommand{\seedGsmLSMin}{49.0}
\newcommand{\seedGsmLSMax}{51.7}
\newcommand{\seedMmluFixMin}{27.3}
\newcommand{\seedMmluFixMax}{31.9}

\newcommand{\seedSLabs}{1.3}
\newcommand{\seedGridText}{the fine static grid used throughout}
\newcommand{\seedMmluStaticN}{2}
\newcommand{\naquaEval}{508}

\newcommand{\ngpqaEval}{198}
\newcommand{\ngpqaTrain}{348}
\newcommand{\ngsmEval}{1{,}319}

\newcommand{\nmathEval}{500}

\newcommand{\shiftPairs}{44}
\newcommand{\shiftViolFix}{18}
\newcommand{\shiftMaxFix}{50.8}
\newcommand{\shiftDiagMaxFix}{3.4}
\newcommand{\shiftUpNFix}{22}
\newcommand{\shiftUpViolFix}{17}
\newcommand{\shiftDownNFix}{22}
\newcommand{\shiftDownViolFix}{1}

\newcommand{\shiftDownSaveFix}{6.7}
\newcommand{\shiftDownSaveOwnFix}{24.3}
\newcommand{\shiftViolLS}{18}
\newcommand{\shiftMaxLS}{45.0}
\newcommand{\shiftDiagMaxLS}{3.6}

\newcommand{\shiftUpViolLS}{16}

\newcommand{\shiftDownViolLS}{2}

\newcommand{\shiftDownSaveLS}{14.8}
\newcommand{\shiftDownSaveOwnLS}{28.0}

\newcommand{\shiftExFix}{22.6}
\newcommand{\shiftExLS}{21.6}
\newcommand{\shiftExFixSave}{70.9}
\newcommand{\shiftExLSSave}{74.1}
\newcommand{\exGsmRef}{940}
\newcommand{\exGsmKvFix}{33.6}

\newcommand{\exGsmKvSel}{49.3}
\newcommand{\exGsmKvSelAcc}{94.3}

\newcommand{\exGsmPcSel}{43.4}

\newcommand{\exGsmBbSel}{25.0}

\newcommand{\exGsmKvSelM}{460}
\newcommand{\exGsmKvFixM}{315}

\newcommand{\exMmluSelGapMax}{2.3}
\newcommand{\atMmluLost}{9.5}

\newcommand{\recalFixMax}{3.4}
\newcommand{\recalFixSave}{20.7}

\newcommand{\recalLSMax}{3.6}

\newcommand{\recalSizeHundredFix}{13.9}
\newcommand{\recalSizeTwoHundredFix}{19.1}
\newcommand{\recalSizeFullFix}{23.8}
\newcommand{\recalSizeHundredLS}{14.5}
\newcommand{\recalSizeTwoHundredLS}{17.9}
\newcommand{\recalSizeFullLS}{21.3}
\newcommand{\recalSizePairs}{37}
\newcommand{\recalMinN}{59}
\newcommand{\latNMain}{167}
\newcommand{\latNRep}{130}
\newcommand{\latNCtl}{50}
\newcommand{\latNNoc}{50}

\newcommand{\latMathFix}{0.93}

\newcommand{\latMathLSd}{0.80}
\newcommand{\latMathLSdLo}{0.72}
\newcommand{\latMathLSdHi}{0.87}

\newcommand{\perTokCtlMathLSd}{1.02}
\newcommand{\perTokNocMathLSd}{1.04}

\newcommand{\latProbesMath}{9.6}

\newcommand{\latPninetyMmluFull}{139}

\newcommand{\latPninetyMmluFix}{48}

\newcommand{\latMmluFix}{0.60}

\newcommand{\latPninetyMmluLSd}{63}

\newcommand{\latMmluLSd}{0.65}

\newcommand{\perTokCtlMmluLSd}{1.01}
\newcommand{\perTokNocMmluLSd}{1.02}

\newcommand{\latProbesMmlu}{8.3}
\newcommand{\latPerTokMin}{0.98}
\newcommand{\latPerTokMax}{1.01}
\newcommand{\latCtlPerTokMin}{0.99}
\newcommand{\latCtlPerTokMax}{1.02}

\newcommand{\latNocFewMin}{0.98}
\newcommand{\latNocFewMax}{1.01}
\newcommand{\agreeFixMin}{85}
\newcommand{\agreeFixMax}{94}

\newcommand{\agreeLSdMin}{53}
\newcommand{\agreeLSdMax}{58}

\newcommand{\agreeLSsMin}{75}
\newcommand{\agreeLSsMax}{77}

\newcommand{\earlyGapMax}{5}
\newcommand{\tokOffGap}{0.06}
\newcommand{\latAccNoise}{9}

\newcommand{\hidAucMath}{0.69}

\newcommand{\hidMathFixZero}{13.7}
\newcommand{\hidMathLSZero}{25.2}

\newcommand{\hidMathHSZero}{14.1}

\newcommand{\hidMathFixHigh}{7.1}

\newcommand{\hidMathHSHigh}{9.5}

\newcommand{\hidVarMinMath}{12.6}
\newcommand{\hidVarMaxMath}{16.6}
\newcommand{\hidRiskMath}{2.2}

\newcommand{\hidAucMmlu}{0.73}

\newcommand{\hidMmluFixZero}{31.1}
\newcommand{\hidMmluLSZero}{26.0}

\newcommand{\hidMmluHSZero}{18.9}

\newcommand{\hidVarMinMmlu}{18.0}
\newcommand{\hidVarMaxMmlu}{22.0}
\newcommand{\hidRiskMmlu}{3.3}
\newcommand{\wcFixZero}{17.0}
\newcommand{\meanFixZero}{20.5}
\newcommand{\wcLSZero}{14.3}
\newcommand{\meanLSZero}{22.7}

\newcommand{\meanSelZero}{23.7}

\newcommand{\wcLSPC}{27.4}

\newcommand{\wcSelPC}{5.2}

\newcommand{\wcLSBB}{25.9}

\newcommand{\wcSelBB}{3.8}

\newcommand{\wcFixMax}{17.0}
\newcommand{\wcLSMax}{27.4}
\newcommand{\wcSelMax}{11.6}
\newcommand{\wcSelZeroEx}{3.6}
\newcommand{\wcSelEx}{4.0}
\newcommand{\svGsmQttSTwo}{23.7}
\newcommand{\recalQPairs}{32}
\newcommand{\recalQFix}{21.1}
\newcommand{\recalQLS}{23.1}
\newcommand{\retrainQLS}{28.2}
\newcommand{\recalRPairs}{12}
\newcommand{\recalRFix}{19.4}
\newcommand{\recalRLS}{8.2}
\newcommand{\retrainRLS}{14.3}
\newcommand{\recalMinNTwo}{72}
\newcommand{\recalMinNFive}{90}

\newcommand{\refRQLSEnd}{8.2}

\newcommand{\refRLLSEnd}{0.0}
\newcommand{\refRQFixResp}{9.4}
\newcommand{\refRQLSResp}{19.6}
\newcommand{\refRLFixResp}{5.2}
\newcommand{\refRLLSResp}{6.9}
\usepackage{hyperref}
\hypersetup{colorlinks=true, linkcolor=blue!50!black, citecolor=blue!50!black, urlcolor=blue!50!black}

\newtheorem{proposition}{Proposition}

\newcommand{\LS}{\textsc{LearnStop}}
\newcommand{\LSC}{\textsc{LearnStop+cap}}
\newcommand{\SLS}{\textsc{Static-or-LearnStop}}

\begin{document}
\renewcommand{\topfraction}{0.9}
\renewcommand{\textfraction}{0.07}
\renewcommand{\floatpagefraction}{0.8}

\title{When Does Learning to Stop Help? Configuring Risk-Controlled Early Exit for Reasoning-Model Services}
\author{Zhe Dong\thanks{Corresponding author.}\\
  University of Maine at Presque Isle\\
  \texttt{zhe.dong@maine.edu}
  \and
  Fang Qin\\
  Stanford University\\
  \texttt{fangq@stanford.edu}
  \and
  Manish Shah\\
  Independent Researcher\\
  \texttt{shahmh@ieee.org}}
\date{}

\maketitle

\begin{abstract}
Early exit lowers the cost of serving a reasoning model by stopping its reasoning once a provisional answer looks
final. An operator must choose between a learned stopping rule and one static token budget, and
savings against full reasoning do not settle the choice. We present a procedure that decides on calibration
data from the workload. It certifies both families with Learn-then-Test to one bound on the fraction of all
questions that full reasoning answers correctly but early exit does not, prices provisional answers as the serving
stack does, and deploys the cheapest certified candidate. Across four models and five benchmarks neither family is
better throughout. At a 5\% risk target the learned rule saves \dQmathAbsMin--\dQmathAbsMax{} percentage points more
tokens than the static budget on free-form mathematics with Qwen3, with accuracy \dropQmathMin--\dropQmathMax{}
points below full reasoning, and less on MMLU-Pro on average. Which
family costs less also changes with the risk target, in \nSignChange{} of \nSettings{} settings, and with the price
of a provisional answer. Across three prices, committing to one family in advance gives up as much as \wcLSMax{}
points on some workload, whereas selecting on calibration data falls at most \wcSelMax{} points short of the better
family, on a task with only \gpqaNcal{} calibration questions. In a vLLM engine with Qwen3-8B, mean response time fell
in proportion to generated tokens.
Policies served on a task other than their calibration task exceeded the risk target in
\shiftViolFix{} of \shiftPairs{} task pairs; recalibration on the new task restored risk control.
\end{abstract}

\section{Introduction}
\label{sec:intro}

Reasoning models write out a chain of intermediate steps before they answer, and they answer better when they are
allowed a longer chain \citep{deepseek2025r1,qwen2025qwen3,muennighoff2025s1}. Whoever runs a service on such a model
pays for every token of that chain, and a good part of it is often unnecessary. On math benchmarks, for instance,
the answer a model would give if it were interrupted typically stops changing after about 60\% of its reasoning
steps \citep{liu2025convergence}. Early-exit methods make use of this. At chosen points they ask the model for a
provisional answer, and they end the reasoning once some signal, such as the model's confidence in that answer or
its agreement with the previous ones, indicates that the answer will not change.

Using early exit in a service requires several decisions, and they have to be made before any traffic is served. The
operator must decide what kind of rule to use: one that watches a single signal, one that is trained to combine
several, or none at all, in which case every question is simply cut off after the same number of tokens. A rule that
watches the reasoning must also be told how often to interrupt it and ask for an answer, because each interruption
costs tokens of its own, and a trained rule needs labelled questions. Finally, every form of early exit will
sometimes return a wrong answer to a question that the model would have answered correctly had it finished
reasoning. We say that such a question is lost, and the operator must decide what share of all questions may be
lost.

Published results give little help with these decisions. Early-exit methods are usually evaluated by the share of
tokens they save compared with letting the model reason to the end. That comparison leaves out the simplest
alternative, which is to give every question the same token budget. We call this a \emph{static budget}. It needs no
trained rule and interrupts each question only once, and over our settings it saves \uniFixA\% of the generated
tokens on average, nearly as much as the learned rule (\uniLSA\%). With Qwen3-32B on
the GSM8K arithmetic problems, and with at most 5\% of the questions lost, a static budget saves \svGsmQttFix\% of
the generated tokens. The learned stopping rule that we study in this paper, held to the same limit, saves
\dGsmQttabs{} percentage points more. On the multiple-choice questions of MMLU-Pro the ranking is reversed, and the
same learned rule saves \dMmluQttabs{} points less than the budget. Measured against full reasoning, the learned rule
looks useful on both workloads. Compared with the static budget, it saves more tokens on only one of them.

\begin{figure}[t]
\centering
\includegraphics[width=\linewidth]{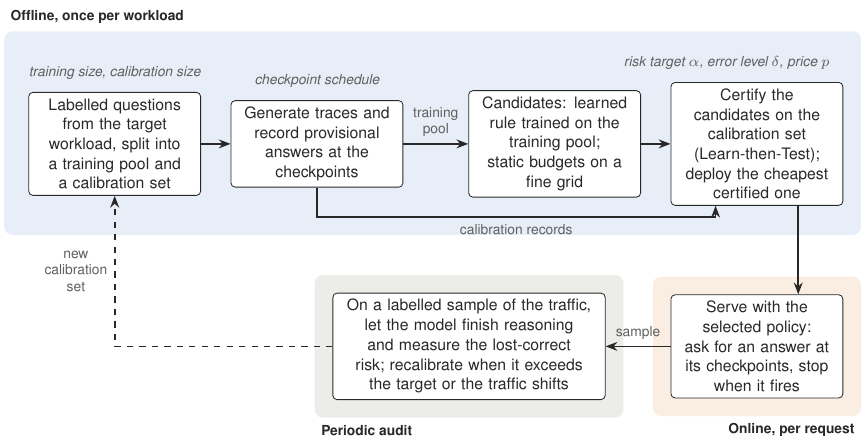}
\caption{The configuration procedure. Offline, once per workload, labelled questions are split into a training pool
and a calibration set, and the model's reasoning and its provisional answers at the checkpoints are recorded for
both. The training pool is used to fit the learned rule. The calibration set is used to certify every candidate, of
either family, at the risk target, and the cheapest certified candidate at the price of the serving stack is
deployed. Online, every request is served with the selected policy. A periodic audit runs both the selected policy
and full reasoning on a labelled sample of the traffic and measures the share of questions that only full reasoning
answers correctly; the policy is recalibrated when this share exceeds the target or the traffic changes. Italic
labels mark the quantities that a deployment sets; $\delta$ is the error level of the certification, so the bound
holds with confidence $1-\delta$.}
\label{fig:pipeline}
\end{figure}

This paper describes a procedure for making these decisions on the workload that will actually be served
(Figure~\ref{fig:pipeline}). The operator collects a sample of questions with known answers and records the
provisional answers of the model on them. From these records the procedure builds two families of candidate
policies. One consists of static budgets of different lengths. The other consists of learned rules, all obtained from
one small classifier that reads features of the provisional answers, by applying different stopping thresholds to
its score. Each candidate is then tested on calibration questions that were held
out, using Learn-then-Test \citep{angelopoulos2025ltt}. This statistical procedure certifies, with a chosen
confidence, that the share of questions a candidate loses stays below a target, provided that the questions served
later come from the same distribution as the calibration questions. We call this share the \emph{lost-correct risk}.
Among the certified candidates the procedure deploys the cheapest one, and the cost of a candidate includes its
provisional answers at the prices of the operator's serving stack.

The same steps tell us what a learned rule adds. Applied to each family separately, they give the cheapest certified
static budget and the cheapest certified learned rule. Both have passed the same test under the same accounting. We call the saving of the learned rule minus that of the
static budget the \emph{increment} of learned stopping, written $\Delta$, so that a positive $\Delta$ means that
learning pays.

We apply the procedure to four open-weight reasoning models from the Qwen3 and DeepSeek-R1 families and to five
benchmarks, \nSettings{} combinations of model and task in all. The experiments follow six questions that an operator
has to answer:
\begin{itemize}
\item[RQ1] Does a learned rule save more than a certified static budget, and on which workloads?
\item[RQ2] How do the risk target and the price of a provisional answer change the comparison, and how often should a
service ask for one?
\item[RQ3] What is gained by choosing the family on calibration data instead of committing to one in advance?
\item[RQ4] How many labelled questions does a learned rule need, and when is collecting them repaid?
\item[RQ5] Do the token savings become shorter response times in a serving engine?
\item[RQ6] Does a deployed policy still meet the risk target after a change of workload, and what restores risk
control if it does not?
\end{itemize}
We also compare the learned rule with two alternatives outside the procedure, a rule that reads the hidden states of
the model and a model fine-tuned to reason less (Section~\ref{sec:robust}).

Neither family is better across conditions. Which one saves more depends on the workload, on the risk target and on
what the serving stack charges for a provisional answer; in \nSignChange{} of the \nSettings{} combinations the better
family changes at least once as the permitted share of lost questions rises from 5\% to 20\%. An operator who always
uses the static budget gives up as much as \wcFixMax{} percentage points of saving on some workload, and one who
always uses the learned rule as much as \wcLSMax{} points. Choosing on calibration data avoids this: the selected
policy falls at most \wcSelEx{} points short of the better family at any of the three prices, except on the task
with only \gpqaNcal{} calibration questions. Two further results concern the
service itself. In a serving engine that lets a provisional answer reuse the cached reasoning, the mean response
time fell in proportion to the tokens generated. And the risk control belongs to the workload on which a policy was
calibrated: policies calibrated on one benchmark and served on another lost more than the permitted 5\% of the
questions in \shiftViolFix{} of \shiftPairs{} pairs of benchmarks, and recalibration on the served benchmark brought
the loss back under the target in every pair.

The paper makes three contributions.
\begin{itemize}
\item A configuration procedure for early exit in reasoning-model services. It applies two existing components,
Learn-then-Test certification and the selection of the cheapest certified candidate, to learned rules and static
budgets under one risk bound and one cost model. Two design choices affect the outcome: the classifier is trained to predict whether stopping
would lose the question, which is not the same as whether the provisional answer is correct, and the static budget
is chosen from a fine grid of lengths (Section~\ref{sec:method}).
\item An evaluation over \nSettings{} combinations of model and task, which shows how the increment of learned
stopping depends on the workload, the risk target and the price of a provisional answer, and explains where each
family leads (Sections~\ref{sec:value}--\ref{sec:conditions}).
\item Measurements that an operator can act on: the gain and cost of selecting on calibration data, the labelled
data a learned rule needs, response times in a serving engine, and what recalibration recovers after a change of
workload (Sections~\ref{sec:select}--\ref{sec:shift} and~\ref{sec:guidance}).
\end{itemize}
The analysis code and its outputs are available at \url{https://github.com/dongzhe1/llm-think-stop}, and
the logged provisional answers, with their token probabilities, from the authors on request. Any stopping rule that uses only these records
can be evaluated on them without a GPU.

\section{Related work}
\label{sec:related}

\subsection{Adaptive computation and cost control}

Stopping a computation early on easy inputs is older than reasoning models: classifiers attached to intermediate
layers let confident inputs leave a network early \citep{teerapittayanon2016branchynet,xin2020deebert,schwartz2020right}.
Services built on large language models also control cost across models, by routing each query to a suitable model
\citep{chen2023frugalgpt,varangot2026routing} or by spending extra computation only when the model appears uncertain
\citep{yin2026failure,ma2026guide}.

Early exit in a reasoning model addresses the same concern within a single request: the model is fixed, and the
decision is how long it reasons. Recent surveys cover test-time compute in general \citep{ji2026testtime} and
efficient inference for reasoning models \citep{liu2026efficient}. The second distinguishes methods that shorten
the written chain of thought from methods that move the reasoning into latent representations, as RuPLaR does
\citep{luo2027ruplar}. Our setting belongs to the first group, and it leaves the weights of the model unchanged.

\subsection{Budgets on reasoning length}

The simplest way to limit the length of a chain is to fix it. Work on test-time scaling has shown that accuracy
rises on average with the number of reasoning tokens \citep{muennighoff2025s1}, and work on token budgets that the
length a question needs varies from one question to the next \citep{han2025tale}. Qwen3 lets the user set a thinking
budget directly \citep{qwen2025qwen3}. A static budget whose length is certified against a risk target is the
plainest policy of this kind. We use it as the baseline that any rule which interrupts the reasoning more than once
has to beat.

\subsection{Exit rules based on provisional answers}

Most early-exit rules for reasoning models watch a provisional answer and stop when it looks final
(Table~\ref{tab:related}). One group takes the model's confidence in the answer as the sign: its probability
\citep{yang2026deer}, the entropy after an appended end-of-reasoning marker \citep{wang2025eat}, or an abrupt rise
in its probability \citep{tikhonov2026confidence}. A second group takes the stability of the answer over
successive requests \citep{fu2024certaindex,liu2025convergence}. PUMA adds a
trained detector that recognises when new reasoning steps add nothing, and then checks that the trial answer is
stable and confident \citep{min2026puma}.

Our comparison includes four signals that are computed from the provisional answer: confidence, entropy, a jump in
confidence, and agreement of consecutive answers. We implement all of them over the same checkpoints and the same form of provisional answer,
so the comparison is between kinds of signal under one protocol. It is not a reproduction of the original systems,
which differ in where they interrupt the model and how they phrase the request for an answer.

\subsection{Learned stopping rules and risk control}

A stopping rule can also be learned from examples. \citet{liu2025convergence} train one on the internal activations
of the model, and TERMINATOR trains a classifier on hidden states to predict where the final answer first appears in
the chain \citep{nagle2026terminator}. Neither attaches a statistical guarantee to the rule. Thought Calibration does:
it trains probes on hidden states and sets their stopping threshold with Learn-then-Test \citep{wu2025thought}.
Learn-then-Test \citep{angelopoulos2025ltt} is a general method for choosing a threshold so that a chosen risk
stays below a target with a stated confidence on a finite calibration sample; conformal risk control
\citep{angelopoulos2024conformalrisk} bounds the expected risk instead. They have been applied to language generation \citep{quach2024conformal}, and
\citet{campos2024conformal} survey conformal methods in language processing.

Conformal Thinking \citep{wang2026conformal} is the closest work to ours. It considers several stopping signals,
among them the confidence of the answer, the entropy signal of EAT, a probe on hidden states and the number of
tokens generated so far. For each signal it calibrates two thresholds: one at which the model stops because it is confident, and one at which
it gives up on a question that it is unlikely to solve. Among the
pairs of signal and threshold whose risk passes a finite-sample bound, it keeps the most efficient one,
and it reports that no single signal is the most efficient at every risk level.

Like Conformal Thinking, we deploy the cheapest candidate that passes a risk certification. We differ from it in
three respects. It treats the number of tokens as one signal among several, whereas we make the certified static
budget the baseline against which every other rule is measured. We charge each provisional answer the tokens it
costs, including the reasoning that has to be sent to the model again when the serving stack does not keep it. And we
train the learned rule on a label that matches the certified risk. We consider only policies that return an answer,
and do not study the option to give up. The statistical test also differs in its details, which
Section~\ref{sec:cert} describes.

\subsection{Changing the model or the prompt}

The length of the reasoning can also be reduced by changing the model. AdaptThink uses reinforcement learning to
teach DeepSeek-R1-Distill-Qwen models to skip the reasoning on easy inputs \citep{zhang2025adaptthink}, and MARM
fine-tunes a medical reasoning model to answer directly or with a short or a long chain of thought, depending on
the question \citep{hou2026marm}. CORE works through the prompt instead, instructing the model to select reasoning
stages according to the problem \citep{wang2026core}. These approaches report average accuracy and token use, and
none of them calibrates the trade-off between the two to a risk bound. We include the released 7B checkpoint of
AdaptThink in our experiments, as the fine-tuned alternative to stopping the same base model at inference time
(Section~\ref{sec:adaptthink}).

\begin{table}[t]
\centering
\caption{Early-exit and adaptive-length methods for reasoning models that are closest to this work. Reads: what the
stopping rule reads when it decides; a dash where there is no stopping rule because the model or the prompt is
changed instead. Risk bound: whether the stopping rule is calibrated to a finite-sample bound on a risk.}
\label{tab:related}
\smallskip
\small
\begin{tabular}{>{\raggedright\arraybackslash}p{0.27\textwidth}>{\raggedright\arraybackslash}p{0.36\textwidth}>{\raggedright\arraybackslash}p{0.17\textwidth}>{\raggedright\arraybackslash}p{0.09\textwidth}}
\toprule
Method & Stopping signal & Reads & Risk bound \\
\midrule
DEER \citep{yang2026deer} & confidence of trial answers at transition points & output probabilities & no \\
EAT \citep{wang2025eat} & entropy after an appended end-of-thinking marker, stopped when its moving variance is low & output probabilities & no \\
Confidence Leaps \citep{tikhonov2026confidence} & jump in answer probability & output probabilities & no \\
Answer convergence \citep{liu2025convergence} & agreement of consecutive answers; learned rule on activations & outputs; activations & no \\
Dynasor/Certaindex \citep{fu2024certaindex} & agreement of provisional answers over a window of steps & outputs & no \\
PUMA \citep{min2026puma} & redundant reasoning steps flagged by a trained detector, then a stable, confident trial answer & reasoning text; output probabilities & no \\
TERMINATOR \citep{nagle2026terminator} & classifier on hidden states & hidden states & no \\
Thought Calibration \citep{wu2025thought} & probes on hidden states & hidden states & yes \\
Conformal Thinking \citep{wang2026conformal} & confidence, entropy, hidden-state probe or token count & outputs or hidden states & yes \\
AdaptThink \citep{zhang2025adaptthink}, MARM \citep{hou2026marm} & none: the model is fine-tuned to choose how much to think & -- & no \\
CORE \citep{wang2026core} & none: the prompt guides the choice of reasoning stages & -- & no \\
\midrule
This work (main procedure) & static budget, or logistic classifier on features of the provisional answers, selected on calibration data & token count; provisional answers and their probabilities & yes \\
This work (comparison, Section~\ref{sec:hidden}) & logistic classifier on hidden states & hidden states & yes \\
\bottomrule
\end{tabular}
\end{table}

\section{Configuration procedure}
\label{sec:method}

This section defines the steps of Figure~\ref{fig:pipeline} in the order in which a reader needs them: how a
provisional answer is obtained, which risk is controlled, what a policy costs, which policies are candidates, and
how they are certified and selected.

\subsection{Provisional answers}
\label{sec:probe}

A reasoning model answers a question $q$ in two phases. It first writes a reasoning trace $r=(r_1,\dots,r_T)$ of $T$
tokens, which it closes with the marker \texttt{</think>}, and then it writes its response. To find out what the
model would answer after only $B$ reasoning tokens, we give it the question followed by the first $B$ tokens of the
trace, close the reasoning ourselves, and ask for the answer. Concretely, the model continues the text
\[
q \oplus r_{1:B} \oplus \texttt{</think>}\,\texttt{Final answer:}
\]
by greedy decoding. Decoding stops at the first line break that follows non-blank text, or at the end of sequence,
or after 48 tokens, and the text up to that point is kept. We call this operation a
\emph{probe} at budget $B$, its result the probe answer, and the number of tokens decoded up to the stop $\ell_B$. A probe works
on a copy of the trace and leaves the trace unchanged; if the policy decides not to stop, the model
continues reasoning from token $B+1$.

An example from our records shows what probes return. Qwen3-32B was asked: ``Darrell and Allen's ages are in the ratio
of 7:11. If their total age now is 162, calculate Allen's age 10 years from now.'' The correct answer is 109, and the
model reasoned for 1{,}060 tokens. Probes after 0, 128, 256, 384, 512, 768 and 1{,}024 reasoning tokens returned 99,
99, 112, 119, 109, 109 and 109, each in four or five decoded tokens. The model had the correct answer after 512
tokens and spent the second half of its reasoning confirming it. A policy that stops at 512 tokens or later returns
the correct answer, and a policy that stops at 256 returns 112. The seven budgets in this example are the first of
14 \emph{checkpoints}, between 0 and 12{,}288 tokens, at which the policies that probe repeatedly do so in our
experiments (Section~\ref{sec:setup} lists them).

The answer of full reasoning is obtained in the same way, by a probe placed at the end of the complete trace, which
we call the end probe. A trace is complete when the model closes it or when it reaches a cap of $T_{\max}$ tokens.
We write $F(q)=1$ if the end-probe answer is correct and $F(q)=0$ otherwise, and we take the cost of full reasoning
to be $T+\ell_{\mathrm{end}}$ tokens. With one format for early and final answers, two answers to the same question differ only in how much reasoning
preceded them. Section~\ref{sec:ref} repeats the main comparison with the model's own response after the trace as the
reference.

\subsection{Lost-correct risk}
\label{sec:risk}

A stopping policy observes probes at increasing budgets $B_0<B_1<\dots$ and either stops at one of them, returning
its probe answer, or lets the trace finish and returns the end-probe answer. We index policies by a parameter
$\lambda$, which will be a budget or a threshold, and write $C_\lambda(q)=1$ if the answer the policy returns for
$q$ is correct. The policy \emph{loses} a question when full reasoning answers it correctly and the policy does not:
\[
L_\lambda(q)=\mathbf{1}\{F(q)=1,\ C_\lambda(q)=0\}.
\]
In the example above, a policy that stops at 256 tokens loses the question. The lost-correct risk of a policy is the
expected loss over the questions of the workload, $R(\lambda)=\mathbb{E}[L_\lambda]$, that is, the share of all
questions that it loses.

We control this risk and not the accuracy of the policy because it isolates the effect of stopping: a question
that full reasoning answers wrongly cannot be lost. The risk still bounds the drop in accuracy, since a policy can
also gain questions by stopping on a correct answer that further reasoning would have abandoned, so that
$\mathrm{Acc}_{\mathrm{full}}-\mathrm{Acc}_\lambda=R(\lambda)-\Pr(F=0,\,C_\lambda=1)\le R(\lambda)$. Every policy in
this paper is held to the same requirement $R(\lambda)\le\alpha$, where $\alpha$ is the risk target chosen by the
operator.

\subsection{Cost of a policy}
\label{sec:cost-model}

The cost of a policy on a question has two parts. The first is the number of tokens the model decodes: the reasoning
tokens up to the point where the policy stops, and the tokens of every probe made on the way. A policy that probes
at $B_0,\dots,B_j$ and stops at $B_j$ decodes $B_j+\sum_{k\le j}\ell_{B_k}$ tokens. In the example, a policy that
probes at every checkpoint and stops at 768 decodes $768+28$ tokens, a static budget of 512 decodes $512+5$, and full
reasoning $1{,}060+5$.

The second part depends on the serving stack. A probe needs the question and the reasoning so far as its input, and
so does the call that resumes the reasoning after a probe that did not stop. If the engine can branch from its
key--value cache of the tokens already processed, which we call a KV-cache fork, this input costs nothing. Otherwise
the prefix has to be sent again and is charged as input. We write $p$ for the price of one re-sent input token
relative to one output token. With a prompt of $P$ tokens, a policy that stops at $B_j$ then costs
\begin{equation}
c = B_j + \sum_{k\le j}\ell_{B_k} + p\Big[\sum_{k\le j}(P+B_k) + \sum_{k<j}(P+B_k)\Big],
\label{eq:cost}
\end{equation}
where the first sum in the bracket counts the probe calls and the second the calls that resume the trace. A policy
that never stops pays for all its probes and resumptions on top of the full trace, namely
$T+\sum_k\ell_{B_k}+\ell_{\mathrm{end}}+2p\sum_k(P+B_k)$. A static budget at $B<T$ makes one probe and no
resumption, and costs $B+\ell_B+p(P+B)$. The end probe is charged its output tokens only: we treat the final answer
as the continuation of the call that produced the trace, for full reasoning and for every policy that does not
stop, so no prefix is counted for it.

Three values of $p$ correspond to common serving stacks. A KV-cache fork has $p=0$. A prefix cache bills cached
input at a discount: if input costs a quarter of output and cached input a tenth of that, $p$ is about 0.025. A
black-box API without caching bills the re-sent prefix as ordinary input, so $p$ is about 0.25. All costs are thus
expressed in output-token equivalents, and with $p=0$ they are counts of generated tokens. The \emph{saving} of a
policy is $1-\bar c_\lambda/\bar c_{\mathrm{ref}}$, where the bars denote means over test questions and
$c_{\mathrm{ref}}=T+\ell_{\mathrm{end}}$ is the cost of full reasoning. Processing the initial prompt is common to
all policies and is left out.

\subsection{Candidate policies}
\label{sec:policies}

\textbf{Static budget.} A static budget $B$ cuts every trace after $B$ tokens and probes once. Questions whose trace
is shorter than $B$ are answered by the end probe. Because the deployed policy makes a single probe, $B$ can be any
length, and the candidates are the lengths of a fine grid: every 128 tokens up to 1{,}024 and every 256 tokens
beyond. With only the 14 checkpoints of the other policies as candidates, the selected budget would often have a risk
well below the target while the next shorter one fails certification (Section~\ref{sec:value}).

\textbf{LearnStop.} The learned rule, which we call \LS, probes at a schedule of checkpoints. At each checkpoint a
logistic regression predicts, from eight features, the probability that stopping now is safe, in a sense made
precise in the next paragraph, and \LS{} stops at the first checkpoint where this probability exceeds a threshold
$\tau$. The candidates of the family are the values of $\tau$. The features fall into four groups (Table~\ref{tab:features-def}): the current budget, the confidence of the
current probe answer, the stability of the answers across checkpoints, and markers of backtracking in the
reasoning. The rule thus reads only what the
model outputs, namely the text and token probabilities of the probes and the text of the reasoning. It needs no
access to hidden states, and it leaves the weights of the model unchanged. Its default schedule is the full set of
14 checkpoints, and Section~\ref{sec:conditions} considers schedules with 7, 4 or 2.

\LS{} is trained to predict whether stopping at a checkpoint is \emph{safe}, meaning that it does not lose the
question. The label is $1-L$: it is one when the probe answer is correct, and also when full reasoning is wrong
anyway. The more obvious label would be whether the probe answer is correct. The two differ only at checkpoints
where the probe answer is wrong and full reasoning is wrong as well. There the correctness label discourages stopping although stopping loses nothing, and Section~\ref{sec:value}
shows what this costs. The classifier is trained on a pool of labelled questions from the same
task, disjoint from the calibration and test questions.

\begin{table}[t]
\centering
\caption{The eight features of \LS{} at a checkpoint.}
\label{tab:features-def}
\smallskip
\small
\begin{tabular}{>{\raggedright\arraybackslash}p{0.14\textwidth}>{\raggedright\arraybackslash}p{0.76\textwidth}}
\toprule
Group & Features \\
\midrule
Budget & logarithm of the current budget; index of the checkpoint \\
Confidence & mean log-probability and mean entropy of the tokens of the current probe answer \\
Stability & whether the current answer repeats the previous one; length of the current run of identical answers;
share of the probes so far that gave this answer \\
Markers & density of words that signal backtracking, such as ``wait'', in the reasoning so far \\
\bottomrule
\end{tabular}
\end{table}

\textbf{Static-or-LearnStop.} The selection step of the procedure treats the static budgets and the \LS{} thresholds
as two families of one policy, which we call \SLS, and deploys the cheapest certified candidate from either family.
Further families can be added in the same way. We use this to choose among checkpoint schedules, by entering each
schedule as a family of its own.

\textbf{Comparison policies.} Three further kinds of policy serve as comparisons. The first are four \emph{scalar
exits}, which stop at the first checkpoint where a single score exceeds a threshold. The scores are the confidence
of the probe answer, the signal used by DEER; its negative entropy, the quantity that EAT monitors (EAT itself stops
when the moving variance of this entropy is small); the number of consecutive
checkpoints that returned the same answer \citep{liu2025convergence}; and a confidence leap, after
\citet{tikhonov2026confidence}. Appendix~\ref{app:data} defines them, and \emph{best scalar} is the policy that
selects among the four.

The second, \LSC, forces \LS{} to stop at a hard cap of 512, 1{,}024, 2{,}048, 4{,}096 or 8{,}192 tokens if it has
not stopped before. Each cap forms a family whose first candidate is the static budget of that length and whose
further candidates are the thresholds of \LS{} under that cap, and \LS{} without a cap is a sixth family. Comparing
\LSC{} with \SLS{} shows whether combining a budget with the learned rule saves more than selecting between them.

The third, \textsc{HiddenStop}, replaces the eight features by a hidden state of the model
(Section~\ref{sec:hiddensetup}).

\subsection{Certification and selection}
\label{sec:cert}

Certification has to establish, from a finite sample, that the risk of a candidate is at most $\alpha$. We use
Learn-then-Test \citep{angelopoulos2025ltt}, which treats this as a hypothesis test. For a candidate $\lambda$ the
null hypothesis is that it is unsafe, $H_0\colon R(\lambda)>\alpha$. If the candidate loses $k_\lambda$ of $n$
calibration questions, the p-value of this hypothesis is the probability that a binomial variable with $n$ trials
and success rate $\alpha$ is at most $k_\lambda$, $p_\lambda=\Pr[\mathrm{Bin}(n,\alpha)\le k_\lambda]$. A small
p-value means that so few losses would be unlikely if the risk really exceeded $\alpha$, and the candidate is then
certified.

Many candidates are tested on the same calibration questions, and the tests must be arranged so that the chance of
certifying any unsafe candidate stays below a level $\delta$. Within a family the candidates have a natural order
from conservative to aggressive: from long budgets to short ones, and from high thresholds to low ones. They are
tested in this order at a fixed level, and testing stops at the first candidate that fails, which is known as
fixed-sequence testing. A policy with $m$ families divides $\delta$ equally, so that each family is tested at level
$\delta/m$. The static budget and \LS{} have one family each, \SLS{} has two, best scalar four and \LSC{} six. The
thresholds of a family are 101 quantiles of its score on the training pool, where the score is the probability
predicted by the classifier or, for a scalar exit, the scalar itself. The candidates and their order are thus fixed
before any calibration label is seen.

Selection then takes place among the certified candidates. To these we add full reasoning as a fallback: the trace
is left to finish without any earlier probe and the end-probe answer is returned, which by definition loses no
question. The procedure deploys the candidate with the lowest mean cost on the calibration questions, computed at the price $p$ of
the serving stack. A policy therefore falls back to full reasoning when none of its certified candidates is cheaper. Conformal Thinking \citep{wang2026conformal} selects in the same way, keeping the feasible candidate with the lowest
estimated efficiency loss; pricing the candidates at the serving stack and the fallback to full reasoning are our
additions. The certification differs: Conformal Thinking compares an upper confidence bound on the risk of each
candidate with the target, where we use the sequential test described above.

The guarantee that results can be stated precisely. It rests on one assumption: the calibration questions and the
questions served later are drawn independently from the same distribution, and the trace and the probes of each
question are generated with randomness that is independent across questions. The losses $L_\lambda(q)$ of a
candidate on the calibration questions are then independent Bernoulli variables with mean $R(\lambda)$, where the
risk $R(\lambda)$ averages over both the question and its generation.

\begin{proposition}
\label{prop:ltt}
Under this assumption, let $\hat\Lambda$ be the set of candidates certified by the procedure above, together with
the fallback to full reasoning. Then $\Pr\big[R(\lambda)\le\alpha \text{ for every } \lambda\in\hat\Lambda\big]\ge 1-\delta$
over the draw of the calibration questions and the generation of their traces and probes. Consequently any candidate chosen from $\hat\Lambda$ on the basis of the
calibration data, and in particular the one with the lowest calibration cost, has $R\le\alpha$ with probability at
least $1-\delta$.
\end{proposition}

The proposition is a direct application of Learn-then-Test, and Appendix~\ref{app:proof} gives the short proof. We
set $\delta=0.05$ throughout and consider risk targets $\alpha$ of 0.05, 0.10, 0.15 and 0.20. Most results are
reported at $\alpha=0.05$, which limits the drop in accuracy to at most five percentage points. Section~\ref{sec:shift}
examines what happens when the assumption fails because the served questions come from a different task.

\subsection{The increment of learned stopping}
\label{sec:delta}

To measure what the learned rule adds, we compare \LS{} and the static budget as two separate policies. For a
setting, a risk target $\alpha$ and a price $p$, each is certified and selected on the same calibration questions,
and its saving is measured on the same test questions. With $S_{\mathrm{LS}}(\alpha,p)$ and
$S_{\mathrm{st}}(\alpha,p)$ the two savings, the increment is
\begin{equation}
\Delta(\alpha,p)=S_{\mathrm{LS}}(\alpha,p)-S_{\mathrm{st}}(\alpha,p).
\label{eq:delta}
\end{equation}
Both policies satisfy the bound, but the risk they actually incur on the test questions can differ, and
Section~\ref{sec:value} therefore also compares them at equal realised risk.

\section{Experimental setup}
\label{sec:setup}

\subsection{Models, traces and probes}

We use four open-weight reasoning models: Qwen3-8B and Qwen3-32B in thinking mode \citep{qwen2025qwen3}, and two
distillations of DeepSeek-R1, DeepSeek-R1-Distill-Qwen-7B and DeepSeek-R1-Distill-Llama-8B \citep{deepseek2025r1},
which we abbreviate to R1-Qwen-7B and R1-Llama-8B. Open weights are needed because the study records the token
probabilities of every probe. For each question one reasoning trace is sampled with the settings recommended by the
developers (temperature 0.6, top-$p$ 0.95, and top-$k$ 20 for Qwen3) and a fixed seed, and all probes are decoded
greedily. The cap $T_{\max}$ on the length of a trace is 8{,}192 tokens for GSM8K and AQuA, 12{,}288 for MMLU-Pro
and 16{,}384 for MATH-500 and GPQA-Diamond. Generation runs in vLLM 0.30.0 with prefix caching on NVIDIA H100 GPUs
\citep{kwon2023vllm}.

Probes are recorded at two sets of budgets. The policies that probe repeatedly use 14 checkpoints, at 0, 128, 256,
384, 512, 768, 1{,}024, 1{,}536, 2{,}048, 3{,}072, 4{,}096, 6{,}144, 8{,}192 and 12{,}288 tokens, with closer spacing
early in the trace. For the static budget, each recorded trace is probed again on the
fine grid of Section~\ref{sec:policies}, which gives between 36 and 68 budgets below $T_{\max}$ depending on the
task. Where a budget of the fine grid coincides with a checkpoint, the analysis keeps the original
probe. The second pass ran as a separate job, so a repeated probe is not guaranteed to be identical to the original;
it receives the same correctness label on \fineAgreeMin--\fineAgreeMax\% of the questions.

\subsection{Tasks and data}

The five tasks cover free-form and multiple-choice answers. GSM8K \citep{cobbe2021gsm8k} and MATH-500
\citep{hendrycks2021math,lightman2024verify} are mathematics problems with free-form answers. AQuA
\citep{ling2017aqua} consists of math word problems in multiple-choice form. MMLU-Pro \citep{wang2024mmlupro} and
GPQA-Diamond \citep{rein2023gpqa}, GPQA-D below, are multiple-choice questions across academic subjects, the latter
at graduate level. Not every model is run on every task: the four models are run on GSM8K, MATH-500 and MMLU-Pro,
Qwen3-8B also on AQuA, and both Qwen3 models on GPQA-D, which gives the \nSettings{} combinations of
Table~\ref{tab:settings}. We refer to a combination of task and model as a \emph{setting}. The table also gives the accuracy of the end probe and of
the model's own response after the trace. For Qwen3 the two differ by at most \gapQwenMax{} percentage points on
GSM8K, MATH-500 and MMLU-Pro and by \gapAqua{} on AQuA.

Each task provides two disjoint sets of questions, an evaluation set and a training pool. The evaluation set is
later divided at random into a calibration half and a test half (Section~\ref{sec:protocol}). The training pool is
used to fit the learned rules and to fix the threshold grids of all policies, and it plays no part in calibration or
testing. The evaluation sets are taken from the benchmarks: the test split of GSM8K (\ngsmEval{} questions),
MATH-500 (\nmathEval), 1{,}000 questions from the test split of MMLU-Pro, the test and validation splits of AQuA
combined (\naquaEval), and GPQA-D (\ngpqaEval). The training pools hold 500 questions from the training split of the task, which for MATH-500 is the
training split of the MATH dataset. There are two exceptions. For MMLU-Pro the pool is drawn from test questions
outside the evaluation set, and for GPQA-D it consists of the \ngpqaTrain{} questions of GPQA-Extended that are not
in Diamond. Answers are graded automatically, by numeric comparison or by the option letter, and for MATH-500 with
the answer normalisation of the benchmark's release code (Appendix~\ref{app:data}).

\subsection{Evaluation protocol}
\label{sec:protocol}

A single division into calibration and test questions would make the results depend on which questions happen to
fall on which side. For each setting we therefore draw 50 random divisions of the evaluation set into two halves. On
each division every policy is certified and selected on the calibration half, and its saving, its accuracy and its
realised risk are measured on the test half. We report means over the 50 divisions, which we call splits.

Comparisons between two policies are paired: the difference in saving is computed on the same test half and then
summarised over the splits by its mean, by the range between its 2.5\% and 97.5\% quantiles, and by the share of
splits in which the first policy saves more. This range describes the variation over divisions of one evaluation
set and shows whether the sign of a difference is stable; it is not a confidence interval for a new sample of
questions. In the tables a difference is set in bold when the range excludes zero. A bootstrap over the questions of a test
half gives intervals of similar width (Appendix~\ref{app:boot}). Averages over
settings give each of the \nSettings{} settings equal weight.

\begin{table}[t]
\centering
\caption{The \nSettings{} settings. Accuracy is given for the model's own response after the trace (natural) and for
the end probe on the complete trace, which is the reference. Thinking tokens: mean length of the trace. Unclosed:
share of traces that reached $T_{\max}$. Probed ckpts: mean number of checkpoints that fall before the end of the
trace. Probe tokens: every recorded probe was decoded to the end of sequence or the 48-token cap; the two columns give
its mean length up to the first line break, where a deployed probe stops and which the costs use, and its full
recorded length.}
\label{tab:settings}
\smallskip
\small\setlength{\tabcolsep}{4pt}
\begin{adjustbox}{max width=\textwidth}\begin{tabular}{lrrrrrrrr}
\toprule
 & & \multicolumn{2}{c}{accuracy} & thinking & unclosed & probed & \multicolumn{2}{c}{probe tokens} \\
\cmidrule(lr){3-4}\cmidrule(lr){8-9}
Task / model & $n$ & natural & end probe & tokens & (\%) & ckpts & to newline & to EOS/cap \\
\midrule
GSM8K / Qwen3-8B & 1319 & .956 & .959 & 1,459 & 0.9 & 6.7 & 6.0 & 37.6 \\
GSM8K / Qwen3-32B & 1319 & .966 & .968 & 932 & 0.1 & 5.5 & 4.7 & 8.4 \\
GSM8K / R1-Qwen-7B & 1319 & .892 & .832 & 189 & 0.0 & 1.8 & 7.4 & 9.5 \\
GSM8K / R1-Llama-8B & 1319 & .790 & .660 & 153 & 0.0 & 1.7 & 5.1 & 42.6 \\
\midrule
MATH-500 / Qwen3-8B & 500 & .914 & .920 & 4,309 & 3.2 & 9.6 & 11.0 & 31.1 \\
MATH-500 / Qwen3-32B & 500 & .924 & .936 & 3,575 & 2.4 & 9.0 & 9.7 & 10.4 \\
MATH-500 / R1-Qwen-7B & 500 & .886 & .882 & 2,767 & 3.6 & 7.3 & 10.7 & 12.4 \\
MATH-500 / R1-Llama-8B & 500 & .778 & .724 & 2,482 & 3.0 & 5.8 & 10.4 & 28.9 \\
\midrule
MMLU-Pro / Qwen3-8B & 1000 & .739 & .759 & 3,062 & 4.2 & 8.2 & 4.8 & 33.2 \\
MMLU-Pro / Qwen3-32B & 1000 & .803 & .823 & 2,106 & 2.3 & 7.0 & 4.6 & 4.9 \\
MMLU-Pro / R1-Qwen-7B & 1000 & .575 & .588 & 2,794 & 4.0 & 7.8 & 5.4 & 18.5 \\
MMLU-Pro / R1-Llama-8B & 1000 & .593 & .604 & 3,171 & 4.4 & 8.2 & 5.6 & 19.8 \\
\midrule
AQuA / Qwen3-8B & 508 & .858 & .900 & 2,564 & 7.7 & 8.0 & 4.5 & 41.8 \\
\midrule
GPQA-D / Qwen3-8B & 198 & .611 & .611 & 6,809 & 2.5 & 11.2 & 4.8 & 35.2 \\
GPQA-D / Qwen3-32B & 198 & .616 & .626 & 4,943 & 1.0 & 10.0 & 4.3 & 4.5 \\
\bottomrule
\end{tabular}\end{adjustbox}
\end{table}

\subsection{Online measurement and hidden states}
\label{sec:onlinesetup}\label{sec:hiddensetup}

Two further experiments go beyond the recorded probes. Both use Qwen3-8B on two tasks: MATH-500, where the learned
rule saves more on average, and MMLU-Pro, where the static budget does.

The first serves the certified policies in the engine and measures response time. The policies are those selected
offline at $\alpha=0.05$ on the first split: the static budget, and \LS{} with 14 and with 4 checkpoints. A request
generates the reasoning up to the next checkpoint of its policy, starts a probe from its own cached prefix, and
either returns the probe answer or resumes, which is the KV-cache fork of Section~\ref{sec:cost-model}. Requests of
all policies are mixed in one queue at a concurrency of 16, so that every policy meets the same load. Two controls
change the cache. In one, requests may also reuse prefixes cached by other requests. In the other the prefix cache
is switched off altogether, so that every probe and every resumption processes its prefix again
(Appendix~\ref{app:latency}).

The second tests a rule that reads the model instead of the probes. \textsc{HiddenStop} applies a logistic
regression, trained on the safe-to-stop label, to the hidden state of the last token at each checkpoint; the layer
and the reduction of the state by principal component analysis are fitted on the training pool
(Appendix~\ref{app:hidden}). Thresholds, calibration and selection are as for \LS. It decides without probing and
makes one probe where it stops, so a stop costs what a static budget of that length costs. Because the rule needs a
self-hosted model, $p>0$ stands here for an engine that processes the prefix again for that probe, not for the
input price of an API.

\section{Results}
\label{sec:results}

Unless stated otherwise, results are for a risk target of $\alpha=0.05$ and a KV-cache fork ($p=0$), so that savings
are shares of generated tokens.

\subsection{RQ1: Learned rule or static budget}
\label{sec:value}

\begin{figure}[t]
\centering
\includegraphics[width=\linewidth]{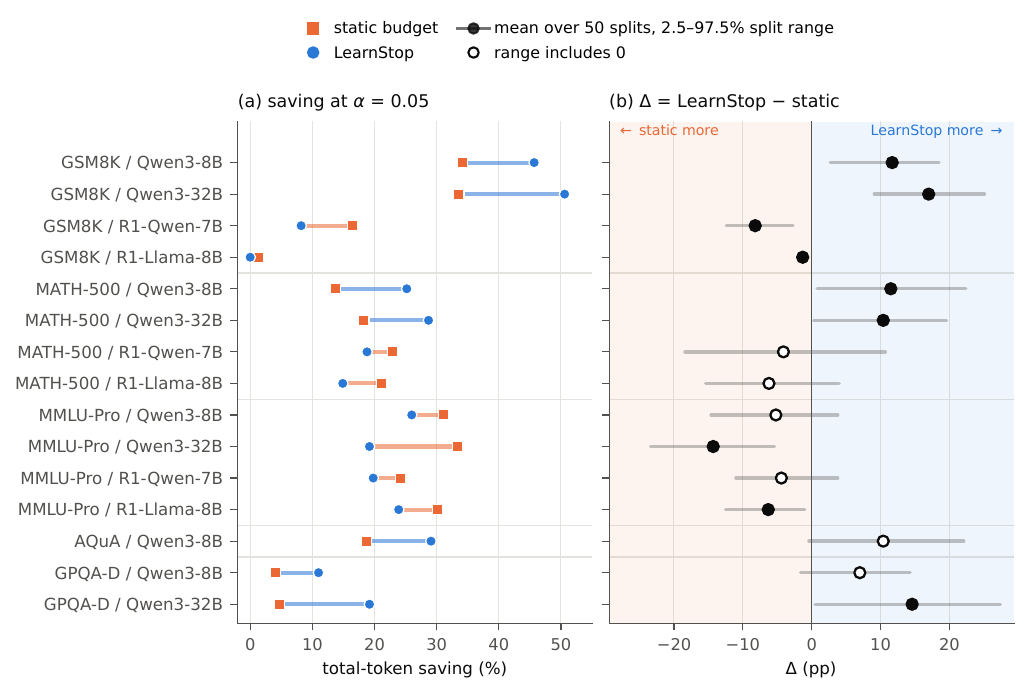}
\caption{Savings at $\alpha=0.05$ with a KV-cache fork. (a) Saving of the static budget and of \LS{} in each setting,
mean over 50 splits; the connecting line is blue where \LS{} saves more and orange where the static budget does.
(b) The increment $\Delta$, \LS{} minus static budget, paired within each split: markers are means over splits, lines
are the 2.5--97.5\% ranges, and filled markers denote ranges that exclude zero.}
\label{fig:core}
\end{figure}

Neither policy saves more in every setting (Table~\ref{tab:main}, Figure~\ref{fig:core}). The learned rule leads
most clearly on the two free-form mathematics tasks with the two Qwen3 models, four settings that we refer to as
Qwen3 math. There its increment over the static budget is between \dQmathAbsMin{} and \dQmathAbsMax{} percentage
points, and the range over splits excludes zero in all four settings. On GSM8K with Qwen3-32B, for example, the
static budget saves \svGsmQttFix\% of the tokens and \LS{} \svGsmQttLS\%. On MMLU-Pro the order is reversed for all
four models. The static budget saves between \dMmluAbsMin{} and \dMmluAbsMax{} points more than \LS, although the
range over splits excludes zero only for \mmluClearList. In the Qwen3 math settings the accuracy of \LS{} is
\dropQmathMin{} to \dropQmathMax{} points below that of full reasoning (Table~\ref{tab:main}).

\begin{table}[t]
\centering
\caption{Savings of each policy at $\alpha=0.05$ and $\delta=0.05$ with a KV-cache fork, in percent of the tokens of
full reasoning; means over 50 splits. Full acc.: accuracy of full reasoning. LS acc.: accuracy of \LS. Scalar: the
four scalar exits certified together, each at $\delta/4$, which can save less than one of them alone. Conv.: the
answer-convergence exit alone. St./LS: \SLS{} (Section~\ref{sec:select}). The last two columns give the increment
$\Delta$ of \LS{} over the static budget: first its mean with its 2.5--97.5\% range over splits (bold when the range
excludes zero), then the share of splits in which it is positive. Bold savings mark the best policy in the row.}
\label{tab:main}
\smallskip
\footnotesize\setlength{\tabcolsep}{3.2pt}
\begin{adjustbox}{max width=\textwidth}\begin{tabular}{lrrrrrrrrlr}
\toprule
 & full & \multicolumn{6}{c}{total-token saving (\%)} & LS & \multicolumn{2}{c}{LearnStop $-$ static budget (pp)} \\
\cmidrule(lr){3-8}\cmidrule(lr){10-11}
Task / model & acc. & static & scalar & conv. & LS & LS+cap & st./LS & acc. & mean [2.5\%, 97.5\%] & win \% \\
\midrule
GSM8K / Qwen3-8B & .960 & 34.1 & 19.8 & 20.7 & \textbf{45.7} & 42.3 & 44.1 & .932 & \textbf{+11.7 [+2.7, +18.5]} & 100 \\
GSM8K / Qwen3-32B & .968 & 33.6 & 38.5 & 23.8 & \textbf{50.6} & 49.8 & 49.8 & .939 & \textbf{+17.0 [+9.1, +25.1]} & 100 \\
GSM8K / R1-Qwen-7B & .830 & \textbf{16.5} & 3.1 & 0.7 & 8.2 & 11.1 & 15.4 & .800 & \textbf{-8.2 [-12.4, -2.7]} & 0 \\
GSM8K / R1-Llama-8B & .658 & \textbf{1.3} & 0.0 & 0.0 & 0.0 & 0.0 & \textbf{1.3} & .658 & \textbf{-1.3 [-1.7, -1.0]} & 0 \\
\midrule
MATH-500 / Qwen3-8B & .922 & 13.7 & 8.1 & 9.6 & \textbf{25.2} & 16.7 & 21.5 & .897 & \textbf{+11.5 [+0.8, +22.4]} & 98 \\
MATH-500 / Qwen3-32B & .938 & 18.3 & 10.2 & 13.7 & \textbf{28.7} & 26.2 & 26.4 & .914 & \textbf{+10.4 [+0.3, +19.6]} & 98 \\
MATH-500 / R1-Qwen-7B & .885 & \textbf{22.9} & 9.7 & 7.1 & 18.8 & 18.2 & 19.8 & .862 & -4.1 [-18.4, +10.7] & 28 \\
MATH-500 / R1-Llama-8B & .730 & \textbf{21.2} & 9.4 & 8.1 & 14.9 & 10.5 & 18.9 & .709 & -6.2 [-15.4, +4.0] & 18 \\
\midrule
MMLU-Pro / Qwen3-8B & .757 & \textbf{31.1} & 16.6 & 17.3 & 26.0 & 27.2 & 30.1 & .736 & -5.2 [-14.6, +3.8] & 16 \\
MMLU-Pro / Qwen3-32B & .823 & \textbf{33.4} & 25.6 & 26.8 & 19.2 & 24.6 & 32.6 & .796 & \textbf{-14.3 [-23.4, -5.4]} & 0 \\
MMLU-Pro / R1-Qwen-7B & .589 & \textbf{24.2} & 12.8 & 13.7 & 19.8 & 18.1 & 23.4 & .573 & -4.4 [-11.0, +3.8] & 14 \\
MMLU-Pro / R1-Llama-8B & .604 & \textbf{30.2} & 16.5 & 18.8 & 23.9 & 22.4 & 29.2 & .587 & \textbf{-6.3 [-12.5, -1.0]} & 2 \\
\midrule
AQuA / Qwen3-8B & .901 & 18.7 & 9.0 & 11.9 & \textbf{29.1} & 23.3 & 26.3 & .879 & +10.4 [-0.4, +22.1] & 96 \\
\midrule
GPQA-D / Qwen3-8B & .617 & 4.0 & 2.1 & 2.2 & \textbf{11.0} & 9.5 & 9.6 & .612 & +7.0 [-1.6, +14.3] & 92 \\
GPQA-D / Qwen3-32B & .630 & 4.7 & 6.1 & 8.8 & \textbf{19.2} & 8.1 & 7.6 & .623 & \textbf{+14.6 [+0.5, +27.4]} & 98 \\
\bottomrule
\end{tabular}\end{adjustbox}
\end{table}

The remaining settings fall on both sides. On AQuA, whose questions are math word problems in multiple-choice form,
and on GPQA-D the learned rule saves more on average: $\Delta$ is \dAqua{} points on AQuA, and \dGpqaQe{} and
\dGpqaQtt{} points on GPQA-D with Qwen3-8B and Qwen3-32B. With the two R1 distillations on MATH-500 the static budget
saves more on average, by \dMathRQabs{} and \dMathRLabs{} points, but the ranges include zero. On GSM8K these two
models close the trace after only \thinkGsmRQ{} and \thinkGsmRL{} tokens on average, so there is little reasoning
for any policy to cut. They work out the solution in their response instead, and Section~\ref{sec:ref} shows that
the comparison in these two settings depends on taking the end probe, and not that response, as the answer of full
reasoning.

\paragraph{Rules that stop on a single signal.}
The scalar exits explain why the static budget is the right baseline. When the four are certified as one policy,
they save less than the static budget in \nBSbelowFix{} of the \nSettings{} settings, and the answer-convergence
exit alone does so in \nConvBelowFix. This is not because probing is expensive: with a KV-cache fork, probing at
every checkpoint adds only \overheadMin--\overheadMax\% to the reasoning tokens. The reason lies in the risk target.
A signal that is sometimes confident about a wrong answer, on a question that full reasoning would have answered
correctly, can only meet a 5\% target with a conservative threshold, and such a threshold stops few questions early.
The static budget does not depend on any signal and shortens every long trace.

\paragraph{Where the learned rule leads.}
On Qwen3 math \LS{} saves more than the static budget because it combines signals. Removing the confidence features
from the classifier lowers its saving by \featDropConf{} points on average, and removing the budget features by
\featDropBudget{} points. A single group of features is rarely enough. The best group alone saves \featAloneMax\% on
average over these four settings, less than the \featFixQm\% of the static budget, and \featAloneBeatsText{}
(Appendix~\ref{app:features}).

GPQA-D shows what limits the static budget. Its certified budgets there are long, and it saves only \svGpqaQeFix\%
and \svGpqaQttFix\% with Qwen3-8B and Qwen3-32B, where \LS{} saves \svGpqaQeLS\% and \svGpqaQttLS\%. Late answers are
the likely reason. Let us say that the answer to a question \emph{settles} at the first checkpoint from which every
later probe is correct. For one in ten of the questions that full reasoning answers correctly on GPQA-D, the answer
settles only after \settleGpqaMin--\settleGpqaMax{} reasoning tokens, compared with
\settleMmluQMin--\settleMmluQMax{} on MMLU-Pro with the same models (Appendix~\ref{app:traj}). One budget for all
questions has to accommodate these late answers, whereas \LS{} can stop other questions earlier.

\paragraph{Where the static budget leads.}
On MMLU-Pro the static budget gains its advantage on the questions that full reasoning answers wrongly. Such
questions cannot be lost, so shortening their traces adds no risk, and the static budget truncates every one of
them that exceeds its length. These questions contribute \stFzMmluMin--\stFzMmluMax{} percentage points to the saving
of the static budget on MMLU-Pro and \lsFzMmluMin--\lsFzMmluMax{} points to that of \LS{} (Table~\ref{tab:label}).
They are also far more common there: \fZeroMmluMin--\fZeroMmluMax\% of the questions, depending on the model,
against \fZeroQmathMin--\fZeroQmathMax\% on Qwen3 math. The signals of \LS{} do not help it here. Removing the
confidence features raises its saving on MMLU-Pro for all four models, and for \nBudgetOnlyMmlu{} models a
classifier that sees only the budget features saves as much as the full one (Appendix~\ref{app:features}).

Neither the subject matter nor the share of wrongly answered questions accounts for the order by itself. Qwen3 math
is mathematics and most of MMLU-Pro is not, but within the mathematics, physics, chemistry and engineering
questions of MMLU-Pro the static budget still saves more for \nCatStatic{} models (Appendix~\ref{app:category}).
GPQA-D has a share of \fZeroGpqaMin--\fZeroGpqaMax\% of such questions and the opposite result, because there the
static budget cannot be short.

\paragraph{Stability across sampled traces.}
Each question has one sampled trace in these runs. We sampled the traces of two settings again with two further
seeds, retrained \LS{} and repeated the certification (Appendix~\ref{app:seeds}). On GSM8K with Qwen3-32B the increment stays between \seedGsmDmin{} and
\seedGsmDmax{} points over the \seedN{} seeds. On MMLU-Pro with Qwen3-8B it ranges from \seedMmluDmin{} to
\seedMmluDmax{} points.
The learned rule leads clearly on GSM8K for every seed. On MMLU-Pro the sign of the small difference
depends on the sample, and the static budget leads for \seedMmluStaticN{} of the \seedN{} seeds.

\paragraph{The two design choices.}
The two design choices of Section~\ref{sec:policies} contribute to these results. The first is the training label.
Table~\ref{tab:label} compares \LS{} with the same classifier trained on the correctness label. With that label the
learned rule saves \labGainQmathMin--\labGainQmathMax{} points less on Qwen3 math. On MMLU-Pro it falls
\labOldMmluMaxAbs--\labOldMmluMinAbs{} points short of the static budget, where \LS{} falls
\dMmluAbsMin--\dMmluAbsMax{} points short. The questions that full reasoning answers wrongly, which are where the two
labels differ, account for \labFzeroShareMin--\labFzeroShareMax\% of the shortfall of the correctness-label rule
relative to the static budget on MMLU-Pro. The safe-to-stop label is not better everywhere: on \labWorseList{} the
correctness label saves more.

\begin{table}[t]
\centering
\caption{Effect of the training label at $\alpha=0.05$ with a KV-cache fork. Saving of the static budget, of \LS{}
(safe-to-stop label) and of the same classifier trained on the correctness label, that is, on whether the probe
answer is correct (p.c.); the increment $\Delta$ of each learned rule over the static budget; the share of questions
that full reasoning answers wrongly (F$=$0); and the part of each policy's saving that is made on those questions,
with the tokens of full reasoning on all questions as the denominator, as for the saving itself. Means over 50 splits.}
\label{tab:label}
\smallskip
\footnotesize\setlength{\tabcolsep}{3.4pt}
\begin{adjustbox}{max width=\textwidth}\begin{tabular}{lrrrrrrrrr}
\toprule
 & \multicolumn{3}{c}{saving (\%)} & \multicolumn{2}{c}{$\Delta$ (pp)} & F$=$0 & \multicolumn{3}{c}{saving on F$=$0 (pp)} \\
\cmidrule(lr){2-4}\cmidrule(lr){5-6}\cmidrule(lr){8-10}
Task / model & static & \LS{} & p.c. & \LS{} & p.c. & (\%) & static & \LS{} & p.c. \\
\midrule
GSM8K / Qwen3-8B & 34.1 & 45.7 & 36.0 & +11.7 & +2.0 & 4 & 7.9 & 8.3 & 6.1 \\
GSM8K / Qwen3-32B & 33.6 & 50.6 & 44.2 & +17.0 & +10.6 & 3 & 7.1 & 6.1 & 3.9 \\
GSM8K / R1-Qwen-7B & 16.5 & 8.2 & 3.5 & -8.2 & -13.0 & 17 & 9.0 & 4.6 & -0.5 \\
GSM8K / R1-Llama-8B & 1.3 & 0.0 & 0.0 & -1.3 & -1.3 & 34 & 0.7 & 0.0 & 0.0 \\
\midrule
MATH-500 / Qwen3-8B & 13.7 & 25.2 & 21.6 & +11.5 & +7.9 & 8 & 4.1 & 2.9 & 2.6 \\
MATH-500 / Qwen3-32B & 18.3 & 28.7 & 23.9 & +10.4 & +5.6 & 6 & 6.4 & 4.9 & 3.8 \\
MATH-500 / R1-Qwen-7B & 22.9 & 18.8 & 19.4 & -4.1 & -3.6 & 12 & 14.4 & 4.2 & 2.0 \\
MATH-500 / R1-Llama-8B & 21.2 & 14.9 & 22.3 & -6.2 & +1.1 & 27 & 13.3 & 6.2 & 6.9 \\
\midrule
MMLU-Pro / Qwen3-8B & 31.1 & 26.0 & 19.9 & -5.2 & -11.3 & 24 & 12.3 & 8.2 & 5.4 \\
MMLU-Pro / Qwen3-32B & 33.4 & 19.2 & 18.5 & -14.3 & -15.0 & 18 & 13.6 & 8.0 & 3.3 \\
MMLU-Pro / R1-Qwen-7B & 24.2 & 19.8 & 12.1 & -4.4 & -12.1 & 41 & 15.1 & 11.9 & 5.3 \\
MMLU-Pro / R1-Llama-8B & 30.2 & 23.9 & 7.6 & -6.3 & -22.6 & 40 & 19.8 & 16.2 & 4.8 \\
\midrule
AQuA / Qwen3-8B & 18.7 & 29.1 & 28.0 & +10.4 & +9.3 & 10 & 8.1 & 7.2 & 5.8 \\
\midrule
GPQA-D / Qwen3-8B & 4.0 & 11.0 & 12.0 & +7.0 & +7.9 & 38 & 1.8 & 5.2 & 5.7 \\
GPQA-D / Qwen3-32B & 4.7 & 19.2 & 3.6 & +14.6 & -1.1 & 37 & 3.0 & 10.8 & 3.0 \\
\bottomrule
\end{tabular}\end{adjustbox}
\end{table}

The second choice is the fine grid of the static budget, and it works in the other direction. Restricted to the 14
checkpoints, the static budget saves \gridGainMean{} points less on average and up to \gridGainMax{} points less, and
the increment on Qwen3 math would appear \gridShrinkQmathMin--\gridShrinkQmathMax{} points larger than it is
(Appendix~\ref{app:grid}).

Certification at the same target leaves the two policies at slightly different realised risks, on average at most \rMaxFix\% for the static budget and \rMaxLS\% for \LS{}
(Appendix~\ref{app:policyfull}). Reading the curve of saving against realised risk of \LS, obtained
by certifying it at a range of targets, at the realised risk of the static budget gives an increment on Qwen3 math
of \dRealQmathMin--\dRealQmathMax{} points (Appendix~\ref{app:realised}). The interpolated point describes the
curve and is not a policy that was deployed.

\subsection{RQ2: Risk target, probe price and probing schedule}
\label{sec:conditions}

\begin{figure}[t]
\centering
\includegraphics[width=\linewidth]{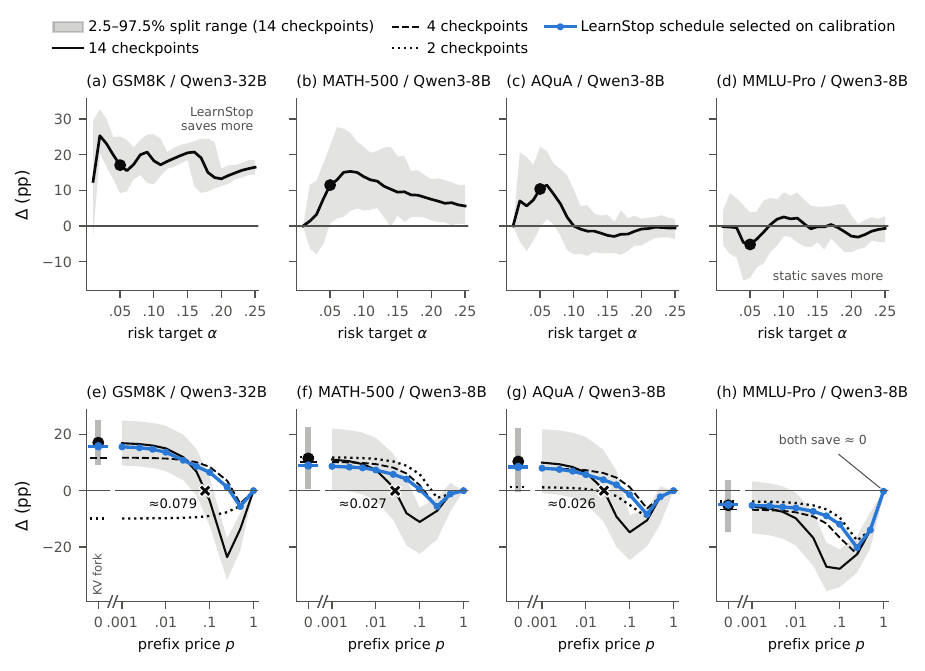}
\caption{The increment $\Delta$ of \LS{} over the static budget in four settings. Top row, (a)--(d): $\Delta$ against
the risk target $\alpha$, with a KV-cache fork and probes at all 14 checkpoints. Bottom row, (e)--(h): $\Delta$
against the price $p$ of a re-sent prefix token at $\alpha=0.05$, for \LS{} probing at 14, 4 or 2 checkpoints and
for the schedule selected on each calibration half among the four of Table~\ref{tab:breakeven}; the narrow panel on the left of each plot is $p=0$, drawn apart
from the logarithmic axis. The static budget always uses its fine grid, and every policy is certified and selected
again at every point. Lines are means over 50 splits and the band is the 2.5--97.5\% range for 14 checkpoints. The
dot marks the operating point of Figure~\ref{fig:core}b and the cross the break-even price of the 14-checkpoint
schedule. At $p=1$ neither policy saves anything in these settings, so $\Delta$ returns to zero. Above the zero line
\LS{} saves more, below it the static budget.}
\label{fig:delta}
\end{figure}

The comparison of Section~\ref{sec:value} holds at one operating point. In \nSignChange{} of the
\nSettings{} settings the policy that saves more changes at least once as $\alpha$ rises from 0.05 to 0.20, and on
Qwen3 math the lead of \LS{} disappears at prices of a few percent of an output token. Figure~\ref{fig:delta} follows
$\Delta$ in four representative settings, and Appendix~\ref{app:alpha} gives every setting.

\paragraph{Risk target.}
The settings fall into three groups as $\alpha$ rises from 0.05 to 0.20. In \nAlphaLS{} settings, \alphaLSList,
\LS{} saves more at all four targets, by at least \alphaLSMinTwenty{} points at $\alpha=0.20$. In \nAlphaFix{}
settings, \alphaFixList, the static budget saves more throughout. In the other \nSignChange{} settings the sign of
$\Delta$ changes at least once. In \nSwing{} of them the change is large, between \swingMin{} and \swingMax{} points
from the smallest to the largest value of $\Delta$: for example \swingExText.

Which way the sign changes depends on which policy gains more from a looser target. On \flipToStaticList{} the
saving of the static budget grows by \flipGainFix{} points on average and that of \LS{} by \flipGainLS; on
\flipToLSList{} the saving of \LS{} grows by \flipLSGainLS{} points and that of the static budget by
\flipLSGainFix. An increment measured at one risk target therefore says little about the choice at another.

\paragraph{Probe price.}
When the prefix has to be sent again, a policy that probes at 14 checkpoints pays for up to 14 probes and as many
resumptions, and the static budget for one probe. On Qwen3 math the increment of the 14-checkpoint schedule
accordingly falls as $p$ rises and reaches zero at a break-even price between \beQmathLo{} and \beQmathHi{}
(Figure~\ref{fig:delta}e,f and Table~\ref{tab:breakeven}). The prefix-cache price of 0.025 lies just below
this range, so dense probing still pays there, narrowly on MATH-500, and the black-box price of 0.25 lies far above
it. Dense probing thus pays with a KV-cache fork or a
steeply discounted prefix cache and usually does not pay through an API that bills the prefix as input. On
MMLU-Pro, where the static budget leads already at $p=0$, a price on the prefix only puts dense probing at a
further disadvantage.

\begin{table}[t]
\centering
\caption{Break-even price: the price of a re-sent prefix token, relative to an output token, at which \LS{} loses
its advantage over the static budget at $\alpha=0.05$, interpolated between the two tested prices around the change
of sign (on a logarithmic scale, and linearly from $p=0$). A price at which both savings are below \zeroTol{} points
counts as a tie at zero and not as a crossing. Columns are
checkpoint schedules: 14 is the full set; 7 is 0, 256, 512, 1{,}024, 2{,}048, 4{,}096, 8{,}192; 4 is 0, 512, 2{,}048,
8{,}192; 2 is 1{,}024, 4{,}096. Last column: the schedule is chosen on each calibration half among the four, which
are certified as families of one policy at $\delta/4$ each. A dash means that \LS{} saves less than the static
budget even at $p=0$. The static budget always uses its fine grid.}
\label{tab:breakeven}
\smallskip
\small
\begin{adjustbox}{max width=\textwidth}\begin{tabular}{lrrrrr}
\toprule
 & \multicolumn{4}{c}{checkpoints in schedule} & chosen on \\
\cmidrule(lr){2-5}
Task / model & 14 & 7 & 4 & 2 & calibration \\
\midrule
GSM8K / Qwen3-8B & 0.043 & 0.078 & -- & 0.77 & 0.30 \\
GSM8K / Qwen3-32B & 0.079 & 0.13 & 0.34 & -- & 0.29 \\
GSM8K / R1-Qwen-7B & -- & -- & -- & -- & -- \\
GSM8K / R1-Llama-8B & -- & 0.013 & 0.036 & -- & 0.027 \\
\midrule
MATH-500 / Qwen3-8B & 0.027 & 0.070 & 0.13 & 0.19 & 0.11 \\
MATH-500 / Qwen3-32B & 0.026 & 0.037 & 0.059 & 0.22 & 0.12 \\
MATH-500 / R1-Qwen-7B & -- & -- & -- & -- & -- \\
MATH-500 / R1-Llama-8B & -- & -- & -- & -- & -- \\
\midrule
MMLU-Pro / Qwen3-8B & -- & -- & -- & -- & -- \\
MMLU-Pro / Qwen3-32B & -- & -- & -- & -- & -- \\
MMLU-Pro / R1-Qwen-7B & -- & -- & -- & -- & -- \\
MMLU-Pro / R1-Llama-8B & -- & -- & -- & 0.006 & -- \\
\midrule
AQuA / Qwen3-8B & 0.026 & 0.028 & 0.12 & 0.018 & 0.075 \\
\midrule
GPQA-D / Qwen3-8B & 0.014 & 0.024 & 0.053 & 0.039 & 0.027 \\
GPQA-D / Qwen3-32B & 0.038 & 0.043 & 0.064 & 0.22 & 0.099 \\
\bottomrule
\end{tabular}\end{adjustbox}
\end{table}

\paragraph{How often to probe.}
A sparser schedule trades chances to stop early for a lower bill. With fewer checkpoints \LS{} can save much
less at $p=0$ (on GSM8K with Qwen3-32B, \svGsmQttSTwo\% with two checkpoints against \svGsmQttLS\% with 14), but its saving falls more slowly as $p$ rises (bottom row of Figure~\ref{fig:delta};
Appendix~\ref{app:breakeven} shows the savings themselves). With four checkpoints the break-even price on GSM8K
with Qwen3-32B is \beGsmQttBest, and with two checkpoints on GSM8K with Qwen3-8B it is \beGsmQeTwo, several times
the break-even price of the 14-checkpoint schedule in the same settings (Table~\ref{tab:breakeven}).

An operator has to choose the schedule without seeing test results, and the selection step can make this choice.
The four schedules are entered as four families of one policy, each certified at $\delta/4$, and the cheapest
certified candidate is deployed. Chosen in this way, \LS{} keeps a positive increment at $p=0.025$ in \nBEselPC{}
settings and at $p=0.25$ in \nBEselBB, as many as when the best schedule is picked with knowledge of the test
results. The choice has a price, which shows at $p=0$: there the selected schedule changes the saving on Qwen3 math
by \selVsFourteenMin{} to \selVsFourteenMax{} points relative to the 14-checkpoint schedule, because $\delta$ is
divided four ways and the choice is made on a finite sample.

\paragraph{Counting probe tokens.}
The comparison also depends on how many tokens a probe is charged. With a stop at the first line break, a probe
decodes \probeLineMin--\probeLineMax{} tokens on average. Without it, some models continue to explain their answer,
and a probe decodes up to \probeFullMax{} tokens on average (Table~\ref{tab:settings}). Charging every probe the 48-token decoding cap instead of its measured
length reverses the result on GSM8K with Qwen3-8B: \LS{} saves \lineGsmQeLS\% against \lineGsmQeFix\% for the static
budget when probes are charged what they decode up to the line break, and \capGsmQeLS\% against \capGsmQeFix\%
under the flat charge (Appendix~\ref{app:accounting}). A deployment should stop its probes at the first line, and an evaluation
should count the tokens that they actually decode.

\subsection{RQ3: Selecting the family on calibration data}
\label{sec:select}

\begin{figure}[t]
\centering
\includegraphics[width=\linewidth]{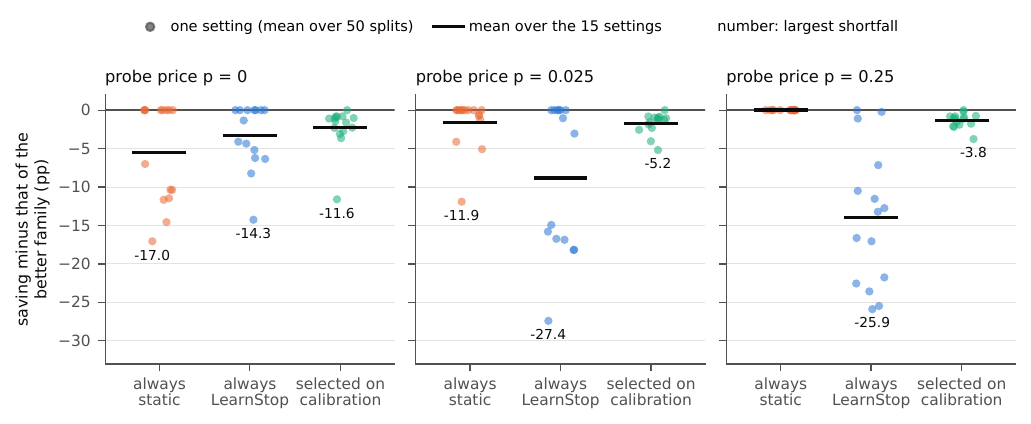}
\caption{Selecting the family on calibration data against committing to one family, at $\alpha=0.05$ and three prices
$p$ of a re-sent prefix token ($p=0$: KV-cache fork). Each dot is one of the \nSettings{} settings: the saving of the
policy minus that of the family, static budget or \LS{} with 14 checkpoints, that saves more in that setting, both
means over 50 splits; zero means that the policy matches the better family. Bars: means over the settings. Numbers:
the largest shortfall. Selected on calibration: \SLS. At $p=0$ its largest shortfall is on GPQA-D with Qwen3-32B,
which has \gpqaNcal{} calibration questions.}
\label{fig:select}
\end{figure}

Since the better family differs from one workload to the next, the procedure does not fix it in advance. \SLS{}
certifies the static budgets and the \LS{} thresholds as two families, each at level $\delta/2$, and deploys the
cheapest certified candidate. We compare it with the two alternatives that an operator has without selection: using
the static budget in every setting, or using \LS{} in every setting. Throughout this subsection \LS{} probes at all
14 checkpoints; the choice of schedule was treated in Section~\ref{sec:conditions}. Figure~\ref{fig:select} shows what each of the
three gives up in a setting relative to the family that saves more there, at the three prices of
Section~\ref{sec:cost-model}. Which family that is can be known only from the test results, so it serves as a
reference and is not a choice that an operator could make.

Committing in advance fails badly somewhere, whichever family is chosen. With a KV-cache fork, the static budget
used everywhere falls up to \wcFixZero{} points short of the better family, and \LS{} used everywhere up to
\wcLSZero{} points. Once a re-sent prefix is billed, the larger shortfalls occur with \LS. The certified thresholds
of its 14-checkpoint schedule are then often no cheaper than full reasoning, so the policy falls back to full
reasoning and saves nothing. Used everywhere, it falls up to \wcLSPC{} points short at $p=0.025$ and \wcLSBB{} at
$p=0.25$, where the static budget is the better family in every setting.

\SLS{} keeps the shortfall small, except in one setting with a small calibration set, which we examine below. It
falls at most \wcSelPC{} and \wcSelBB{} points short at the two higher prices, and at $p=0$ at most \wcSelZeroEx{}
points in every other setting. Avoiding the bad cases does not lower its average saving. At $p=0$ it saves
\meanSelZero\% on average over the settings, against \meanFixZero\% for the static budget used everywhere and
\meanLSZero\% for \LS{} used everywhere, and at the two higher prices it stays within \uniGapPC{} and \uniGapBB{}
points of the static budget. An operator who selects therefore does not need to know in advance which family the
workload favours.

The shortfall that remains has two causes. Each family is tested at $\delta/2$ instead of $\delta$, which makes
certification harder, and the choice between the families is made on a finite calibration sample. At $p=0$ the
shortfall is \slGapMean{} points on average. The cost of the stricter test is most visible in the one large value,
on \slGapWorst, which has only \gpqaNcal{} calibration questions. With so few, \gpqaLossText. The thresholds that
\LS{} would deploy at level $\delta$ have one loss in most splits and therefore fail at $\delta/2$. \SLS{} then has
to fall back to more conservative thresholds or to the static budget, and it saves \svSLGpqaQtt\% where \LS{} alone
saves \svLSGpqaQtt\%. The risk guarantee is not affected in any of these cases; what is lost is saving.

Selection also does better than building the budget into the learned rule. \LSC{} saves \capGapMean{} points less
than the better family on average and up to \capGapMax{} points less. Two restrictions account for this: its static
budgets are limited to the five cap lengths, and each of its six families is certified at $\delta/6$.
Appendix~\ref{app:decomp} measures both on the static budget alone.

The same selection can include further families, at the price of dividing $\delta$ further.
Section~\ref{sec:conditions} used it to choose the checkpoint schedule, and a service with several similar tasks can
enter a classifier trained on their pooled questions, with thresholds still calibrated per task
(Appendix~\ref{app:learners}).

\subsection{RQ4: Labelled data and when it is repaid}
\label{sec:data}

The learned rule needs something that the static budget does not: a pool of labelled questions on which to train
the classifier. How large this pool has to be depends on the task. Table~\ref{tab:train} retrains \LS{} on random
subsets of 50, 100 and 200 training questions, five subsets of each size, and certifies it as before. On GSM8K with
Qwen3-32B a pool of 50 questions already gives an increment of \trainGsmQttFifty{} points, against \trainGsmQttAll{}
with the full pool. On MATH-500 with Qwen3-8B the same 50 questions give \trainMathQeFifty{} points against
\trainMathQeAll. Among the \nTrainWinners{} settings in which \LS{} leads with the full pool, 50 questions retain at
least three quarters of the increment in \trainMostN{} and less than half in \trainLessN. In the settings where the
static budget leads, \trainNoFlipText.

\begin{table}[t]
\centering
\caption{Increment $\Delta$ at $\alpha=0.05$ with a KV-cache fork when \LS{} is trained on a random subset of $n$
questions from the training pool; means over five subsets and 50 splits. All: the full pool, 348 questions for
GPQA-D and 500 otherwise. Tokens per question: tokens decoded to generate one training trace and its probes, in
thousands. Requests to repay: the tokens spent on collecting a pool of the given size, divided by the tokens that \LS{}
saves per request relative to the static budget when trained on that pool. Settings in which \LS{} saves less than the static budget with the
full pool are omitted.}
\label{tab:train}
\smallskip
\small\setlength{\tabcolsep}{3.6pt}
\begin{adjustbox}{max width=\textwidth}\begin{tabular}{lrrrrrrr}
\toprule
 & \multicolumn{4}{c}{difference (pp) with $n$ training questions} & tokens per & \multicolumn{2}{c}{requests to repay} \\
\cmidrule(lr){2-5}\cmidrule(lr){7-8}
Task / model & 50 & 100 & 200 & all & question (k) & $n=50$ & all \\
\midrule
GSM8K / Qwen3-8B & +6.6 & +9.1 & +9.2 & +11.7 & 1.4 & 750 & 4,150 \\
GSM8K / Qwen3-32B & +15.0 & +17.1 & +16.0 & +17.0 & 0.9 & 330 & 2,880 \\
MATH-500 / Qwen3-8B & +4.3 & +8.2 & +11.9 & +11.5 & 4.7 & 1,330 & 4,740 \\
MATH-500 / Qwen3-32B & +8.0 & +11.6 & +9.5 & +10.4 & 3.9 & 710 & 5,210 \\
AQuA / Qwen3-8B & +3.8 & +7.7 & +11.6 & +10.4 & 2.6 & 1,350 & 4,810 \\
GPQA-D / Qwen3-8B & +3.3 & +5.7 & +6.9 & +7.0 & 6.3 & 1,480 & 4,570 \\
GPQA-D / Qwen3-32B & +5.3 & +5.4 & +6.4 & +14.6 & 4.5 & 920 & 2,170 \\
\bottomrule
\end{tabular}\end{adjustbox}
\end{table}

Both policies need calibration questions, and their number matters as well. When 30\% of the evaluation set is
used for calibration instead of 50\%, both policies save less in most settings, because fewer candidates can be
certified (Appendix~\ref{app:calfrac}).

The increment $\Delta$ already accounts for the probes made while serving, but not for the one-off cost of
generating the training traces and their probes. We count this cost in decoded tokens and divide it by the tokens
that \LS{} saves per request relative to the static budget, which gives the number of requests after which the
collection is repaid. In the settings where \LS{} leads, this is \paybackAllMin--\paybackAllMax{} requests for the
full pool and \paybackFiftyMin--\paybackFiftyMax{} for a pool of 50 questions (Table~\ref{tab:train}). If the traces
would be generated anyway and only the probes are an extra cost, the full pool is repaid after
\paybackProbeMin--\paybackProbeMax{} requests. These figures count decoded tokens only. They leave out the calibration records, which both policies need, and the
cost of obtaining the correct answers, which for a workload that has to be labelled by hand can be the larger cost.

\subsection{RQ5: From token savings to response time}
\label{sec:latency}

\begin{figure}[t]
\centering
\includegraphics[width=0.86\linewidth]{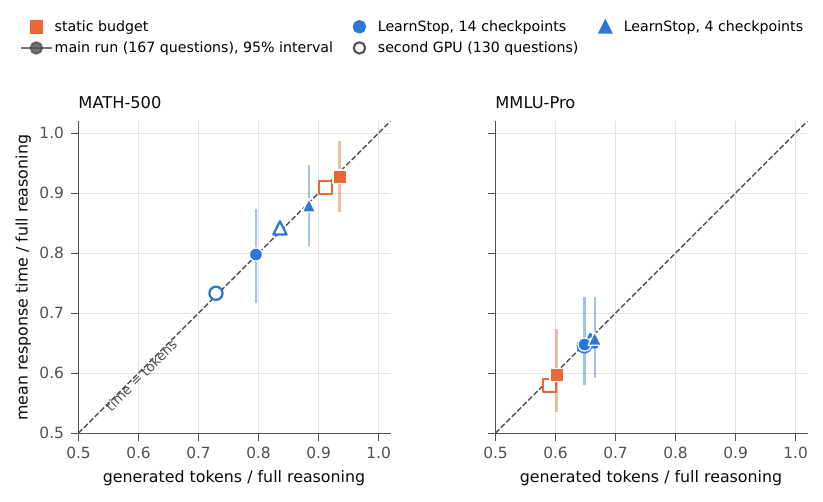}
\caption{Mean response time against generated tokens in a serving engine (Qwen3-8B, vLLM, one H100, concurrency 16),
both relative to full reasoning on the same questions. Filled markers: main run, with a 95\% question-level bootstrap
interval for the time; open markers: a second run at the same time on another GPU. On the dashed line the time per
generated token equals that of full reasoning. The policies were certified offline at $\alpha=0.05$ on one split.}
\label{fig:latency}
\end{figure}

In the serving engine the mean response time fell in proportion to the tokens generated (Figure~\ref{fig:latency}).
For every policy the mean time per generated token stayed between \latPerTokMin{} and \latPerTokMax{} of that of
full reasoning, on \latNMain{} questions per task at a concurrency of 16. On MATH-500, for example, \LS{} with 14
checkpoints took \latMathLSd{} of the time of full reasoning (95\% interval over questions
\latMathLSdLo--\latMathLSdHi) and the static budget \latMathFix; on MMLU-Pro the static budget took \latMmluFix{}
and \LS{} \latMmluLSd. The order by mean response time
is therefore the order by tokens, which is the order of Section~\ref{sec:value}: the learned rule leads on
mathematics and the static budget on MMLU-Pro. A second run on another GPU of the same type, with \latNRep{} questions per task, gave the same order
(Appendix~\ref{app:latency}).

Probes added little time beyond their own tokens. \LS{} with 14 checkpoints made \latProbesMath{} of them per
question on MATH-500 and \latProbesMmlu{} on MMLU-Pro, each started from the cached reasoning, and its time per
generated token stayed within the range above; this is the KV-cache fork that $p=0$ describes. The two families
shortened different requests, however. The static budget cuts only the long traces, so it shortened the slowest
requests most: on MMLU-Pro the 90th percentile of the response time was \latPninetyMmluFix{} s for the static budget,
\latPninetyMmluLSd{} s for \LS{} and \latPninetyMmluFull{} s for full reasoning.

Online, the policies behaved in aggregate as the records predict, though not request by request. The traces are
sampled again, so a question can stop at another checkpoint than in the records, and \LS{} with 14 checkpoints
stopped where the offline analysis predicts in only \agreeLSdMin--\agreeLSdMax\% of the requests. In aggregate,
however, the share of questions stopped early differed from the offline prediction by at most \earlyGapMax{}
percentage points, and the token ratios were within \tokOffGap{} of the predicted ones. The online run does not
test the risk bound. The bound rests on the offline certification, which covers newly sampled traces as long as the
workload and the generation settings are unchanged. Each policy ran on its own sampled traces, and the
differences in accuracy between policies are within the sampling error of about \latAccNoise{} points at this size
(Table~\ref{tab:latency}).

Two controls with \latNCtl{} questions per task varied the cache (Appendix~\ref{app:latency}). In the main run a
request reuses only its own cached prefix. Letting requests also reuse prefixes cached by other requests changed
little: the time per token stayed between \latCtlPerTokMin{} and \latCtlPerTokMax{} of that of full reasoning.
Switching the prefix cache off, so that every probe and every resumption processes the reasoning again, raised the
time per generated token of \LS{} with 14 checkpoints only to \perTokNocMathLSd{} of that of full reasoning on
MATH-500 and \perTokNocMmluLSd{} on MMLU-Pro. For Qwen3-8B on these two tasks at a concurrency of 16, frequent
probing thus added little time beyond its decoded tokens even without a prefix cache, because processing a prefix
is much faster than generating it. The prices $p>0$ of Section~\ref{sec:conditions} therefore matter where re-sent
input is billed, as through an API, and much less for the response time of a self-hosted engine.

\subsection{RQ6: A change of workload}
\label{sec:shift}

\begin{figure}[t]
\centering
\includegraphics[width=\linewidth]{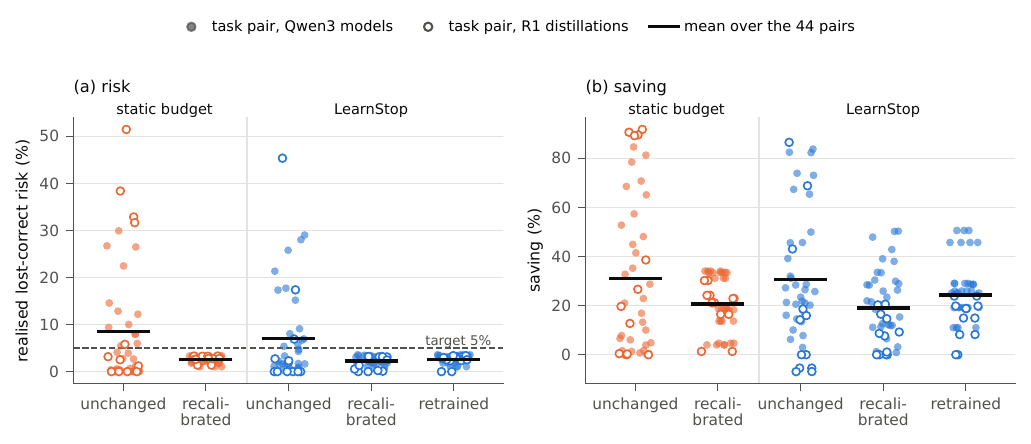}
\caption{Policies certified at $\alpha=0.05$ on one task and served on another task of the same model: the
\shiftPairs{} ordered pairs of the four models, KV-cache fork. (a) Realised lost-correct risk and (b) saving, both measured on the
held-out test halves of the served task; means over 50 splits per pair. Unchanged: the policy certified on the calibration task; recalibrated:
certification and selection repeated on a calibration half of the served task, \LS{} keeping its classifier;
retrained: \LS{} also retrained on the training pool of the served task. Filled dots: pairs with a Qwen3 model; open
dots: pairs with an R1 distillation; bars: means over all pairs.}
\label{fig:recover}
\end{figure}

The guarantee of Proposition~\ref{prop:ltt} assumes that the served questions come from the distribution of the
calibration questions. To see how the policies behave when this fails, we move them between tasks. A policy is
certified and selected on the calibration half of one task exactly as before, with \LS{} trained on the training
pool of that task. The selected budget, or the selected classifier with its threshold, is then applied without
change to another task run with the same model, and its risk is measured against the full-reasoning answers of that
task. The transfer alone is measured on all evaluation questions of the served task. Where it is compared with
recalibration, which takes a calibration half of the served task, every policy is measured on the held-out test
halves instead (Figure~\ref{fig:recover}). No question is used both to calibrate a policy and to measure it. With
four models this gives \shiftPairs{} ordered pairs of a
calibration task and a served task.

On the task where a policy was calibrated, its mean realised risk stays below the 5\% target for every model: it is
at most \shiftDiagMaxFix\% for the static budget and \shiftDiagMaxLS\% for \LS. On a different task the target is
exceeded in \shiftViolFix{} of the \shiftPairs{} pairs for the static budget and in \shiftViolLS{} pairs for \LS,
and the risk reaches \shiftMaxFix\% and \shiftMaxLS\% (Appendix~\ref{app:shift} gives every pair).

The direction of the change decides the outcome. A policy calibrated on a task with short traces is too aggressive
for a task with long ones. In the \shiftUpNFix{} pairs where the calibration task has the shorter mean trace, the
target is exceeded \shiftUpViolFix{} times by the static budget and \shiftUpViolLS{} times by \LS. Transferred from
GSM8K to MATH-500 with Qwen3-8B, for example, the static budget loses \shiftExFix\% of the questions and \LS{}
\shiftExLS\%. In the opposite direction the policies are too
cautious. The target is exceeded in only \shiftDownViolFix{} and \shiftDownViolLS{} of the \shiftDownNFix{} pairs,
but the static budget saves \shiftDownSaveFix\% on average where a budget calibrated on the served task would save
\shiftDownSaveOwnFix\%, and \LS{} saves \shiftDownSaveLS\% instead of \shiftDownSaveOwnLS\%. The features of \LS{}
respond to each question, but this does not make it more robust than the static budget: the two exceed the target
in the same number of pairs.

Recalibration on the new task restores risk control for both families (Figure~\ref{fig:recover}). The certification
and selection are repeated on a calibration half of the served task, with the classifier of \LS{} left as it was
trained on the old task, and risk and saving are then measured on the held-out test half of the served task. The figure shows the
unchanged policies on the same test halves. The
mean risk exceeds the target in none of the \shiftPairs{} pairs. It is at most \recalFixMax\% for the static budget
and \recalLSMax\% for \LS. The recalibrated static budget also recovers its saving: \recalFixSave\% on average over
the pairs, which is what a budget calibrated on the served task from the outset saves.

For \LS{} the saving after recalibration depends on the model, as it did on the calibrated tasks. On average over
the \recalQPairs{} pairs with a Qwen3 model, the old classifier still saves more than the static budget, \recalQLS\%
against \recalQFix\%, and retraining it on the training pool of the new task raises its saving to \retrainQLS\%. On
average over the \recalRPairs{} pairs with an R1 distillation it saves less, \recalRLS\% against \recalRFix\%, and
\retrainRLS\% when retrained, still below the static budget. Two decisions follow a change of workload. Which family to deploy is
decided as before, by certifying both on the new calibration set. Whether to collect new training data is a
separate question, which compares the saving that retraining adds with the cost of the data, as in
Section~\ref{sec:data}.

Recalibration needs a minimum of data. A candidate that loses no calibration question passes the test at
$\alpha=0.05$ and $\delta=0.05$ only if there are at least \recalMinN{} of them, and at least \recalMinNTwo{} when
$\delta$ is shared between two families, as in selection. With 25 or 50 calibration questions no early-exit candidate
is certified, and the procedure falls back to full reasoning. With 200 questions of the served task the recalibrated
static budget saves \recalSizeTwoHundredFix\% on average, against \recalSizeFullFix\% with the full calibration
half, and \LS{} with the old classifier \recalSizeTwoHundredLS\% against \recalSizeFullLS\% (averages over the
\recalSizePairs{} pairs whose served task has at least 200 calibration questions; Appendix~\ref{app:shift} gives the
other sizes).

\section{Comparisons and robustness checks}
\label{sec:robust}

The results above compare the two families of the procedure under one definition of the full-reasoning answer and
one certification test. This section places the learned rule next to two alternatives outside the procedure, a rule
that reads hidden states and a model fine-tuned to reason less, and then varies the reference and the test.

\subsection{A rule that reads hidden states}
\label{sec:hidden}

A learned rule can read signals other than the provisional answers. \textsc{HiddenStop}, a linear probe on the hidden
state of one layer (Section~\ref{sec:hiddensetup}), is a simple rule of this kind. Table~\ref{tab:hidden} sets it
beside the static budget and \LS{} on the same splits. At $p=0$ it saves \hidMathHSZero\% on
MATH-500, about as much as the static budget (\hidMathFixZero\%) and less than \LS{} (\hidMathLSZero\%), and
\hidMmluHSZero\% on MMLU-Pro, less than both (\hidMmluFixZero\% and \hidMmluLSZero\%). Its realised risk stays below
the target, at \hidRiskMath\% and \hidRiskMmlu\%.

With a KV-cache fork, answer probes add little cost, so the comparison reflects the quality of the signal, and
the hidden-state classifier tested here discriminates weakly. The layer chosen on the training pool reaches a
cross-validated AUC of only \hidAucMath{} on MATH-500 and \hidAucMmlu{} on MMLU-Pro, and other layers, added budget
features or thresholds from out-of-fold scores do not change the picture (Appendix~\ref{app:hidden}). On MATH-500
the lead of \LS{} therefore comes from the features of the probe answers, which this classifier does not match.

At $p=0.25$ the order changes. A rule that reads hidden states cannot run behind an API, so for this comparison $p=0.25$ is an
accounting weight for processing the reasoning again on a self-hosted engine. It is not a measured time ratio, which
Section~\ref{sec:latency} found to be much smaller. Under this accounting the
repeated probes of \LS{} with 14 checkpoints make every certified threshold more expensive than full reasoning, and
\LS{} falls back to full reasoning. \textsc{HiddenStop} makes a single answer probe, when it stops, and still saves
tokens: more than the static budget on MATH-500 (\hidMathHSHigh\% against \hidMathFixHigh\%) and less on MMLU-Pro
(Table~\ref{tab:hidden}). Which rule is worth deploying thus depends on the signal and on how it is obtained, as
well as on the workload and the price. The comparison covers one linear classifier on one model and two tasks, and a stronger reader of hidden states
may do better.

\begin{table}[t]
\centering
\caption{A stopping rule on hidden states (Qwen3-8B, $\alpha=0.05$, 50 splits): saving (\%) of the static budget,
\LS{} with 14 checkpoints and \textsc{HiddenStop}, and the paired difference of \textsc{HiddenStop} to each (mean and
2.5--97.5\% range over splits, percentage points). At $p=0.25$, ``full'': no certified threshold of \LS{} is cheaper
than full reasoning, so the policy falls back to full reasoning and saves nothing.}
\label{tab:hidden}
\smallskip
\footnotesize\setlength{\tabcolsep}{3.4pt}
\begin{adjustbox}{max width=\textwidth}\begin{tabular}{lrrrrrrrr}
\toprule
 & \multicolumn{5}{c}{$p=0$} & \multicolumn{3}{c}{$p=0.25$} \\
\cmidrule(lr){2-6}\cmidrule(lr){7-9}
Qwen3-8B & static & \LS & \textsc{HiddenStop} & $-$ static & $-$ \LS & static & \LS & \textsc{HiddenStop} \\
\midrule
MATH-500 & 13.7 & 25.2 & 14.1 & +0.4 [$-$11.5, +8.3] & $-$11.1 [$-$21.9, +1.3] & 7.1 & 0 (full) & 9.5 \\
MMLU-Pro & 31.1 & 26.0 & 18.9 & $-$12.2 [$-$23.8, +1.3] & $-$7.0 [$-$16.4, +2.9] & 22.5 & 0 (full) & 14.7 \\
\bottomrule
\end{tabular}\end{adjustbox}
\end{table}

\subsection{A model fine-tuned to reason less}
\label{sec:adaptthink}

All policies above leave the model as it is. The alternative is to change the model so that it reasons less.
AdaptThink-7B \citep{zhang2025adaptthink} is such a model, fine-tuned from R1-Qwen-7B (we use the released variant
with its own hyperparameter $\delta=0.05$).
Table~\ref{tab:adaptthink} compares it with the certified policies applied to that base model. Because AdaptThink
often skips the reasoning and answers in its response, this comparison takes the model's own response as the
answer of full reasoning, for both models, and measures savings in all generated tokens. A policy that stops early
still returns its probe answer. A policy that does not stop returns the response of the base model, and the tokens
of that response count towards its cost.

On the two mathematics tasks fine-tuning saves much more. AdaptThink uses \atGsmsave\% fewer tokens than its base
model on GSM8K and \atMathsave\% fewer on MATH-500, with accuracy equal to or above that of the base model and with
fewer than 5\% of the questions lost relative to it (Table~\ref{tab:adaptthink}). The best certified policies on the
base model save less: \atBestText. On MMLU-Pro the result is the other way round. AdaptThink saves less than the
best certified policy, and on \atMmluLost\% of the questions the base model is right and AdaptThink is wrong, which
is above the 5\% to which the certified policies are held. On the two mathematics tasks fine-tuning is thus the
stronger option where it is available. Stopping at inference time applies to any model without retraining, and it
is the option that comes with a bound on the risk (Appendix~\ref{app:adaptthink} extends the comparison to the other
risk targets).

\begin{table}[t]
\centering
\caption{AdaptThink-7B and certified policies on its base model, R1-Qwen-7B. Accuracy is that of the model's own
responses, and savings are in all generated tokens relative to the responses of the base model. Lost-correct: share
of all questions on which the base model is right and AdaptThink is wrong. The policies are certified at
$\alpha=0.05$ on the calibration halves, against the correctness of the base model's responses, and all methods are
compared on the same test halves.}
\label{tab:adaptthink}
\smallskip
\small\setlength{\tabcolsep}{3.2pt}
\begin{adjustbox}{max width=\textwidth}\begin{tabular}{lrrrrrrrrrr}
\toprule
 & base & \multicolumn{3}{c}{AdaptThink-7B (not certified)} & \multicolumn{6}{c}{R1-Qwen-7B + exit certified at $\alpha=0.05$} \\
\cmidrule(lr){3-5}\cmidrule(lr){6-11}
 & & & & lost- & \multicolumn{2}{c}{static} & \multicolumn{2}{c}{LS} & \multicolumn{2}{c}{st./LS} \\
Task & acc. & acc. & saving & correct & acc. & saving & acc. & saving & acc. & saving \\
\midrule
GSM8K & .890 & .904 & 36.4 & 0.045 & .866 & 9.4 & .860 & 19.6 & .863 & 18.5 \\
MATH-500 & .890 & .899 & 56.3 & 0.034 & .868 & 24.9 & .866 & 22.8 & .866 & 23.9 \\
MMLU-Pro & .577 & .567 & 19.6 & 0.095 & .559 & 28.5 & .558 & 19.3 & .560 & 27.9 \\
\bottomrule
\end{tabular}\end{adjustbox}
\end{table}

\subsection{The definition of the full-reasoning answer}
\label{sec:ref}

Apart from the comparison with AdaptThink, all results above take the end probe as the answer of full reasoning
(Section~\ref{sec:probe}). This isolates the
reasoning phase, because early and final answers are produced in the same way. The alternative is to take the
response that the model writes by itself after the trace. Under this definition full reasoning costs the trace plus
the response, and a policy that stops early and answers with a short probe saves the response as well. We repeated
the main comparison with this reference (Appendix~\ref{app:natref}). Everything that depends on the reference is
recomputed: which questions full reasoning answers correctly, and with it the losses, the training label of \LS{}
and the certification.

On Qwen3 math the conclusion is unchanged: $\Delta$ lies between \natQmathDmin{} and \natQmathDmax{} points. The sign of $\Delta$ reverses in \nRefFlips{} settings, both of them GSM8K with an R1 distillation: the static budget
leads under the end-probe reference and \LS{} under the response reference. As noted in Section~\ref{sec:value},
these models close the trace after fewer than 200 tokens on this task. They then work out the solution in their
response, and their end probe, which has to answer without that work, is \gapGsmRQ{} and \gapGsmRL{} points less
accurate than the response. Changing the reference therefore changes two things. More questions count as answered
correctly by full reasoning, and so can be lost, and a policy that stops also saves the long response. In these two
settings \LS{} then saves much more, \refRQLSResp\% instead of \refRQLSEnd\% with R1-Qwen-7B and \refRLLSResp\%
instead of \refRLLSEnd\% with R1-Llama-8B, enough to overtake the static budget, which saves \refRQFixResp\% and
\refRLFixResp\% (Appendix~\ref{app:natref}). For such models the definition of the reference decides the result, and we report
both.

\subsection{The certification procedure}
\label{sec:certrobust}

The certification of Section~\ref{sec:cert} combines an exact binomial test with fixed-sequence testing. Replacing
either by a cruder alternative, Hoeffding's inequality or a union bound over all thresholds, lowers the saving of
\LS, from \certLSBS\% on average to \certLSBU\% with the union bound and to \certLSHS\% with Hoeffding's inequality
(Appendix~\ref{app:cert}). At our calibration sizes of 99 to 660 questions the choice of the test thus matters more
than the treatment of multiplicity, and a loose test can remove the saving altogether at a small risk target.

\section{Discussion}
\label{sec:discussion}

\subsection{Relation to earlier findings}

The results qualify how savings from early exit should be read. Against full reasoning our rules save
tokens in every setting but GSM8K with R1-Llama-8B, where little reasoning is left to cut. Against a static budget that meets the same risk target, the rules that rely on one signal save less in
most settings, and the learned rule saves more in some settings and less in others. A method evaluated only against full
reasoning may be credited with savings that a single budget would also deliver. Our scalar rules implement the
published signals under one protocol and do not rank the original systems.

Conformal Thinking reports that no stopping signal is the most efficient at every risk level
\citep{wang2026conformal}. Our results extend this observation in two directions. The dependence reaches the more
basic choice of whether to learn a rule at all: the better of the two families changes at least once between risk
targets of 5\% and 20\% in about half of the settings. And the risk level is not the only condition. The ranking also changes with the price of a probe, so a result
obtained at one serving price does not determine the better policy at another.

Thought Calibration certifies a stopping rule that reads hidden states \citep{wu2025thought}.
\textsc{HiddenStop}, a linear classifier on one layer of Qwen3-8B, follows the same idea. With a KV-cache fork it
showed no stable advantage over a certified static budget. At $p=0.25$, where a rule that probes at every
checkpoint is no longer affordable, \textsc{HiddenStop} makes one answer probe per question, as the static budget
does, and it saved more than the budget on MATH-500 and less on MMLU-Pro (Section~\ref{sec:hidden}). A reader of
hidden states, like any learned rule, has to show its increment over that baseline at the price of the deployment.

Published measurements of response time point in different directions. The authors of DEER measure, with the
Hugging Face transformers library on MATH-500 and AMC 2023, that it reduces latency by 27.9--40.1\%
\citep{yang2026deer}, whereas \citet{min2026puma} measure the same method as slower than full reasoning in their
setup and attribute this to its trial-answer probes. We did not reproduce either setup. In our engine, with a
KV-cache fork, the mean time saving of \LS{} followed its token saving, and with the prefix cache switched off its
time per generated token rose only to \perTokNocMathLSd{} of that of full reasoning (Section~\ref{sec:latency}).
Processing the reasoning again therefore does not by itself slow a probing policy down to the extent that
\citet{min2026puma} report for DEER.

\subsection{Guidance for operators}
\label{sec:guidance}

Table~\ref{tab:decisions} turns the results into rules: for a condition that an operator can observe, the action that
the results support and the limit of that evidence. Two points apply across them. A static budget certified to the
risk target should be the first thing an operator tries, because it needs no training data and already removes much
of the redundant reasoning. And whether a learned rule is worth adding cannot be taken from results on another
workload, another risk target or another serving stack; it can be decided on the operator's own calibration data,
because one certification covers both families.

\begin{table}[t]
\centering
\caption{Conditions, actions and limits for an operator. Results are for $\alpha=0.05$ unless stated otherwise.}
\label{tab:decisions}
\small
\begin{tabular}{>{\raggedright\arraybackslash}p{0.23\textwidth}>{\raggedright\arraybackslash}p{0.38\textwidth}>{\raggedright\arraybackslash}p{0.30\textwidth}}
\toprule
Condition & Action & Limit of the evidence \\
\midrule
Any workload, before anything else & Certify a static budget on a fine grid at the risk target; it needs no training
data (Section~\ref{sec:value}). & It saves \dQmathAbsMin--\dQmathAbsMax{} points less than the learned rule on Qwen3
math, where answers settle at very different points. \\
A labelled training pool is available & Train the learned rule and certify both families together; deploy the cheapest
certified candidate (Section~\ref{sec:select}). & With \gpqaNcal{} calibration questions selection fell \wcSelMax{}
points short of the better family; otherwise at most \wcSelEx{} at any of the three prices. \\
Re-sent prefixes are billed ($p>0$) & Probe sparsely, or let the certification choose the schedule; behind an API
that bills the prefix as input, expect the static budget (Section~\ref{sec:conditions}). & Break-even prices of
\beQmathLo--\beQmathHi{} hold for 14 checkpoints on Qwen3 math. \\
Only about 50 labelled questions & A learned rule may keep most of its increment; the collection is repaid after
\paybackFiftyMin--\paybackFiftyMax{} requests (Section~\ref{sec:data}). & Less than half of the increment is kept in
\trainLessN{} of \nTrainWinners{} settings; labelling cost is not counted. \\
Response time matters, self-hosted engine with a prefix cache & Expect the mean response time to follow the tokens;
probes from the cached reasoning add little time (Section~\ref{sec:latency}). & One model, one engine, concurrency
16; mean times, not every percentile. \\
Audited risk above the target, or a known change of traffic & Recalibrate both families on at least \recalMinN{}
labelled questions of the new workload (\recalMinNTwo{} when selecting between two); retrain the learned rule if new
training data are cheap, and certify again (Section~\ref{sec:shift}). & Retraining kept the learned rule ahead with
Qwen3, not with the R1 distillations; moves between benchmarks are larger than most drift. \\
\bottomrule
\end{tabular}
\end{table}

Consider two services that run Qwen3-32B at a risk target of 5\%, one with questions like those of GSM8K and one
with questions like those of MMLU-Pro, each calibrated on its own workload. Table~\ref{tab:example} gives what the
procedure deploys for each on three serving stacks, when it selects among the static budgets and four checkpoint
schedules of \LS. On the GSM8K workload with a KV-cache fork the selected policy saves \exGsmKvSel\% of the
generated tokens at an accuracy of \exGsmKvSelAcc\%, where the static budget alone saves \exGsmKvFix\%; with
full reasoning at about \exGsmRef{} tokens per request, the selected policy saves about \exGsmKvSelM{} million
generated tokens per million requests, and the static budget about \exGsmKvFixM{} million.
Behind a prefix cache, where savings include the billed prefix, the selected policy still saves \exGsmPcSel\%.
Through a black-box API no certified threshold of the 14-checkpoint schedule is cheaper than full reasoning, so the
procedure selects a sparse schedule or the static budget and saves \exGsmBbSel\% on average. On the MMLU-Pro
workload the selected policy stays within \exMmluSelGapMax{} points of the static budget on all three stacks. That
small loss buys the operator freedom from having to know in advance which workload favours which family.

\begin{table}[t]
\centering
\caption{Worked example: saving (\%, in output-token equivalents) with Qwen3-32B at $\alpha=0.05$ on three serving
stacks. Static: the certified static budget. \LS: the learned rule with probes at all 14 checkpoints. Selected: the
cheapest certified candidate among the static budgets and four checkpoint schedules of \LS, chosen on each
calibration half. Bold marks the better of Static and \LS. Means over 50 splits.}
\label{tab:example}
\smallskip
\small
\begin{adjustbox}{max width=\textwidth}\begin{tabular}{llrrrrrr}
\toprule
 & & \multicolumn{3}{c}{GSM8K workload} & \multicolumn{3}{c}{MMLU-Pro workload} \\
\cmidrule(lr){3-5}\cmidrule(lr){6-8}
Serving stack & $p$ & static & \LS{} & selected & static & \LS{} & selected \\
\midrule
KV-cache fork & 0 & 33.6 & \textbf{50.6} & 49.3 & \textbf{33.4} & 19.2 & 31.2 \\
Prefix cache & 0.025 & 32.6 & \textbf{44.5} & 43.4 & \textbf{32.6} & 5.3 & 30.3 \\
Black-box API & 0.25 & \textbf{23.6} & 0.0 & 25.0 & \textbf{25.5} & 0.0 & 23.3 \\
\bottomrule
\end{tabular}\end{adjustbox}
\end{table}

After deployment the operator has to monitor the policy, and the served answers are of no help for that: a
loss is defined relative to an answer that the policy has prevented the model from producing. The realised risk can
only be estimated on a sample of the traffic for which the correct answer is known and the model is also allowed to
finish reasoning. When that estimate, or a known change in the mix of questions, indicates that the workload has
moved, Section~\ref{sec:shift} shows what to do: recalibrate both families on a new calibration set, which restores
the bound, retrain the learned rule where new training data are cheap, and let the certification choose between the
families again.

\subsection{Limitations}

Each question has one sampled trace in the main runs, so the savings do not
include the variation between traces that a service would see, and in one setting the sign of a small difference
depended on the sample (Section~\ref{sec:value}). The tasks are mathematics problems and multiple-choice questions
with short answers that can be graded automatically. The design is not a full grid: AQuA and GPQA-D were run with
the Qwen3 models only, so counts over the \nSettings{} settings give more weight to GSM8K, MATH-500, MMLU-Pro and
Qwen3. The hidden-state rule covers two settings with one model.

The method has limits of its own. Probes are decoded greedily and capped at 48 tokens, which suits short answers
and truncates long free-form ones (Appendix~\ref{app:bbh}). The learned rule is a logistic regression on eight
features, and a stronger classifier might alter the comparison in either direction. The change of workload in Section~\ref{sec:shift} is a move between benchmarks, which is a larger change
than most drifts in production traffic, and we did not test a method for detecting the change.

Costs are counts of tokens under a price model. The response times of Section~\ref{sec:latency} were measured in
two settings on one engine, and elsewhere a token saving should not be read as a saving in time, because each probe
adds a call whose overhead depends on the engine. Energy was not measured.

Finally, early exit trades compute for a bounded risk. The bound holds on average over questions from the
calibrated distribution, with the stated confidence, and says nothing about any individual answer: a lost answer is
returned with the same appearance as every other answer. Where a wrong answer is costly, full reasoning should
remain available.

\section{Conclusion}

An operator who adds early exit to a reasoning-model service should measure a learned stopping rule against a
certified static budget, and can do so on calibration data from the workload itself. Under one risk bound and one
cost model neither is better throughout: the learned rule saved considerably more on free-form mathematics with
Qwen3, and the static budget saved more on MMLU-Pro on average. The order also changed with the risk target, in
about half of the settings, and with the price of a probe. Committing to either family in advance therefore cost a
large part of the saving on some workload, whereas selecting between them on calibration data fell at most \wcSelEx{} points
short of the better family except on the task with the smallest calibration set.

A training pool of fifty questions preserved at least three quarters of the increment of the learned rule in
\trainMostN{} of \nTrainWinners{} settings and less than half in \trainLessN; calibration requires separate labelled questions in either case. In a serving engine
the mean response times followed the token savings. The risk control did not carry over to a different task: both
families had to be recalibrated there, and whether the learned rule then kept its lead depended on the model. The
analysis code is released and the recorded probes are available on request, so that further stopping rules can be evaluated under the same
protocol.

\appendix
\counterwithin{table}{section}
\counterwithin{figure}{section}
\counterwithin{table}{section}
\counterwithin{figure}{section}
\counterwithin{equation}{section}
\makeatletter\let\eswa@section\section\renewcommand{\section}{\FloatBarrier\eswa@section}\makeatother

\section{Proof of Proposition~\ref{prop:ltt}}
\label{app:proof}

\textit{Proof.} For $R(\lambda)>\alpha$, $\Pr[\mathrm{Bin}(n,R(\lambda))\le k]\le\Pr[\mathrm{Bin}(n,\alpha)\le k]$
for every $k$, so $p_\lambda$ is a valid p-value. Fixed-sequence testing at level $\delta/m$ rejects a true null within
a family with probability at most $\delta/m$, because the first true null in the fixed order must be rejected for any
later one to be tested; a union bound over the $m$ families gives probability at most $\delta$ of any false rejection
\citep{angelopoulos2025ltt}. The fallback to full reasoning has $L\equiv0$. On the complementary event every element of
$\hat\Lambda$ has $R\le\alpha$, whichever element the selection picks. \qed

\section{Data, grading and decoding details}
\label{app:data}

Evaluation and training pools are drawn with a fixed seed, and no training question appears in an evaluation set. MATH
training problems come from the MATH training split; their gold answer is the content of the last \verb|\boxed{}| in the
reference solution, extracted with brace matching. GPQA options are shuffled with a seed derived from the question text.
A probe answer is extracted from the kept text of the probe, that is, its first line: it is the content of the
first \verb|\boxed{}| in that text if there is one, and the text itself otherwise. Numeric answers are compared as numbers after removing thousands separators and \LaTeX{} spacing; MATH answers are
compared after the answer normalisation of the MATH release code \citep{hendrycks2021math} with a numeric fallback; multiple-choice answers are the
first standalone option letter. The free response is graded from its last \verb|\boxed{}|, the text after its last
``Final answer:'', or, for multiple choice, its last explicit answer statement. Probe confidence and entropy are
averaged over the tokens up to the newline stop, with entropy over the top-20 log-probabilities returned by the engine.
The four scalar exits use the following scores: the confidence of the probe answer, which is the exponential of its
mean token log-probability; its negative mean entropy; the number of consecutive checkpoints that returned the same
answer; and a confidence leap, the increase in confidence over the previous checkpoint multiplied by the current
confidence.

\section{Results at other risk targets}
\label{app:alpha}

Table~\ref{tab:alpha} gives $\Delta$ at the four risk targets in every setting. In \nSignEnds{} of the \nSignChange{}
settings in which its sign changes, the policy that saves more at $\alpha=0.20$ is not the one that saves more at
$\alpha=0.05$. In the other \nSmallFlip{} the change from the smallest to the largest value of $\Delta$ is at most
\smallFlipMax{} points.

\begin{table}[!htbp]
\centering
\caption{The sign of $\Delta$ depends on the risk target. $\Delta$ (\LS{} minus static budget, percentage points) at each
$\alpha$, mean over 50 calibration/test splits, bold when the 2.5--97.5\% range over splits excludes zero; each policy's
saving at $\alpha=0.05$ and $0.20$. KV-cache fork, measured probe cost. Last column: sign of the mean difference across
the four targets.}
\label{tab:alpha}
\smallskip
\footnotesize\setlength{\tabcolsep}{3.4pt}
\begin{adjustbox}{max width=\textwidth}\begin{tabular}{lrrrr@{\hspace{10pt}}rr@{\hspace{10pt}}rr@{\hspace{10pt}}l}
\toprule
 & \multicolumn{4}{c}{\LS{} $-$ static budget (pp)} & \multicolumn{2}{c}{static saving} & \multicolumn{2}{c}{\LS{} saving} & \\
\cmidrule(lr){2-5}\cmidrule(lr){6-7}\cmidrule(lr){8-9}
Task / model & $\alpha$=0.05 & 0.10 & 0.15 & 0.20 & 0.05 & 0.20 & 0.05 & 0.20 & over $\alpha$ \\
\midrule
GSM8K / Qwen3-8B & \textbf{+11.7} & \textbf{+6.6} & \textbf{+4.6} & +4.4 & 34.1 & 61.6 & 45.7 & 65.9 & \LS{} ahead \\
GSM8K / Qwen3-32B & \textbf{+17.0} & \textbf{+18.3} & \textbf{+20.5} & \textbf{+13.2} & 33.6 & 59.4 & 50.6 & 72.6 & \LS{} ahead \\
GSM8K / R1-Qwen-7B & \textbf{-8.2} & -1.4 & \textbf{+4.7} & \textbf{+10.9} & 16.5 & 17.7 & 8.2 & 28.6 & ranking changes \\
GSM8K / R1-Llama-8B & \textbf{-1.3} & +1.4 & \textbf{+6.2} & \textbf{+12.8} & 1.3 & 1.3 & 0.0 & 14.1 & ranking changes \\
\midrule
MATH-500 / Qwen3-8B & \textbf{+11.5} & \textbf{+13.8} & \textbf{+9.5} & \textbf{+7.5} & 13.7 & 59.0 & 25.2 & 66.5 & \LS{} ahead \\
MATH-500 / Qwen3-32B & \textbf{+10.4} & +10.1 & +9.9 & \textbf{+8.2} & 18.3 & 60.0 & 28.7 & 68.2 & \LS{} ahead \\
MATH-500 / R1-Qwen-7B & -4.1 & -3.2 & +0.4 & -1.9 & 22.9 & 58.2 & 18.8 & 56.3 & ranking changes \\
MATH-500 / R1-Llama-8B & -6.2 & -1.8 & -0.5 & -2.6 & 21.2 & 68.4 & 14.9 & 65.8 & static ahead \\
\midrule
MMLU-Pro / Qwen3-8B & -5.2 & +2.5 & -0.2 & -2.9 & 31.1 & 74.1 & 26.0 & 71.3 & ranking changes \\
MMLU-Pro / Qwen3-32B & \textbf{-14.3} & -3.9 & -4.4 & +2.0 & 33.4 & 70.6 & 19.2 & 72.5 & ranking changes \\
MMLU-Pro / R1-Qwen-7B & -4.4 & -4.7 & -4.4 & -1.8 & 24.2 & 70.0 & 19.8 & 68.2 & static ahead \\
MMLU-Pro / R1-Llama-8B & \textbf{-6.3} & -3.7 & \textbf{-3.9} & -2.2 & 30.2 & 71.2 & 23.9 & 69.0 & static ahead \\
\midrule
AQuA / Qwen3-8B & +10.4 & -0.0 & -2.7 & -0.9 & 18.7 & 70.8 & 29.1 & 69.9 & ranking changes \\
\midrule
GPQA-D / Qwen3-8B & +7.0 & +3.7 & -6.9 & -7.1 & 4.0 & 54.9 & 11.0 & 47.8 & ranking changes \\
GPQA-D / Qwen3-32B & \textbf{+14.6} & +9.5 & +1.6 & -3.6 & 4.7 & 55.0 & 19.2 & 51.4 & ranking changes \\
\bottomrule
\end{tabular}\end{adjustbox}
\end{table}

\begin{figure}[!htbp]
\centering
\includegraphics[width=\linewidth]{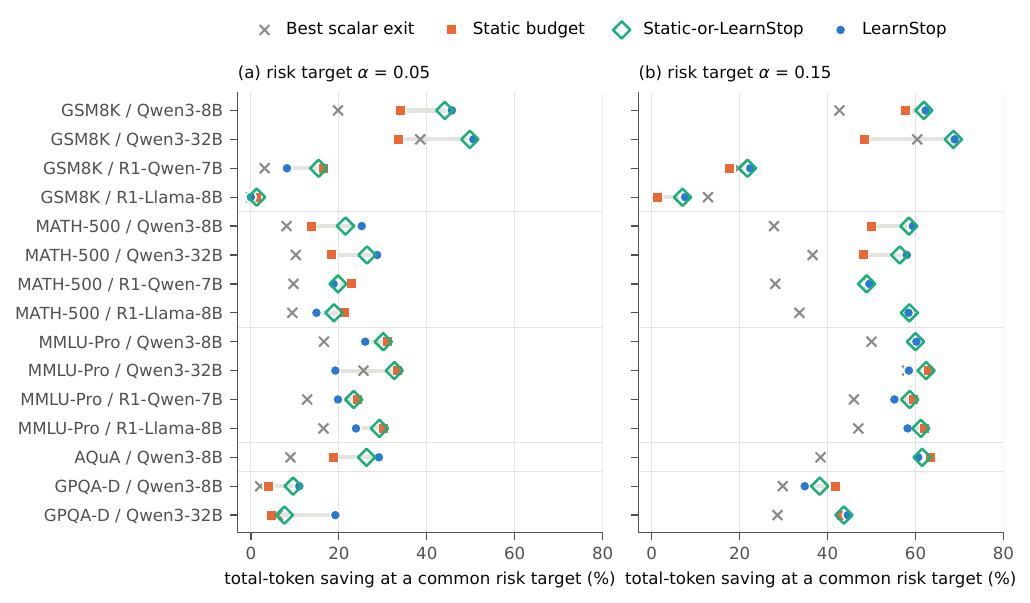}
\caption{From $\alpha=0.05$ to $0.15$ the ranking of the two policies changes in several settings, in both
directions. Total-token saving of each
policy at a common risk target, KV-cache fork, measured probe cost, mean over 50 calibration/test splits; both
panels share the $x$-axis. The gray bar joins the static budget and \LS; the open diamond is \SLS.}
\label{fig:map}
\end{figure}

\begin{table}[!htbp]
\centering
\caption{As Table~\ref{tab:main}, at $\alpha=0.15$.}
\smallskip
\footnotesize\setlength{\tabcolsep}{3.2pt}
\begin{adjustbox}{max width=\textwidth}\begin{tabular}{lrrrrrrrrlr}
\toprule
 & full & \multicolumn{6}{c}{total-token saving (\%)} & LS & \multicolumn{2}{c}{LearnStop $-$ static budget (pp)} \\
\cmidrule(lr){3-8}\cmidrule(lr){10-11}
Task / model & acc. & static & scalar & conv. & LS & LS+cap & st./LS & acc. & mean [2.5\%, 97.5\%] & win \% \\
\midrule
GSM8K / Qwen3-8B & .960 & 57.7 & 42.7 & 40.2 & 62.3 & \textbf{62.5} & 61.9 & .848 & \textbf{+4.6 [+2.6, +7.0]} & 100 \\
GSM8K / Qwen3-32B & .968 & 48.4 & 60.4 & 42.2 & \textbf{68.9} & 68.7 & 68.6 & .851 & \textbf{+20.5 [+18.4, +23.2]} & 100 \\
GSM8K / R1-Qwen-7B & .830 & 17.7 & 19.7 & 10.6 & 22.4 & \textbf{23.1} & 21.8 & .715 & \textbf{+4.7 [+0.8, +8.9]} & 98 \\
GSM8K / R1-Llama-8B & .658 & 1.3 & \textbf{12.8} & -0.2 & 7.6 & 6.5 & 7.0 & .539 & \textbf{+6.2 [+3.8, +9.3]} & 100 \\
\midrule
MATH-500 / Qwen3-8B & .922 & 49.9 & 27.8 & 32.6 & \textbf{59.4} & 57.3 & 58.5 & .832 & \textbf{+9.5 [+3.0, +19.5]} & 98 \\
MATH-500 / Qwen3-32B & .938 & 48.1 & 36.6 & 40.3 & \textbf{58.0} & 56.6 & 56.4 & .833 & +9.9 [-2.3, +20.2] & 94 \\
MATH-500 / R1-Qwen-7B & .885 & 49.2 & 28.1 & 24.9 & 49.5 & \textbf{53.5} & 48.9 & .783 & +0.4 [-7.3, +9.7] & 46 \\
MATH-500 / R1-Llama-8B & .730 & \textbf{58.9} & 33.6 & 39.0 & 58.4 & 56.7 & 58.6 & .626 & -0.5 [-7.1, +6.3] & 38 \\
\midrule
MMLU-Pro / Qwen3-8B & .757 & 60.4 & 50.0 & 52.8 & 60.2 & \textbf{60.9} & 60.0 & .657 & -0.2 [-4.6, +5.4] & 48 \\
MMLU-Pro / Qwen3-32B & .823 & \textbf{62.9} & 58.1 & 59.1 & 58.5 & 61.9 & 62.4 & .716 & -4.4 [-10.7, +0.9] & 8 \\
MMLU-Pro / R1-Qwen-7B & .589 & \textbf{59.6} & 46.0 & 42.5 & 55.2 & 57.5 & 58.7 & .500 & -4.4 [-11.2, +3.6] & 10 \\
MMLU-Pro / R1-Llama-8B & .604 & \textbf{62.1} & 47.0 & 48.1 & 58.2 & 59.6 & 61.2 & .518 & \textbf{-3.9 [-8.0, -0.3]} & 2 \\
\midrule
AQuA / Qwen3-8B & .901 & \textbf{63.3} & 38.4 & 40.3 & 60.6 & 60.3 & 61.5 & .794 & -2.7 [-7.6, +3.2] & 28 \\
\midrule
GPQA-D / Qwen3-8B & .617 & \textbf{41.7} & 29.8 & 37.5 & 34.8 & 33.7 & 38.2 & .548 & -6.9 [-24.7, +8.6] & 30 \\
GPQA-D / Qwen3-32B & .630 & 43.1 & 28.6 & 33.7 & \textbf{44.6} & 43.9 & 43.7 & .563 & +1.6 [-11.0, +11.9] & 62 \\
\bottomrule
\end{tabular}\end{adjustbox}
\end{table}

\begin{table}[!htbp]
\centering
\caption{Savings (\%) across risk targets: static budget (st.), \LS{} (LS) and \LSC{} (cap). KV-cache fork, measured probe
cost.}
\smallskip
\footnotesize\setlength{\tabcolsep}{2.6pt}
\begin{adjustbox}{max width=\textwidth}\begin{tabular}{lrrrrrrrrrrrr}
\toprule
 & \multicolumn{3}{c}{$\alpha=0.05$} & \multicolumn{3}{c}{$\alpha=0.10$} & \multicolumn{3}{c}{$\alpha=0.15$} & \multicolumn{3}{c}{$\alpha=0.20$} \\
\cmidrule(lr){2-4}\cmidrule(lr){5-7}\cmidrule(lr){8-10}\cmidrule(lr){11-13}
Task / model & st. & LS & cap & st. & LS & cap & st. & LS & cap & st. & LS & cap \\
\midrule
GSM8K / Qwen3-8B & 34.1 & 45.7 & 42.3 & 49.7 & 56.3 & 55.3 & 57.7 & 62.3 & 62.5 & 61.6 & 65.9 & 66.3 \\
GSM8K / Qwen3-32B & 33.6 & 50.6 & 49.8 & 45.4 & 63.7 & 62.2 & 48.4 & 68.9 & 68.7 & 59.4 & 72.6 & 72.4 \\
GSM8K / R1-Qwen-7B & 16.5 & 8.2 & 11.1 & 17.7 & 16.4 & 16.8 & 17.7 & 22.4 & 23.1 & 17.7 & 28.6 & 28.3 \\
GSM8K / R1-Llama-8B & 1.3 & 0.0 & 0.0 & 1.3 & 2.7 & 1.9 & 1.3 & 7.6 & 6.5 & 1.3 & 14.1 & 12.4 \\
\midrule
MATH-500 / Qwen3-8B & 13.7 & 25.2 & 16.7 & 35.7 & 49.5 & 45.8 & 49.9 & 59.4 & 57.3 & 59.0 & 66.5 & 64.9 \\
MATH-500 / Qwen3-32B & 18.3 & 28.7 & 26.2 & 33.9 & 44.0 & 42.3 & 48.1 & 58.0 & 56.6 & 60.0 & 68.2 & 66.1 \\
MATH-500 / R1-Qwen-7B & 22.9 & 18.8 & 18.2 & 40.5 & 37.3 & 42.9 & 49.2 & 49.5 & 53.5 & 58.2 & 56.3 & 60.2 \\
MATH-500 / R1-Llama-8B & 21.2 & 14.9 & 10.5 & 42.1 & 40.3 & 34.3 & 58.9 & 58.4 & 56.7 & 68.4 & 65.8 & 68.5 \\
\midrule
MMLU-Pro / Qwen3-8B & 31.1 & 26.0 & 27.2 & 47.3 & 49.8 & 48.2 & 60.4 & 60.2 & 60.9 & 74.1 & 71.3 & 70.6 \\
MMLU-Pro / Qwen3-32B & 33.4 & 19.2 & 24.6 & 48.4 & 44.5 & 46.3 & 62.9 & 58.5 & 61.9 & 70.6 & 72.5 & 74.4 \\
MMLU-Pro / R1-Qwen-7B & 24.2 & 19.8 & 18.1 & 43.6 & 38.8 & 40.1 & 59.6 & 55.2 & 57.5 & 70.0 & 68.2 & 69.5 \\
MMLU-Pro / R1-Llama-8B & 30.2 & 23.9 & 22.4 & 49.4 & 45.6 & 45.4 & 62.1 & 58.2 & 59.6 & 71.2 & 69.0 & 69.4 \\
\midrule
AQuA / Qwen3-8B & 18.7 & 29.1 & 23.3 & 51.5 & 51.4 & 50.0 & 63.3 & 60.6 & 60.3 & 70.8 & 69.9 & 69.5 \\
\midrule
GPQA-D / Qwen3-8B & 4.0 & 11.0 & 9.5 & 18.7 & 22.3 & 15.3 & 41.7 & 34.8 & 33.7 & 54.9 & 47.8 & 50.6 \\
GPQA-D / Qwen3-32B & 4.7 & 19.2 & 8.1 & 24.8 & 34.3 & 28.7 & 43.1 & 44.6 & 43.9 & 55.0 & 51.4 & 53.0 \\
\bottomrule
\end{tabular}\end{adjustbox}
\end{table}

\section{Realised risk, accuracy, certification and deployment of every policy}
\label{app:policyfull}

Across policies and settings, the empirical risk on a test half exceeds $\alpha$ in \violMean\% of
policy--setting--split cases and in at most \violMax\% of the splits of any policy and setting, which a finite test
half can show when the true risk is below $\alpha$.

\begin{table}[!htbp]
\centering
\caption{At $\alpha=0.05$, KV-cache fork: mean test lost-correct risk and test accuracy over 50 splits. The
certification target is the same for every policy; realised risk and accuracy are not.}
\smallskip
\footnotesize\setlength{\tabcolsep}{3.2pt}
\begin{adjustbox}{max width=\textwidth}\begin{tabular}{lrrrrrrrrrrr}
\toprule
 & full & \multicolumn{2}{c}{static} & \multicolumn{2}{c}{scalar} & \multicolumn{2}{c}{LS} & \multicolumn{2}{c}{LS+cap} & \multicolumn{2}{c}{st./LS} \\
\cmidrule(lr){3-4}\cmidrule(lr){5-6}\cmidrule(lr){7-8}\cmidrule(lr){9-10}\cmidrule(lr){11-12}
Task / model & acc. & risk & acc. & risk & acc. & risk & acc. & risk & acc. & risk & acc. \\
\midrule
GSM8K / Qwen3-8B & .960 & 0.032 & .934 & 0.028 & .935 & 0.036 & .932 & 0.031 & .936 & 0.033 & .935 \\
GSM8K / Qwen3-32B & .968 & 0.034 & .938 & 0.030 & .940 & 0.034 & .939 & 0.030 & .943 & 0.030 & .942 \\
GSM8K / R1-Qwen-7B & .830 & 0.028 & .804 & 0.027 & .805 & 0.034 & .800 & 0.011 & .820 & 0.024 & .808 \\
GSM8K / R1-Llama-8B & .658 & 0.014 & .644 & 0.000 & .658 & 0.000 & .658 & 0.000 & .658 & 0.014 & .644 \\
\midrule
MATH-500 / Qwen3-8B & .922 & 0.025 & .900 & 0.022 & .906 & 0.029 & .897 & 0.018 & .908 & 0.024 & .901 \\
MATH-500 / Qwen3-32B & .938 & 0.022 & .917 & 0.019 & .921 & 0.026 & .914 & 0.020 & .920 & 0.022 & .918 \\
MATH-500 / R1-Qwen-7B & .885 & 0.026 & .863 & 0.016 & .871 & 0.025 & .862 & 0.021 & .867 & 0.027 & .861 \\
MATH-500 / R1-Llama-8B & .730 & 0.028 & .705 & 0.023 & .708 & 0.023 & .709 & 0.019 & .713 & 0.025 & .708 \\
\midrule
MMLU-Pro / Qwen3-8B & .757 & 0.031 & .738 & 0.027 & .737 & 0.031 & .736 & 0.030 & .737 & 0.031 & .737 \\
MMLU-Pro / Qwen3-32B & .823 & 0.033 & .797 & 0.028 & .805 & 0.031 & .796 & 0.024 & .805 & 0.030 & .800 \\
MMLU-Pro / R1-Qwen-7B & .589 & 0.033 & .568 & 0.023 & .574 & 0.030 & .573 & 0.027 & .573 & 0.033 & .569 \\
MMLU-Pro / R1-Llama-8B & .604 & 0.033 & .589 & 0.024 & .587 & 0.032 & .587 & 0.027 & .592 & 0.032 & .590 \\
\midrule
AQuA / Qwen3-8B & .901 & 0.021 & .889 & 0.023 & .879 & 0.025 & .879 & 0.018 & .886 & 0.022 & .883 \\
\midrule
GPQA-D / Qwen3-8B & .617 & 0.018 & .601 & 0.012 & .607 & 0.011 & .612 & 0.000 & .622 & 0.001 & .621 \\
GPQA-D / Qwen3-32B & .630 & 0.010 & .637 & 0.018 & .615 & 0.018 & .623 & 0.012 & .624 & 0.012 & .625 \\
\bottomrule
\end{tabular}\end{adjustbox}
\end{table}

\begin{table}[!htbp]
\centering
\caption{At $\alpha=0.05$, KV-cache fork, share of splits (\%) in which some candidate other than full reasoning is
certified (cert.) and in which the deployed candidate is not full reasoning (dep.), that is, some certified candidate
is cheaper on the calibration half. A conservative threshold can be certified and still cost more than full
reasoning, because it pays for probes at many checkpoints and rarely stops.}
\smallskip
\footnotesize\setlength{\tabcolsep}{4pt}
\begin{adjustbox}{max width=\textwidth}\begin{tabular}{lrrrrrrrrrr}
\toprule
 & \multicolumn{2}{c}{static} & \multicolumn{2}{c}{scalar} & \multicolumn{2}{c}{LS} & \multicolumn{2}{c}{LS+cap} & \multicolumn{2}{c}{st./LS} \\
\cmidrule(lr){2-3}\cmidrule(lr){4-5}\cmidrule(lr){6-7}\cmidrule(lr){8-9}\cmidrule(lr){10-11}
Task / model & cert. & dep. & cert. & dep. & cert. & dep. & cert. & dep. & cert. & dep. \\
\midrule
GSM8K / Qwen3-8B & 100 & 100 & 100 & 100 & 100 & 100 & 100 & 100 & 100 & 100 \\
GSM8K / Qwen3-32B & 100 & 100 & 100 & 100 & 100 & 100 & 100 & 100 & 100 & 100 \\
GSM8K / R1-Qwen-7B & 100 & 100 & 100 & 100 & 100 & 100 & 100 & 100 & 100 & 100 \\
GSM8K / R1-Llama-8B & 100 & 100 & 100 & 0 & 100 & 0 & 100 & 0 & 100 & 100 \\
\midrule
MATH-500 / Qwen3-8B & 100 & 100 & 100 & 100 & 100 & 100 & 100 & 98 & 100 & 100 \\
MATH-500 / Qwen3-32B & 100 & 100 & 100 & 100 & 100 & 100 & 100 & 100 & 100 & 100 \\
MATH-500 / R1-Qwen-7B & 100 & 100 & 100 & 100 & 100 & 100 & 100 & 100 & 100 & 100 \\
MATH-500 / R1-Llama-8B & 100 & 100 & 100 & 100 & 100 & 100 & 100 & 100 & 100 & 100 \\
\midrule
MMLU-Pro / Qwen3-8B & 100 & 100 & 100 & 100 & 100 & 100 & 100 & 100 & 100 & 100 \\
MMLU-Pro / Qwen3-32B & 100 & 100 & 100 & 100 & 100 & 100 & 100 & 100 & 100 & 100 \\
MMLU-Pro / R1-Qwen-7B & 100 & 100 & 100 & 100 & 100 & 100 & 100 & 100 & 100 & 100 \\
MMLU-Pro / R1-Llama-8B & 100 & 100 & 100 & 100 & 100 & 100 & 100 & 100 & 100 & 100 \\
\midrule
AQuA / Qwen3-8B & 100 & 100 & 100 & 100 & 100 & 100 & 100 & 100 & 100 & 100 \\
\midrule
GPQA-D / Qwen3-8B & 100 & 100 & 76 & 46 & 100 & 100 & 100 & 100 & 100 & 100 \\
GPQA-D / Qwen3-32B & 100 & 100 & 100 & 98 & 100 & 100 & 100 & 52 & 100 & 100 \\
\bottomrule
\end{tabular}\end{adjustbox}
\end{table}

\section{Question-level uncertainty}
\label{app:boot}

With the policies frozen on each calibration half, a question-level bootstrap gives intervals of median half-width
\bootHalfMin--\bootHalfMax{} points, of the same order as the standard deviation of $\Delta$ across splits
(\bootSdMin--\bootSdMax{} points). Such an interval holds the calibration draw fixed, so it does not show how $\Delta$
varies when the calibration questions change.

\begin{table}[!htbp]
\centering
\caption{\LS{} minus static budget at $\alpha=0.05$, KV-cache fork (pp). Left: mean, 2.5--97.5\% range and standard
deviation of the paired difference over 50 calibration/test splits. Right: for the two policies frozen on each split's
calibration half, a paired bootstrap over the questions of its test half (2{,}000 resamples); median half-width of the
95\% interval, and share of splits whose interval lies entirely above or below zero.}
\smallskip
\small\setlength{\tabcolsep}{4pt}
\begin{adjustbox}{max width=\textwidth}\begin{tabular}{lrlrrrr}
\toprule
 & \multicolumn{3}{c}{across 50 splits} & \multicolumn{3}{c}{question-level bootstrap within a split} \\
\cmidrule(lr){2-4}\cmidrule(lr){5-7}
Task / model & mean & [2.5\%, 97.5\%] & s.d. & median $\pm$ & $>0$ (\%) & $<0$ (\%) \\
\midrule
GSM8K / Qwen3-8B & +11.7 & [+2.7, +18.5] & 4.3 & 2.4 & 100 & 0 \\
GSM8K / Qwen3-32B & +17.0 & [+9.1, +25.1] & 4.4 & 3.6 & 100 & 0 \\
GSM8K / R1-Qwen-7B & -8.2 & [-12.4, -2.7] & 2.9 & 3.5 & 0 & 98 \\
GSM8K / R1-Llama-8B & -1.3 & [-1.7, -1.0] & 0.2 & 0.6 & 0 & 100 \\
\midrule
MATH-500 / Qwen3-8B & +11.5 & [+0.8, +22.4] & 6.0 & 5.0 & 82 & 0 \\
MATH-500 / Qwen3-32B & +10.4 & [+0.3, +19.6] & 5.6 & 6.7 & 72 & 0 \\
MATH-500 / R1-Qwen-7B & -4.1 & [-18.4, +10.7] & 7.4 & 8.2 & 8 & 30 \\
MATH-500 / R1-Llama-8B & -6.2 & [-15.4, +4.0] & 5.7 & 6.9 & 0 & 46 \\
\midrule
MMLU-Pro / Qwen3-8B & -5.2 & [-14.6, +3.8] & 5.0 & 4.3 & 4 & 56 \\
MMLU-Pro / Qwen3-32B & -14.3 & [-23.4, -5.4] & 5.1 & 5.0 & 0 & 98 \\
MMLU-Pro / R1-Qwen-7B & -4.4 & [-11.0, +3.8] & 4.3 & 4.6 & 2 & 54 \\
MMLU-Pro / R1-Llama-8B & -6.3 & [-12.5, -1.0] & 3.6 & 2.7 & 2 & 86 \\
\midrule
AQuA / Qwen3-8B & +10.4 & [-0.4, +22.1] & 5.7 & 4.9 & 84 & 2 \\
\midrule
GPQA-D / Qwen3-8B & +7.0 & [-1.6, +14.3] & 4.5 & 5.3 & 68 & 0 \\
GPQA-D / Qwen3-32B & +14.6 & [+0.5, +27.4] & 8.7 & 7.2 & 78 & 2 \\
\bottomrule
\end{tabular}\end{adjustbox}
\end{table}

\section{Common risk target and equal realised risk}
\label{app:realised}

\begin{figure}[!htbp]
\centering
\includegraphics[width=\linewidth]{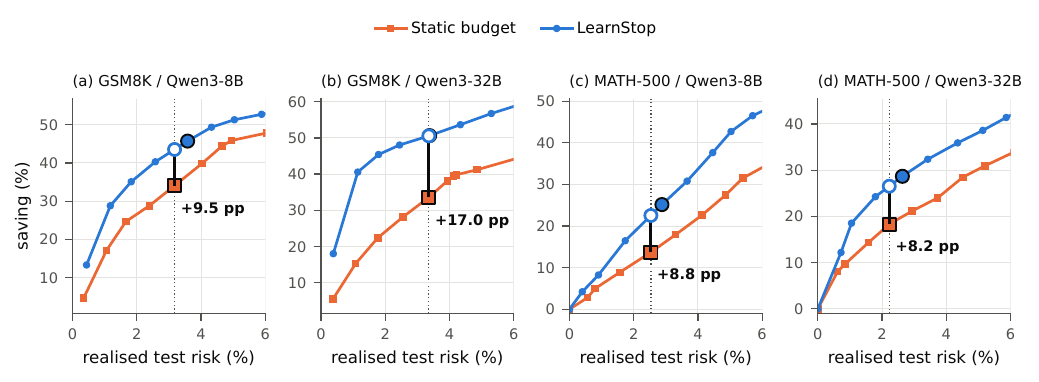}
\caption{With the fine static grid the two policies reach similar realised risks, and $\Delta$ at the static budget's
realised risk is close to $\Delta$ at the common risk target. Mean test risk against saving over 50 splits for $\alpha$
from \alphaGridMin{} upward, each point certified and selected on the calibration half (KV-cache fork); realised risks
up to 6\% are shown (full range in Figure~\ref{fig:realisedfull}). Large markers: $\alpha=0.05$. Dotted line: the
static budget's mean realised risk at $\alpha=0.05$; the open circle is \LS's curve interpolated there (post hoc), and
the labelled segment is the interpolated difference.}
\label{fig:realised}
\end{figure}

\begin{figure}[!htbp]
\centering
\includegraphics[width=\linewidth]{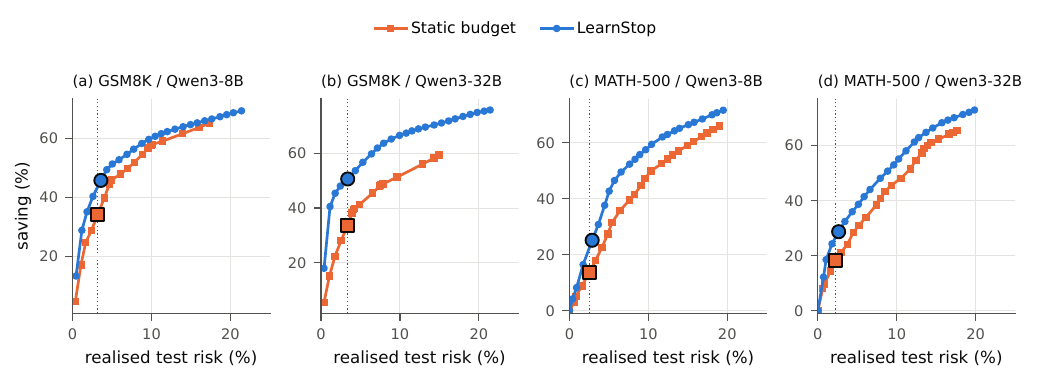}
\caption{Figure~\ref{fig:realised} over the whole $\alpha$ grid (\alphaGridMin--\alphaGridMax). Large markers:
$\alpha=0.05$; dotted line: the static budget's mean realised risk at $\alpha=0.05$.}
\label{fig:realisedfull}
\end{figure}

\begin{table}[!htbp]
\centering
\caption{Mean test risk of each policy at $\alpha=0.05$ and the saving difference at the same target or at the same
realised risk. Same realised risk: \LS's risk--saving curve over $\alpha\in\{\alphaGridMin,\dots,\alphaGridMax\}$ (made
non-decreasing in risk by a running maximum of the saving, then interpolated linearly) read at the static budget's realised risk; -- where the static budget's risk is
below every point of \LS's curve. KV-cache fork.}
\smallskip
\small
\begin{adjustbox}{max width=\textwidth}\begin{tabular}{lrrrr}
\toprule
 & \multicolumn{2}{c}{realised test risk at $\alpha=0.05$ (\%)} & \multicolumn{2}{c}{\LS{} $-$ static budget (pp)} \\
\cmidrule(lr){2-3}\cmidrule(lr){4-5}
Task / model & static & \LS & same target & same realised risk \\
\midrule
GSM8K / Qwen3-8B & 3.2 & 3.6 & +11.7 & +9.5 \\
GSM8K / Qwen3-32B & 3.4 & 3.4 & +17.0 & +17.0 \\
GSM8K / R1-Qwen-7B & 2.8 & 3.4 & -8.2 & -9.3 \\
GSM8K / R1-Llama-8B & 1.4 & 0.0 & -1.3 & -1.2 \\
\midrule
MATH-500 / Qwen3-8B & 2.5 & 2.9 & +11.5 & +8.8 \\
MATH-500 / Qwen3-32B & 2.2 & 2.6 & +10.4 & +8.2 \\
MATH-500 / R1-Qwen-7B & 2.6 & 2.5 & -4.1 & -3.4 \\
MATH-500 / R1-Llama-8B & 2.8 & 2.3 & -6.2 & -3.4 \\
\midrule
MMLU-Pro / Qwen3-8B & 3.1 & 3.1 & -5.2 & -5.4 \\
MMLU-Pro / Qwen3-32B & 3.3 & 3.1 & -14.3 & -13.2 \\
MMLU-Pro / R1-Qwen-7B & 3.3 & 3.0 & -4.4 & -2.8 \\
MMLU-Pro / R1-Llama-8B & 3.3 & 3.2 & -6.3 & -5.6 \\
\midrule
AQuA / Qwen3-8B & 2.1 & 2.5 & +10.4 & +6.9 \\
\midrule
GPQA-D / Qwen3-8B & 1.8 & 1.1 & +7.0 & +8.6 \\
GPQA-D / Qwen3-32B & 1.0 & 1.8 & +14.6 & +3.7 \\
\bottomrule
\end{tabular}\end{adjustbox}
\end{table}

\section{Probe accounting}
\label{app:accounting}

\begin{table}[!htbp]
\centering
\caption{Savings (\%) at $\alpha=0.05$ under three probe accountings: decoding to the first newline (a deployed stop
string), decoding to end of sequence or the 48-token cap, and a flat 48 tokens per probe.}
\smallskip
\small
\begin{adjustbox}{max width=\textwidth}\begin{tabular}{lrrrrrr}
\toprule
 & \multicolumn{2}{c}{to newline (deployed)} & \multicolumn{2}{c}{to EOS or 48} & \multicolumn{2}{c}{48 per probe} \\
\cmidrule(lr){2-3}\cmidrule(lr){4-5}\cmidrule(lr){6-7}
Task / model & static & LS & static & LS & static & LS \\
\midrule
GSM8K / Qwen3-8B & 34.1 & 45.7 & 33.3 & 26.7 & 33.1 & 23.8 \\
GSM8K / Qwen3-32B & 33.6 & 50.6 & 33.4 & 46.9 & 32.1 & 32.5 \\
GSM8K / R1-Qwen-7B & 16.5 & 8.2 & 16.7 & 8.7 & 14.0 & 0.0 \\
GSM8K / R1-Llama-8B & 1.3 & 0.0 & 0.9 & 0.0 & 1.0 & 0.0 \\
\midrule
MATH-500 / Qwen3-8B & 13.7 & 25.2 & 14.1 & 20.8 & 14.0 & 17.5 \\
MATH-500 / Qwen3-32B & 18.3 & 28.7 & 18.3 & 28.6 & 18.1 & 20.2 \\
MATH-500 / R1-Qwen-7B & 22.9 & 18.8 & 25.6 & 20.7 & 25.3 & 11.9 \\
MATH-500 / R1-Llama-8B & 21.2 & 14.9 & 21.1 & 9.6 & 20.8 & 5.6 \\
\midrule
MMLU-Pro / Qwen3-8B & 31.1 & 26.0 & 30.9 & 20.2 & 30.7 & 16.6 \\
MMLU-Pro / Qwen3-32B & 33.4 & 19.2 & 33.5 & 18.9 & 32.8 & 5.2 \\
MMLU-Pro / R1-Qwen-7B & 24.2 & 19.8 & 24.1 & 19.0 & 23.9 & 11.0 \\
MMLU-Pro / R1-Llama-8B & 30.2 & 23.9 & 30.1 & 22.5 & 29.8 & 15.4 \\
\midrule
AQuA / Qwen3-8B & 18.7 & 29.1 & 18.4 & 16.6 & 18.4 & 14.8 \\
\midrule
GPQA-D / Qwen3-8B & 4.0 & 11.0 & 4.0 & 6.0 & 4.0 & 4.2 \\
GPQA-D / Qwen3-32B & 4.7 & 19.2 & 4.7 & 18.1 & 4.6 & 10.5 \\
\bottomrule
\end{tabular}\end{adjustbox}
\end{table}

\section{Free response as the reference}
\label{app:natref}

\begin{table}[!htbp]
\centering
\caption{Savings (\%) at $\alpha=0.05$, KV-cache fork, with full reasoning scored by the end probe (left) or by the
model's free response (right). $\Delta$ is \LS{} minus static budget; $\leftrightarrow$ marks a reversal of sign with a
difference of at least one point on either side.}
\label{tab:ref}
\smallskip
\small
\begin{adjustbox}{max width=\textwidth}\begin{tabular}{lrrrrrrc}
\toprule
 & \multicolumn{3}{c}{end probe after the full trace} & \multicolumn{3}{c}{model's free response} & \\
\cmidrule(lr){2-4}\cmidrule(lr){5-7}
Task / model & static & LS & $\Delta$ & static & LS & $\Delta$ & sign \\
\midrule
GSM8K / Qwen3-8B & 34.1 & 45.7 & +11.7 & 38.1 & 51.2 & +13.1 &  \\
GSM8K / Qwen3-32B & 33.6 & 50.6 & +17.0 & 37.9 & 55.4 & +17.6 &  \\
GSM8K / R1-Qwen-7B & 16.5 & 8.2 & -8.2 & 9.4 & 19.6 & +10.1 & $\leftrightarrow$ \\
GSM8K / R1-Llama-8B & 1.3 & 0.0 & -1.3 & 5.2 & 6.9 & +1.7 & $\leftrightarrow$ \\
\midrule
MATH-500 / Qwen3-8B & 13.7 & 25.2 & +11.5 & 22.2 & 34.5 & +12.3 &  \\
MATH-500 / Qwen3-32B & 18.3 & 28.7 & +10.4 & 20.7 & 35.4 & +14.7 &  \\
MATH-500 / R1-Qwen-7B & 22.9 & 18.8 & -4.1 & 24.9 & 22.8 & -2.1 &  \\
MATH-500 / R1-Llama-8B & 21.2 & 14.9 & -6.2 & 21.3 & 15.6 & -5.7 &  \\
\midrule
MMLU-Pro / Qwen3-8B & 31.1 & 26.0 & -5.2 & 38.2 & 32.9 & -5.3 &  \\
MMLU-Pro / Qwen3-32B & 33.4 & 19.2 & -14.3 & 37.8 & 22.4 & -15.5 &  \\
MMLU-Pro / R1-Qwen-7B & 24.2 & 19.8 & -4.4 & 28.5 & 19.3 & -9.2 &  \\
MMLU-Pro / R1-Llama-8B & 30.2 & 23.9 & -6.3 & 34.3 & 25.9 & -8.4 &  \\
\midrule
AQuA / Qwen3-8B & 18.7 & 29.1 & +10.4 & 28.8 & 37.6 & +8.8 &  \\
\midrule
GPQA-D / Qwen3-8B & 4.0 & 11.0 & +7.0 & 4.7 & 11.9 & +7.2 &  \\
GPQA-D / Qwen3-32B & 4.7 & 19.2 & +14.6 & 5.5 & 12.6 & +7.1 &  \\
\bottomrule
\end{tabular}\end{adjustbox}
\end{table}

\begin{table}[!htbp]
\centering
\caption{Savings (\%) at $\alpha=0.05$ with the free response as the full-reasoning reference (Section~\ref{sec:ref}).
Natural acc.\ is the accuracy of the free response averaged over the test halves.}
\label{tab:natfull}
\smallskip
\small
\begin{adjustbox}{max width=\textwidth}\begin{tabular}{lrrrrr}
\toprule
Task / model & natural acc. & static & scalar & LearnStop & LS+cap \\
\midrule
GSM8K / Qwen3-8B & .957 & 38.1 & 22.5 & 51.2 & 47.6 \\
GSM8K / Qwen3-32B & .966 & 37.9 & 42.0 & 55.4 & 54.4 \\
GSM8K / R1-Qwen-7B & .890 & 9.4 & 15.7 & 19.6 & 18.7 \\
GSM8K / R1-Llama-8B & .790 & 5.2 & 4.2 & 6.9 & 5.7 \\
\midrule
MATH-500 / Qwen3-8B & .916 & 22.2 & 10.5 & 34.5 & 26.9 \\
MATH-500 / Qwen3-32B & .926 & 20.7 & 11.7 & 35.4 & 31.3 \\
MATH-500 / R1-Qwen-7B & .890 & 24.9 & 11.5 & 22.8 & 22.8 \\
MATH-500 / R1-Llama-8B & .783 & 21.3 & 7.6 & 15.6 & 11.3 \\
\midrule
MMLU-Pro / Qwen3-8B & .737 & 38.2 & 19.6 & 32.9 & 36.6 \\
MMLU-Pro / Qwen3-32B & .803 & 37.8 & 27.8 & 22.4 & 27.9 \\
MMLU-Pro / R1-Qwen-7B & .577 & 28.5 & 15.9 & 19.3 & 19.1 \\
MMLU-Pro / R1-Llama-8B & .593 & 34.3 & 16.9 & 25.9 & 26.4 \\
\midrule
AQuA / Qwen3-8B & .861 & 28.8 & 12.6 & 37.6 & 31.3 \\
\midrule
GPQA-D / Qwen3-8B & .618 & 4.7 & 2.8 & 11.9 & 6.0 \\
GPQA-D / Qwen3-32B & .621 & 5.5 & 6.8 & 12.6 & 5.7 \\
\bottomrule
\end{tabular}\end{adjustbox}
\end{table}

\section{Fine-tuned comparison across risk targets}
\label{app:adaptthink}

\begin{figure}[!htbp]
\centering
\includegraphics[width=\linewidth]{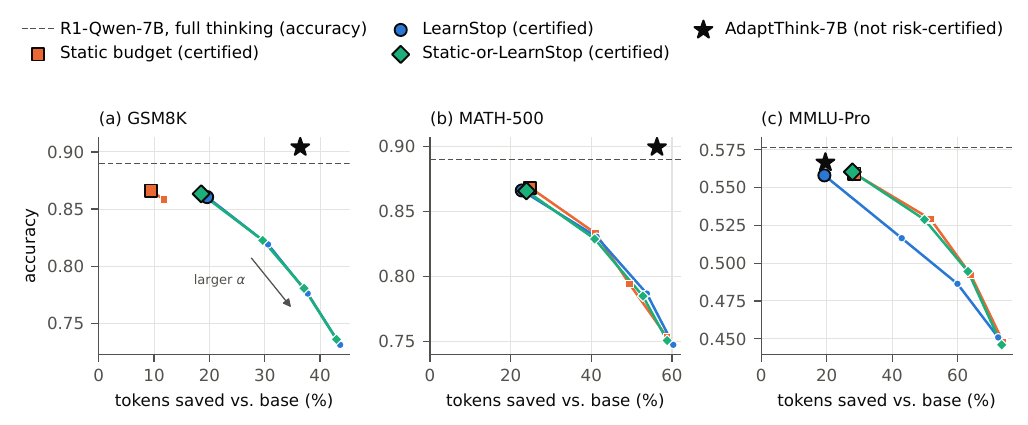}
\caption{At $\alpha=0.05$, AdaptThink-7B saves more tokens than the evaluated certified exits on GSM8K and MATH-500,
while maintaining or improving the base model's observed accuracy; on MMLU-Pro it saves less. At larger $\alpha$ the
exits save more, at lower accuracy. AdaptThink is not risk-certified, so the comparison does not hold risk fixed.
Accuracy against tokens saved relative to R1-Qwen-7B (free-response reference). Lines trace each exit over
$\alpha\in\{0.05,0.10,0.15,0.20\}$; outlined markers are $\alpha=0.05$ and the arrow in (a) points toward larger
$\alpha$. Dashed line: accuracy of the base model thinking fully.}
\label{fig:adaptthink}
\end{figure}

Figure~\ref{fig:adaptthink} follows the comparison of Section~\ref{sec:adaptthink} over the four risk targets.

\section{The static budget's grid}
\label{app:grid}

The static budget probes once per question, so its candidate budgets need not coincide with the checkpoints at which
the probing policies look. Table~\ref{tab:grid} compares the 14 probing checkpoints with the fine grid of the main
results (every 128 tokens up to 1{,}024 and every 256 after that). With the coarse grid the static budget often stops
well below $\alpha$, because the next shorter budget fails certification.

\begin{table}[!htbp]
\centering
\caption{Static-budget grid at $\alpha=0.05$, KV-cache fork: saving and mean realised test risk of the static budget on
the 14 checkpoints and on the fine grid, realised risk of \LS, and $\Delta$ against each static budget. Mean over 50
splits.}
\label{tab:grid}
\smallskip
\small
\begin{adjustbox}{max width=\textwidth}\begin{tabular}{lrrrrrrr}
\toprule
 & \multicolumn{2}{c}{static saving (\%)} & \multicolumn{3}{c}{realised test risk (\%)} & \multicolumn{2}{c}{$\Delta$ (pp)} \\
\cmidrule(lr){2-3}\cmidrule(lr){4-6}\cmidrule(lr){7-8}
Task / model & 14 points & fine & static, 14 & static, fine & \LS & vs.\ 14 & vs.\ fine \\
\midrule
GSM8K / Qwen3-8B & 30.7 & 34.1 & 2.8 & 3.2 & 3.6 & +15.0 & +11.7 \\
GSM8K / Qwen3-32B & 31.5 & 33.6 & 3.1 & 3.4 & 3.4 & +19.1 & +17.0 \\
GSM8K / R1-Qwen-7B & 16.5 & 16.5 & 2.8 & 2.8 & 3.4 & -8.2 & -8.2 \\
GSM8K / R1-Llama-8B & 1.3 & 1.3 & 1.4 & 1.4 & 0.0 & -1.3 & -1.3 \\
\midrule
MATH-500 / Qwen3-8B & 12.2 & 13.7 & 2.2 & 2.5 & 2.9 & +12.9 & +11.5 \\
MATH-500 / Qwen3-32B & 16.7 & 18.3 & 1.9 & 2.2 & 2.6 & +12.0 & +10.4 \\
MATH-500 / R1-Qwen-7B & 19.6 & 22.9 & 2.2 & 2.6 & 2.5 & -0.7 & -4.1 \\
MATH-500 / R1-Llama-8B & 16.9 & 21.2 & 2.1 & 2.8 & 2.3 & -1.9 & -6.2 \\
\midrule
MMLU-Pro / Qwen3-8B & 28.5 & 31.1 & 2.6 & 3.1 & 3.1 & -2.6 & -5.2 \\
MMLU-Pro / Qwen3-32B & 30.6 & 33.4 & 2.2 & 3.3 & 3.1 & -11.4 & -14.3 \\
MMLU-Pro / R1-Qwen-7B & 19.1 & 24.2 & 2.3 & 3.3 & 3.0 & +0.7 & -4.4 \\
MMLU-Pro / R1-Llama-8B & 25.3 & 30.2 & 2.7 & 3.3 & 3.2 & -1.4 & -6.3 \\
\midrule
AQuA / Qwen3-8B & 11.7 & 18.7 & 1.6 & 2.1 & 2.5 & +17.4 & +10.4 \\
\midrule
GPQA-D / Qwen3-8B & 2.6 & 4.0 & 1.4 & 1.8 & 1.1 & +8.4 & +7.0 \\
GPQA-D / Qwen3-32B & 3.8 & 4.7 & 1.1 & 1.0 & 1.8 & +15.5 & +14.6 \\
\bottomrule
\end{tabular}\end{adjustbox}
\end{table}

\section{Savings against the probe price}
\label{app:breakeven}

\begin{figure}[!htbp]
\centering
\includegraphics[width=\linewidth]{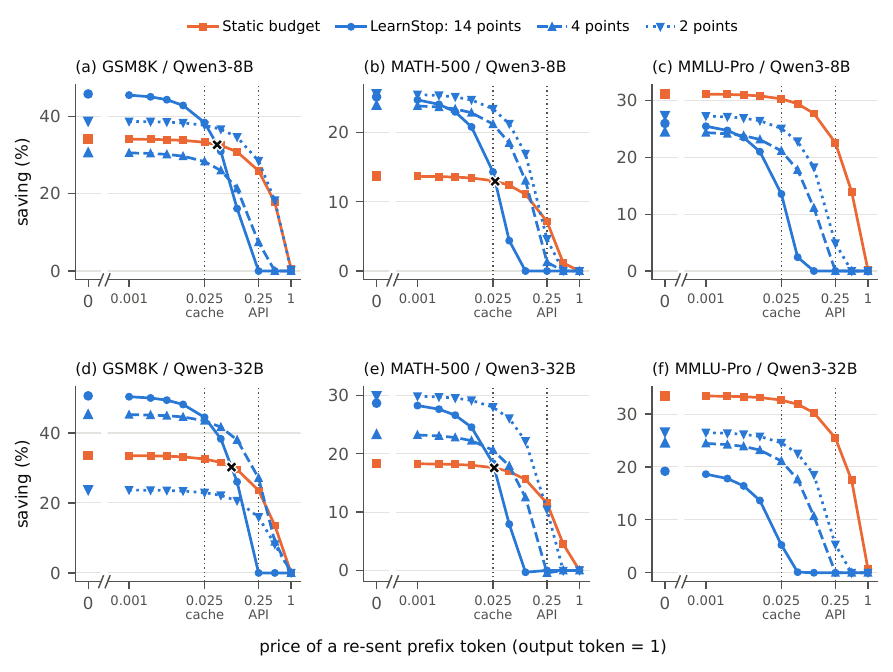}
\caption{Fewer checkpoints make \LS's saving fall more slowly as re-sent prefix tokens get more expensive. Savings at
$\alpha=0.05$ against the price of a re-sent prefix token (output token = 1; Equation~\ref{eq:cost}), each policy
re-certified and re-selected at every price. The left stub of each panel is $p=0$ (KV-cache fork), drawn apart from the
log axis. Dotted lines mark two example prices: 0.025 (a prefix cache billing cached input at a tenth of an input price
of 0.25; tick label ``cache'') and 0.25 (uncached input, as through a black-box API; tick label ``API''). \LS{} on the 14-point schedule (all
budgets), a 4-point schedule (0, 512, 2{,}048, 8{,}192) and a 2-point schedule (1{,}024, 4{,}096); the static budget
probes once. Crosses: break-even price of the 14-point schedule.}
\label{fig:price}
\end{figure}

\section{Certification procedure}
\label{app:cert}

With Hoeffding's inequality and a union bound together, no candidate is certified in any split and \LS{} saves
\certLSHU\%.

\begin{table}[!htbp]
\centering
\caption{Certification procedure: tail bound $\times$ multiplicity correction. Cert.: share of (setting, split) pairs in
which some candidate other than full reasoning is certified. Dep.: share in which the deployed candidate is not full
reasoning, that is, a certified candidate is cheaper on the calibration half. Saving: mean over settings. Same
candidates, same splits, KV-cache fork.}
\label{tab:cert}
\smallskip
\small
\begin{adjustbox}{max width=\textwidth}\begin{tabular}{llrrrrrr}
\toprule
 & & \multicolumn{3}{c}{LearnStop} & \multicolumn{3}{c}{static budget} \\
\cmidrule(lr){3-5}\cmidrule(lr){6-8}
$\alpha$ & tail bound, multiplicity & cert. & dep. & saving & cert. & dep. & saving \\
\midrule
0.05 & binomial, fixed sequence (ours) & 100 & 93 & 22.7 & 100 & 100 & 20.5 \\
0.05 & binomial, union bound & 87 & 80 & 13.8 & 87 & 87 & 14.8 \\
0.05 & Hoeffding, fixed sequence & 27 & 11 & 0.9 & 27 & 25 & 1.1 \\
0.05 & Hoeffding, union bound & 0 & 0 & 0.0 & 0 & 0 & 0.0 \\
\midrule
0.10 & binomial, fixed sequence (ours) & 100 & 100 & 39.8 & 100 & 100 & 36.7 \\
0.10 & binomial, union bound & 100 & 100 & 31.0 & 100 & 100 & 29.1 \\
0.10 & Hoeffding, fixed sequence & 87 & 83 & 22.5 & 87 & 87 & 21.7 \\
0.10 & Hoeffding, union bound & 53 & 46 & 8.5 & 53 & 53 & 10.3 \\
\midrule
0.15 & binomial, fixed sequence (ours) & 100 & 100 & 50.6 & 100 & 100 & 48.3 \\
0.15 & binomial, union bound & 100 & 100 & 43.8 & 100 & 100 & 42.4 \\
0.15 & Hoeffding, fixed sequence & 100 & 100 & 41.9 & 100 & 100 & 38.4 \\
0.15 & Hoeffding, union bound & 87 & 87 & 27.0 & 87 & 87 & 27.3 \\
\midrule
0.20 & binomial, fixed sequence (ours) & 100 & 100 & 59.2 & 100 & 100 & 56.8 \\
0.20 & binomial, union bound & 100 & 100 & 52.9 & 100 & 100 & 50.9 \\
0.20 & Hoeffding, fixed sequence & 100 & 100 & 53.4 & 100 & 100 & 50.3 \\
0.20 & Hoeffding, union bound & 100 & 97 & 42.6 & 98 & 98 & 39.8 \\
\bottomrule
\end{tabular}\end{adjustbox}
\end{table}

\section{Calibration-set size}
\label{app:calfrac}

\begin{table}[!htbp]
\centering
\caption{Savings (\%) at $\alpha=0.05$ when 30\%, 50\% or 70\% of the evaluation set is used for calibration. The
50 splits of this table are drawn separately from those of Table~\ref{tab:main}, so its 50\% columns differ slightly
from that table.}
\smallskip
\small
\begin{adjustbox}{max width=\textwidth}\begin{tabular}{lrrrrrr}
\toprule
 & \multicolumn{2}{c}{30\% calibration} & \multicolumn{2}{c}{50\% calibration} & \multicolumn{2}{c}{70\% calibration} \\
\cmidrule(lr){2-3}\cmidrule(lr){4-5}\cmidrule(lr){6-7}
Task / model & static & LS & static & LS & static & LS \\
\midrule
GSM8K / Qwen3-8B & 31.9 & 43.7 & 35.0 & 45.4 & 35.2 & 47.1 \\
GSM8K / Qwen3-32B & 30.4 & 49.6 & 32.5 & 50.8 & 34.0 & 51.8 \\
GSM8K / R1-Qwen-7B & 15.2 & 7.2 & 16.1 & 8.0 & 17.6 & 8.9 \\
GSM8K / R1-Llama-8B & 1.4 & 0.0 & 1.4 & 0.0 & 1.3 & 0.0 \\
\midrule
MATH-500 / Qwen3-8B & 10.7 & 16.7 & 15.9 & 23.8 & 16.3 & 26.6 \\
MATH-500 / Qwen3-32B & 14.7 & 25.2 & 19.7 & 28.6 & 20.5 & 30.5 \\
MATH-500 / R1-Qwen-7B & 15.2 & 14.4 & 22.6 & 17.3 & 26.7 & 19.3 \\
MATH-500 / R1-Llama-8B & 17.9 & 10.2 & 22.6 & 15.5 & 24.4 & 17.7 \\
\midrule
MMLU-Pro / Qwen3-8B & 27.2 & 22.7 & 30.4 & 25.3 & 32.3 & 27.1 \\
MMLU-Pro / Qwen3-32B & 30.6 & 15.8 & 32.7 & 19.1 & 33.3 & 21.6 \\
MMLU-Pro / R1-Qwen-7B & 20.3 & 17.3 & 23.9 & 20.3 & 25.9 & 22.2 \\
MMLU-Pro / R1-Llama-8B & 27.5 & 22.6 & 29.2 & 23.2 & 32.0 & 25.6 \\
\midrule
AQuA / Qwen3-8B & 16.4 & 24.2 & 20.1 & 29.2 & 22.2 & 32.5 \\
\midrule
GPQA-D / Qwen3-8B & 2.8 & 10.8 & 4.8 & 10.3 & 5.9 & 10.2 \\
GPQA-D / Qwen3-32B & 4.0 & 14.0 & 5.9 & 19.0 & 7.9 & 23.3 \\
\bottomrule
\end{tabular}\end{adjustbox}
\end{table}

\section{Feature ablation}
\label{app:features}

Removing the confidence features raises the saving of \LS{} on MMLU-Pro for all four models, by
\featConfMmluMin--\featConfMmluMax{} points.

\begin{table}[!htbp]
\centering
\caption{\LS{} savings (\%) at $\alpha=0.05$, KV-cache fork, with one feature group removed or kept alone. Budget: log
budget, checkpoint index. Confidence: mean log-probability and mean entropy of the probe tokens. Stability: repeat indicator, run length, vote share.
Markers: backtracking-marker density.}
\smallskip
\footnotesize\setlength{\tabcolsep}{3.5pt}
\begin{adjustbox}{max width=\textwidth}\begin{tabular}{lrrrrrrrrr}
\toprule
 & & \multicolumn{4}{c}{one group removed} & \multicolumn{4}{c}{one group alone} \\
\cmidrule(lr){3-6}\cmidrule(lr){7-10}
Task / model & all & $-$budget & $-$conf. & $-$stab. & $-$markers & budget & conf. & stab. & markers \\
\midrule
GSM8K / Qwen3-8B & 45.7 & 35.8 & 38.2 & 44.6 & 46.0 & 28.3 & 16.0 & 27.8 & 3.7 \\
GSM8K / Qwen3-32B & 50.6 & 47.6 & 39.2 & 53.2 & 50.7 & 29.2 & 41.5 & 33.1 & 5.4 \\
GSM8K / R1-Qwen-7B & 8.2 & 6.0 & 4.7 & 8.1 & 8.4 & 10.1 & 4.8 & 5.2 & 13.2 \\
GSM8K / R1-Llama-8B & 0.0 & 0.0 & 0.0 & 0.0 & 0.0 & 0.0 & 0.0 & 0.0 & 0.0 \\
\midrule
MATH-500 / Qwen3-8B & 25.2 & 20.6 & 15.7 & 21.3 & 24.5 & 9.8 & 11.7 & 12.9 & 1.7 \\
MATH-500 / Qwen3-32B & 28.7 & 23.7 & 18.2 & 32.2 & 27.8 & 14.3 & 18.8 & 15.0 & 1.0 \\
MATH-500 / R1-Qwen-7B & 18.8 & 17.7 & 18.1 & 15.2 & 18.5 & 16.8 & 13.6 & 7.5 & 2.1 \\
MATH-500 / R1-Llama-8B & 14.9 & 12.8 & 14.0 & 21.1 & 14.9 & 14.5 & 11.9 & 9.1 & 3.9 \\
\midrule
MMLU-Pro / Qwen3-8B & 26.0 & 21.4 & 27.1 & 26.3 & 27.0 & 27.3 & 12.1 & 25.6 & 10.5 \\
MMLU-Pro / Qwen3-32B & 19.2 & 22.4 & 22.9 & 26.4 & 20.2 & 29.1 & 20.9 & 28.1 & 8.3 \\
MMLU-Pro / R1-Qwen-7B & 19.8 & 16.8 & 27.0 & 15.6 & 19.7 & 17.7 & 13.4 & 15.1 & 8.7 \\
MMLU-Pro / R1-Llama-8B & 23.9 & 19.1 & 25.4 & 24.2 & 25.7 & 24.0 & 8.7 & 20.0 & 6.1 \\
\midrule
AQuA / Qwen3-8B & 29.1 & 18.3 & 27.6 & 17.7 & 29.6 & 10.3 & 5.8 & 15.3 & 4.3 \\
\midrule
GPQA-D / Qwen3-8B & 11.0 & 5.8 & 6.5 & 5.0 & 12.0 & 2.3 & 5.5 & 5.4 & 1.6 \\
GPQA-D / Qwen3-32B & 19.2 & 15.8 & 13.0 & 9.2 & 17.9 & 3.4 & 5.0 & 8.3 & 1.1 \\
\bottomrule
\end{tabular}\end{adjustbox}
\end{table}

\section{Other classifiers and shared training pools}
\label{app:learners}

Training \LS{} on the pooled training questions of all tasks and calibrating the threshold per task keeps savings within
3 points of the rule trained on the task itself in \poolWithin{} of \nSettings{} settings; training only on the other tasks does so in
\lotoWithin. The pooled rule loses most on \poolLossText{} and gains most on \poolGainText. A single guarantee over the task mixture, rather than one per task, is a different target that one
pooled calibration set can serve.

\begin{table}[!htbp]
\centering
\caption{\LS{} variants at $\alpha=0.05$: logistic regression or gradient-boosted trees trained on the task's own training
pool; logistic regression trained on the pooled training questions of all tasks for the same model, or on all tasks
except the target. Thresholds are always calibrated on the target task.}
\smallskip
\small
\begin{adjustbox}{max width=\textwidth}\begin{tabular}{lrrrr}
\toprule
 & \multicolumn{2}{c}{in-task training pool} & \multicolumn{2}{c}{other tasks' pools} \\
\cmidrule(lr){2-3}\cmidrule(lr){4-5}
Task / model & logistic & boosted trees & pooled (all tasks) & leave-task-out \\
\midrule
GSM8K / Qwen3-8B & 45.7 & 48.1 & 44.5 & 44.0 \\
GSM8K / Qwen3-32B & 50.6 & 52.9 & 45.9 & 44.3 \\
GSM8K / R1-Qwen-7B & 8.2 & 10.5 & 12.5 & 12.7 \\
GSM8K / R1-Llama-8B & 0.0 & 0.0 & 0.0 & 0.1 \\
\midrule
MATH-500 / Qwen3-8B & 25.2 & 19.9 & 25.6 & 20.0 \\
MATH-500 / Qwen3-32B & 28.7 & 28.3 & 24.2 & 20.4 \\
MATH-500 / R1-Qwen-7B & 18.8 & 25.6 & 19.0 & 19.9 \\
MATH-500 / R1-Llama-8B & 14.9 & 16.5 & 12.1 & 16.3 \\
\midrule
MMLU-Pro / Qwen3-8B & 26.0 & 27.9 & 27.6 & 27.8 \\
MMLU-Pro / Qwen3-32B & 19.2 & 21.7 & 28.1 & 28.3 \\
MMLU-Pro / R1-Qwen-7B & 19.8 & 15.9 & 22.3 & 18.1 \\
MMLU-Pro / R1-Llama-8B & 23.9 & 16.5 & 23.6 & 19.5 \\
\midrule
AQuA / Qwen3-8B & 29.1 & 35.2 & 29.6 & 25.9 \\
\midrule
GPQA-D / Qwen3-8B & 11.0 & 12.5 & 8.1 & 8.8 \\
GPQA-D / Qwen3-32B & 19.2 & 12.0 & 2.9 & 4.6 \\
\bottomrule
\end{tabular}\end{adjustbox}
\end{table}

\section{Selecting between the static budget and LearnStop}
\label{app:decomp}

\begin{table}[!htbp]
\centering
\caption{Savings (\%) at $\alpha=0.05$, KV-cache fork. The static budget over the full grid at level $\delta$, restricted to
the five caps of \LSC{} at $\delta$, and at $\delta/6$ (the level of each \LSC{} family); \LSC; \LS; and \SLS{} (both
families at $\delta/2$).}
\smallskip
\small
\begin{adjustbox}{max width=\textwidth}\begin{tabular}{lrrrrrr}
\toprule
 & \multicolumn{3}{c}{static budget} & & & \\
\cmidrule(lr){2-4}
Task / model & full grid, $\delta$ & 5 caps, $\delta$ & 5 caps, $\delta/6$ & LS+cap & LS & st./LS \\
\midrule
GSM8K / Qwen3-8B & 34.1 & 24.9 & 21.3 & 42.3 & 45.7 & 44.1 \\
GSM8K / Qwen3-32B & 33.6 & 23.7 & 23.7 & 49.8 & 50.6 & 49.8 \\
GSM8K / R1-Qwen-7B & 16.5 & 11.2 & 11.2 & 11.1 & 8.2 & 15.4 \\
GSM8K / R1-Llama-8B & 1.3 & 0.0 & 0.0 & 0.0 & 0.0 & 1.3 \\
\midrule
MATH-500 / Qwen3-8B & 13.7 & 11.1 & 3.0 & 16.7 & 25.2 & 21.5 \\
MATH-500 / Qwen3-32B & 18.3 & 11.1 & 11.1 & 26.2 & 28.7 & 26.4 \\
MATH-500 / R1-Qwen-7B & 22.9 & 14.5 & 6.6 & 18.2 & 18.8 & 19.8 \\
MATH-500 / R1-Llama-8B & 21.2 & 14.3 & 3.2 & 10.5 & 14.9 & 18.9 \\
\midrule
MMLU-Pro / Qwen3-8B & 31.1 & 27.6 & 19.8 & 27.2 & 26.0 & 30.1 \\
MMLU-Pro / Qwen3-32B & 33.4 & 22.8 & 22.8 & 24.6 & 19.2 & 32.6 \\
MMLU-Pro / R1-Qwen-7B & 24.2 & 11.3 & 10.2 & 18.1 & 19.8 & 23.4 \\
MMLU-Pro / R1-Llama-8B & 30.2 & 20.4 & 12.4 & 22.4 & 23.9 & 29.2 \\
\midrule
AQuA / Qwen3-8B & 18.7 & 5.8 & 1.3 & 23.3 & 29.1 & 26.3 \\
\midrule
GPQA-D / Qwen3-8B & 4.0 & 1.0 & 0.0 & 9.5 & 11.0 & 9.6 \\
GPQA-D / Qwen3-32B & 4.7 & 6.0 & 1.6 & 8.1 & 19.2 & 7.6 \\
\bottomrule
\end{tabular}\end{adjustbox}
\end{table}

The selection can also be made at a non-zero price and over the checkpoint schedule. Table~\ref{tab:selprice} reports,
at $\alpha=0.05$, the split-averaged shortfall of three calibration-time choices: between the static budget and \LS{}
with 14 checkpoints (\SLS), among the four \LS{} schedules, and over both at once (the static family plus the four
\LS{} schedules, each family at $\delta/5$). The comparator is, in each setting, the single policy of the set that has the highest mean saving over the test
halves, as in Section~\ref{sec:select}, so the shortfall includes the cost of not knowing the winner in advance.

At $\alpha=0.15$ with a KV-cache fork, \SLS{} saves \uniSLB\% on average over the settings, the static budget used
everywhere \uniFixB\% and \LS{} used everywhere \uniLSB\%. The two restrictions of \LSC{} can be measured on the
static budget alone: limiting it to the five cap lengths lowers its saving by \decCoverage{} points on average, and
testing it at $\delta/6$ instead of $\delta$, as each of the six families of \LSC{} is tested, lowers it by a
further \decPower{} points.

\begin{table}[!htbp]
\centering
\caption{Calibration selection gap (pp) at $\alpha=0.05$: in each setting, the policy of the set with the highest mean saving over the test halves, minus the policy
selected on each calibration half; mean and maximum over the 15 settings.}
\label{tab:selprice}
\smallskip
\small
\begin{adjustbox}{max width=\textwidth}\begin{tabular}{lrrrrrr}
\toprule
 & \multicolumn{2}{c}{static vs.\ \LS{} (14 ckpts)} & \multicolumn{2}{c}{among \LS{} schedules} & \multicolumn{2}{c}{both choices} \\
\cmidrule(lr){2-3}\cmidrule(lr){4-5}\cmidrule(lr){6-7}
price $p$ & mean & max & mean & max & mean & max \\
\midrule
0 & 2.3 & 11.6 & 2.7 & 8.3 & 3.4 & 8.5 \\
0.025 & 1.7 & 5.2 & 3.0 & 6.2 & 3.4 & 6.0 \\
0.25 & 1.3 & 3.8 & 1.7 & 4.7 & 2.7 & 6.3 \\
\bottomrule
\end{tabular}\end{adjustbox}
\end{table}

\section{Serving another workload: other models}
\label{app:shift}

Table~\ref{tab:shift} gives the realised risk of every pair for Qwen3-8B and Table~\ref{tab:shiftmore} for the other
three models.
Table~\ref{tab:recal} gives, per model, the realised risk and the saving on the served
tasks without recalibration, after recalibration and, for \LS, after retraining on the served task. With R1-Llama-8B, \LS{}
calibrated on GSM8K does not probe (Table~\ref{tab:main}), so it carries no risk to another task. In the example of
Section~\ref{sec:shift}, the policies calibrated on GSM8K with Qwen3-8B cut \shiftExFixSave\% (static budget) and
\shiftExLSSave\% (\LS) of the tokens on MATH-500.

With 100 calibration questions of the served task the recalibrated static budget saves \recalSizeHundredFix\% on
average and \LS{} with the old classifier \recalSizeHundredLS\%, over the same \recalSizePairs{} pairs as the
figures for 200 questions in Section~\ref{sec:shift}. When $\delta$ is shared among five families, as when the
checkpoint schedule is selected as well, a candidate without any calibration loss needs at least \recalMinNFive{}
calibration questions to be certified.

\begin{table}[!htbp]
\centering
\caption{Realised lost-correct risk (\%) with Qwen3-8B when a policy certified at $\alpha=0.05$ on one task (rows)
is served unchanged on another (columns), with a KV-cache fork; means over the 50 calibration splits of the row
task. Off the diagonal the risk is measured on all evaluation questions of the column task. Italic: the calibrated
task itself, measured on its held-out test halves. Bold: above the 5\% target.}
\label{tab:shift}
\smallskip
\footnotesize\setlength{\tabcolsep}{3.4pt}
\begin{adjustbox}{max width=\textwidth}\begin{tabular}{lrrrrrrrrrr}
\toprule
 & \multicolumn{5}{c}{static budget, deployed on} & \multicolumn{5}{c}{\LS, deployed on} \\
\cmidrule(lr){2-6}\cmidrule(lr){7-11}
Calibrated on & GSM8K & MATH-500 & MMLU-Pro & AQuA & GPQA-D & GSM8K & MATH-500 & MMLU-Pro & AQuA & GPQA-D \\
\midrule
GSM8K & \textit{3.2} & \textbf{22.6} & \textbf{13.0} & \textbf{7.9} & \textbf{26.3} & \textit{3.6} & \textbf{21.6} & \textbf{15.2} & \textbf{17.8} & \textbf{27.4} \\
MATH-500 & 0.0 & \textit{2.5} & 1.2 & 0.1 & 4.0 & 0.3 & \textit{2.9} & 1.7 & 0.7 & 4.2 \\
MMLU-Pro & 0.5 & \textbf{6.0} & \textit{3.1} & 2.8 & \textbf{9.8} & 1.5 & \textbf{8.2} & \textit{3.1} & 1.6 & \textbf{5.3} \\
AQuA & 0.4 & \textbf{5.5} & 2.8 & \textit{2.1} & \textbf{9.3} & 1.1 & \textbf{9.3} & 3.4 & \textit{2.5} & \textbf{6.6} \\
GPQA-D & 0.0 & 0.6 & 0.2 & 0.0 & \textit{1.8} & 1.6 & \textbf{5.8} & 1.8 & 1.8 & \textit{1.1} \\
\bottomrule
\end{tabular}\end{adjustbox}
\end{table}

\begin{table}[!htbp]
\centering
\caption{Realised lost-correct risk / saving (both \%) on the served task, $\alpha=0.05$, averaged over the ordered
pairs of a calibration task and another served task (number in parentheses) and over 50 splits of the served task.
Unchanged: the policy certified on the calibration task; recalibrated: certification and selection repeated on the
served task's calibration half (\LS{} keeps its classifier); retrained: \LS{} also retrained on the served task.}
\label{tab:recal}
\smallskip
\footnotesize\setlength{\tabcolsep}{3.6pt}
\begin{adjustbox}{max width=\textwidth}\begin{tabular}{lrrrrr}
\toprule
 & \multicolumn{2}{c}{static budget} & \multicolumn{3}{c}{\LS} \\
\cmidrule(lr){2-3}\cmidrule(lr){4-6}
Model (pairs) & unchanged & recalibrated & unchanged & recalibrated & retrained \\
\midrule
all pairs (44) & 8.6 / 31.1 & 2.6 / 20.7 & 7.1 / 30.7 & 2.3 / 19.1 & 2.6 / 24.4 \\
\midrule
Qwen3-8B (20) & 5.7 / 26.7 & 2.5 / 20.3 & 6.9 / 35.3 & 2.5 / 21.7 & 2.6 / 27.4 \\
Qwen3-32B (12) & 8.1 / 31.3 & 2.5 / 22.5 & 8.1 / 35.0 & 2.4 / 25.5 & 2.7 / 29.4 \\
R1-Qwen-7B (6) & 14.8 / 37.7 & 2.9 / 21.2 & 11.3 / 29.4 & 2.4 / 11.8 & 3.0 / 15.6 \\
R1-Llama-8B (6) & 13.1 / 39.0 & 2.5 / 17.6 & 1.5 / 7.7 & 0.8 / 4.7 & 1.8 / 12.9 \\
\bottomrule
\end{tabular}\end{adjustbox}
\end{table}

\begin{table}[!htbp]
\centering
\caption{As Table~\ref{tab:shift}, for Qwen3-32B, R1-Qwen-7B and R1-Llama-8B: realised lost-correct risk (\%) when a
policy certified at $\alpha=0.05$ on the row task is served unchanged on the column task. Italic: the calibrated
task; bold: above the 5\% target.}
\label{tab:shiftmore}
\smallskip
\footnotesize\setlength{\tabcolsep}{3.4pt}
\begin{tabular}{lrrrrrrrr}
\toprule
\multicolumn{9}{l}{\textit{Qwen3-32B}} \\
 & \multicolumn{4}{c}{static budget, deployed on} & \multicolumn{4}{c}{\LS, deployed on} \\
\cmidrule(lr){2-5}\cmidrule(lr){6-9}
Calibrated on & GSM8K & MATH-500 & MMLU-Pro & GPQA-D & GSM8K & MATH-500 & MMLU-Pro & GPQA-D \\
\midrule
GSM8K & \textit{3.4} & \textbf{29.8} & \textbf{14.4} & \textbf{27.0} & \textit{3.4} & \textbf{28.8} & \textbf{17.2} & \textbf{26.3} \\
MATH-500 & 0.0 & \textit{2.2} & 0.7 & 4.1 & 0.2 & \textit{2.6} & 1.8 & \textbf{7.0} \\
MMLU-Pro & 0.4 & \textbf{8.1} & \textit{3.3} & \textbf{11.8} & 0.4 & \textbf{7.1} & \textit{3.1} & \textbf{5.1} \\
GPQA-D & 0.0 & 0.4 & 0.1 & \textit{1.0} & 0.3 & 2.2 & 0.9 & \textit{1.8} \\
\bottomrule
\end{tabular}

\medskip
\begin{tabular}{lrrrrrr}
\toprule
\multicolumn{7}{l}{\textit{R1-Qwen-7B}} \\
 & \multicolumn{3}{c}{static budget, deployed on} & \multicolumn{3}{c}{\LS, deployed on} \\
\cmidrule(lr){2-4}\cmidrule(lr){5-7}
Calibrated on & GSM8K & MATH-500 & MMLU-Pro & GSM8K & MATH-500 & MMLU-Pro \\
\midrule
GSM8K & \textit{2.8} & \textbf{50.8} & \textbf{31.7} & \textit{3.4} & \textbf{45.0} & \textbf{17.4} \\
MATH-500 & 0.0 & \textit{2.6} & 2.5 & 0.0 & \textit{2.5} & 2.3 \\
MMLU-Pro & 0.0 & 3.0 & \textit{3.3} & 0.0 & 2.7 & \textit{3.0} \\
\bottomrule
\end{tabular}

\medskip
\begin{tabular}{lrrrrrr}
\toprule
\multicolumn{7}{l}{\textit{R1-Llama-8B}} \\
 & \multicolumn{3}{c}{static budget, deployed on} & \multicolumn{3}{c}{\LS, deployed on} \\
\cmidrule(lr){2-4}\cmidrule(lr){5-7}
Calibrated on & GSM8K & MATH-500 & MMLU-Pro & GSM8K & MATH-500 & MMLU-Pro \\
\midrule
GSM8K & \textit{1.4} & \textbf{37.8} & \textbf{32.8} & \textit{0.0} & 0.0 & 0.0 \\
MATH-500 & 0.0 & \textit{2.8} & 1.1 & 0.0 & \textit{2.3} & 1.9 \\
MMLU-Pro & 0.0 & \textbf{5.8} & \textit{3.3} & 0.0 & \textbf{7.1} & \textit{3.2} \\
\bottomrule
\end{tabular}
\end{table}

\section{MMLU-Pro by subject}
\label{app:category}

\begin{table}[!htbp]
\centering
\caption{Savings (\%) at $\alpha=0.05$ within MMLU-Pro subject groups; calibration and test splits are drawn within each
group.}
\smallskip
\small
\begin{adjustbox}{max width=\textwidth}\begin{tabular}{llrrrrr}
\toprule
Model & MMLU-Pro categories & $n$ & static & scalar & LearnStop & LS+cap \\
\midrule
Qwen3-8B & math, physics, chemistry, engineering & 399 & 11.4 & 5.8 & 13.5 & 5.3 \\
Qwen3-8B & other 10 categories & 601 & 25.7 & 23.1 & 22.8 & 20.2 \\
Qwen3-32B & math, physics, chemistry, engineering & 399 & 27.0 & 18.0 & 5.4 & 7.1 \\
Qwen3-32B & other 10 categories & 601 & 22.3 & 30.1 & 30.2 & 27.3 \\
R1-Qwen-7B & math, physics, chemistry, engineering & 399 & 6.1 & 7.5 & 5.4 & 2.2 \\
R1-Qwen-7B & other 10 categories & 601 & 32.6 & 17.5 & 23.0 & 24.2 \\
R1-Llama-8B & math, physics, chemistry, engineering & 399 & 17.5 & 11.1 & 8.9 & 4.5 \\
R1-Llama-8B & other 10 categories & 601 & 30.3 & 20.0 & 29.8 & 25.5 \\
\bottomrule
\end{tabular}\end{adjustbox}
\end{table}

\section{Trajectory decomposition}
\label{app:traj}

Each question's probe correctness sequence ends with the end probe. Types are assigned in this order: \emph{early} if the
first probe and the end probe are correct (whatever happens in between); otherwise, if the end probe is correct,
\emph{beneficial} when correctness flips at most once and \emph{oscillating} when it flips more than once; if the end
probe is wrong, \emph{harmful} when some probe is correct and \emph{unsolved} when none is. Unsolved questions contribute
no lost-correct risk. For the questions that full reasoning answers correctly, the answer \emph{settles} at the earliest
checkpoint after which every probe is correct, or at the end of the trace when the probe at the last checkpoint before the end is wrong. The sequence uses the 14
checkpoints.

\begin{table}[!htbp]
\centering
\caption{Trajectory types, the median and 90th percentile of the settling point (thinking tokens), probe-to-thinking
token ratio when probing at all 14 checkpoints, and the policy with the largest saving at $\alpha=0.05$ (\looWinners).}
\label{tab:traj}
\smallskip
\footnotesize\setlength{\tabcolsep}{3.4pt}
\begin{adjustbox}{max width=\textwidth}\begin{tabular}{lrrrrrrrrl}
\toprule
 & \multicolumn{5}{c}{trajectory type (\% of questions)} & \multicolumn{2}{c}{settles by} & probe/think & winner \\
\cmidrule(lr){2-6}\cmidrule(lr){7-8}
Task / model & early & benef. & oscill. & harmful & unsolved & median & p90 & tokens (\%) & at $\alpha=0.05$ \\
\midrule
GSM8K / Qwen3-8B & 26.2 & 64.8 & 4.9 & 1.6 & 2.5 & 384 & 768 & 2.7 & LearnStop \\
GSM8K / Qwen3-32B & 40.1 & 52.4 & 4.3 & 0.9 & 2.3 & 256 & 512 & 2.8 & LearnStop \\
GSM8K / R1-Qwen-7B & 22.4 & 60.4 & 0.3 & 2.3 & 14.6 & 128 & 228 & 7.2 & static \\
GSM8K / R1-Llama-8B & 6.1 & 60.0 & 0.0 & 1.7 & 32.2 & 128 & 214 & 5.6 & static \\
\midrule
MATH-500 / Qwen3-8B & 26.4 & 55.8 & 9.8 & 2.8 & 5.2 & 768 & 3,336 & 2.5 & LearnStop \\
MATH-500 / Qwen3-32B & 34.6 & 50.6 & 8.4 & 1.6 & 4.8 & 512 & 3,072 & 2.4 & LearnStop \\
MATH-500 / R1-Qwen-7B & 26.4 & 51.6 & 10.2 & 2.2 & 9.6 & 768 & 3,072 & 2.8 & static \\
MATH-500 / R1-Llama-8B & 10.2 & 57.6 & 4.6 & 4.0 & 23.6 & 384 & 4,024 & 2.4 & static \\
\midrule
MMLU-Pro / Qwen3-8B & 44.4 & 28.3 & 3.2 & 6.8 & 17.3 & 0 & 3,072 & 1.3 & static \\
MMLU-Pro / Qwen3-32B & 52.3 & 25.6 & 4.4 & 5.4 & 12.3 & 0 & 2,048 & 1.5 & static \\
MMLU-Pro / R1-Qwen-7B & 22.8 & 27.8 & 8.2 & 13.0 & 28.2 & 512 & 4,096 & 1.5 & static \\
MMLU-Pro / R1-Llama-8B & 21.6 & 34.4 & 4.4 & 12.9 & 26.7 & 512 & 3,072 & 1.4 & static \\
\midrule
AQuA / Qwen3-8B & 43.3 & 42.1 & 4.5 & 3.3 & 6.7 & 384 & 1,536 & 1.4 & LearnStop \\
\midrule
GPQA-D / Qwen3-8B & 30.3 & 24.7 & 6.1 & 12.6 & 26.3 & 1,536 & 7,524 & 0.8 & LearnStop \\
GPQA-D / Qwen3-32B & 31.8 & 24.2 & 6.6 & 14.1 & 23.2 & 1,024 & 5,835 & 0.9 & LearnStop \\
\bottomrule
\end{tabular}\end{adjustbox}
\end{table}

\section{Free-form BBH}
\label{app:bbh}

On the BBH \citep{suzgun2023bbh} dyck-language and word-sorting tasks with Qwen3-8B (150 evaluation questions
each), the free response has accuracy \bbhNat{} and the end probe \bbhEnd; \bbhCapHit\% of probes reach the 48-token cap because word lists are long. The
end-probe reference is therefore unreliable for this task and we exclude it from the main comparison.

\section{Trace-sampling seeds}
\label{app:seeds}

We re-sampled every trace of GSM8K with Qwen3-32B and MMLU-Pro with Qwen3-8B with \seedN{} seeds in total, re-trained
\LS{} on each seed's own training pool and re-certified all policies. The static budget uses
\seedGridText{} for every seed. \LS{} minus static budget ranges from
\seedGsmDmin{} to \seedGsmDmax{} points on GSM8K and from \seedMmluDmin{} to \seedMmluDmax{} on MMLU-Pro across seeds:
the GSM8K result is stable, while on MMLU-Pro with Qwen3-8B the sign depends on the sampled traces, with the static
budget ahead for \seedMmluStaticN{} of the \seedN{} seeds. The static budget
saves \seedMmluFixMin--\seedMmluFixMax\% on MMLU-Pro and \LS{} \seedGsmLSMin--\seedGsmLSMax\% on GSM8K depending on the
seed. \SLS{} stays within \seedSLabs{} points of the better of the two for every seed.

\begin{table}[!htbp]
\centering
\caption{Savings (\%) at $\alpha=0.05$, KV-cache fork, for independently sampled traces (seed 0 is the main run), with
the static budget on the fine grid for every seed.}
\smallskip
\small
\begin{adjustbox}{max width=\textwidth}\begin{tabular}{lrrrrrrr}
\toprule
Task / model & seed & full acc. & thinking & static & LS & st./LS & LS $-$ static \\
\midrule
GSM8K / Qwen3-32B & 0 & .968 & 932 & 33.6 & 50.6 & 49.8 & +17.0 \\
GSM8K / Qwen3-32B & 1 & .968 & 925 & 33.3 & 49.0 & 47.7 & +15.7 \\
GSM8K / Qwen3-32B & 2 & .967 & 926 & 32.4 & 51.7 & 50.7 & +19.4 \\
\midrule
MMLU-Pro / Qwen3-8B & 0 & .759 & 3,062 & 31.1 & 26.0 & 30.1 & -5.2 \\
MMLU-Pro / Qwen3-8B & 1 & .771 & 3,131 & 27.3 & 30.7 & 30.2 & +3.4 \\
MMLU-Pro / Qwen3-8B & 2 & .762 & 3,089 & 31.9 & 29.0 & 30.7 & -2.9 \\
\bottomrule
\end{tabular}\end{adjustbox}
\end{table}

\section{Online measurement: details and controls}
\label{app:latency}

The engine is vLLM 0.30.0 on one H100 GPU with prefix caching. In the main
run and in its repetition each request carries its own cache key, so that its reasoning segments and provisional
answers reuse its own cached reasoning and nothing else. Reasoning is sampled as in the main runs (temperature 0.6,
top-$p$ 0.95, top-$k$ 20) and provisional answers are decoded greedily and stopped at their first line. The policies
are full reasoning, the static budget and \LS{} with 14 and with 4 checkpoints, certified offline at $\alpha=0.05$ on
the first split; the questions are the first questions of that split's test half, the same for every policy.
Requests of all task--policy pairs are shuffled into one queue served at a concurrency of 16, and the response time
of a request runs from the start of its processing to its final answer. A pilot of 8 questions per task and policy
measured the generation rate, from which the number of questions was set to fit the time of the run: \latNMain{} per
task and policy in the main run and \latNRep{} in the repetition, which ran at the same time on another GPU of the
same type. Before measuring, each job compared the provisional answer at $B=0$ with the logged one for 4 questions per
task, and text and length agreed for all of them. In the
first control requests may reuse any cached prefix, and in the second the prefix cache is switched off, so that every
reasoning segment and provisional answer processes its prefix again (\latNCtl{} and \latNNoc{} questions per task and
policy).
Table~\ref{tab:latency} gives the tokens, times and accuracy of every policy in the main run, and
Table~\ref{tab:latency_runs} the token and time ratios of all four runs. Each policy ran on its own sampled traces,
so differences in accuracy between policies are within the sampling error of about \latAccNoise{} points.
Request by request, \LS{} with 4 checkpoints stopped where the offline analysis predicts in
\agreeLSsMin--\agreeLSsMax\% of the requests and the static budget in \agreeFixMin--\agreeFixMax\%. With a shared
cache the time per generated token of \LS{} with 14 checkpoints was \perTokCtlMathLSd{} of that of full reasoning on
MATH-500 and \perTokCtlMmluLSd{} on MMLU-Pro. With the prefix cache off, that of the static budget and of the
4-checkpoint rule stayed between \latNocFewMin{} and \latNocFewMax{} (\latNNoc{} questions per task and policy).

\begin{table}[!htbp]
\centering
\caption{Response time in a serving engine (Qwen3-8B, vLLM, one H100, concurrency 16, \latNMain{} questions per task
and policy). Tokens and time: generated tokens (reasoning and provisional answers) and mean response time relative
to full reasoning on the same questions, with a 95\% question-level bootstrap interval; median and 90th percentile
in seconds; accuracy of the returned answers. The policies were certified offline at $\alpha=0.05$ on one split.}
\label{tab:latency}
\smallskip
\footnotesize\setlength{\tabcolsep}{3.4pt}
\begin{adjustbox}{max width=\textwidth}\begin{tabular}{lrrrrrrrrrr}
\toprule
 & \multicolumn{5}{c}{MATH-500} & \multicolumn{5}{c}{MMLU-Pro} \\
\cmidrule(lr){2-6}\cmidrule(lr){7-11}
Policy & tokens & time & median (s) & p90 (s) & acc. (\%) & tokens & time & median (s) & p90 (s) & acc. (\%) \\
\midrule
full reasoning & 1.00 & 1.00 & 40 & 119 & 91.0 & 1.00 & 1.00 & 31 & 139 & 76.6 \\
static budget & 0.94 & 0.93 [0.87, 0.99] & 41 & 117 & 92.8 & 0.60 & 0.60 [0.54, 0.67] & 32 & 48 & 70.7 \\
\LS{}, 14 checkpoints & 0.80 & 0.80 [0.72, 0.87] & 31 & 86 & 89.2 & 0.65 & 0.65 [0.58, 0.73] & 21 & 63 & 72.5 \\
\LS{}, 4 checkpoints & 0.88 & 0.88 [0.81, 0.94] & 31 & 110 & 91.0 & 0.67 & 0.66 [0.59, 0.72] & 25 & 95 & 74.9 \\
\bottomrule
\end{tabular}\end{adjustbox}
\end{table}

\begin{table}[!htbp]
\centering
\caption{Generated tokens and mean response time relative to full reasoning in the main run, its repetition on
a second GPU, and the controls in which requests share cached prefixes or the prefix cache is off (questions per task
and policy in
parentheses). Qwen3-8B, concurrency 16.}
\label{tab:latency_runs}
\smallskip
\footnotesize\setlength{\tabcolsep}{3.6pt}
\begin{adjustbox}{max width=\textwidth}\begin{tabular}{llrrrrrrrr}
\toprule
 & & \multicolumn{2}{c}{main run (167)} & \multicolumn{2}{c}{second GPU (130)} & \multicolumn{2}{c}{shared cache (50)} & \multicolumn{2}{c}{no prefix cache (50)} \\
\cmidrule(lr){3-4}\cmidrule(lr){5-6}\cmidrule(lr){7-8}\cmidrule(lr){9-10}
Task & Policy & tokens & time & tokens & time & tokens & time & tokens & time \\
\midrule
MATH-500 & static budget & 0.94 & 0.93 & 0.91 & 0.91 & 0.87 & 0.88 & 0.90 & 0.91 \\
MATH-500 & \LS{}, 14 checkpoints & 0.80 & 0.80 & 0.73 & 0.73 & 0.72 & 0.74 & 0.82 & 0.85 \\
MATH-500 & \LS{}, 4 checkpoints & 0.88 & 0.88 & 0.84 & 0.84 & 0.94 & 0.94 & 0.82 & 0.82 \\
\midrule
MMLU-Pro & static budget & 0.60 & 0.60 & 0.59 & 0.58 & 0.66 & 0.65 & 0.64 & 0.63 \\
MMLU-Pro & \LS{}, 14 checkpoints & 0.65 & 0.65 & 0.65 & 0.65 & 0.73 & 0.74 & 0.69 & 0.70 \\
MMLU-Pro & \LS{}, 4 checkpoints & 0.67 & 0.66 & 0.66 & 0.65 & 0.72 & 0.71 & 0.75 & 0.75 \\
\bottomrule
\end{tabular}\end{adjustbox}
\end{table}

\section{Hidden-state rule: layer selection and ablations}
\label{app:hidden}

\textsc{HiddenStop} reads the residual stream at the last token of each
checkpoint after layers 18, 24, 30 and 36 of the 36 layers of Qwen3-8B, extracted with one forward pass over each
logged trace and stored in half precision. On the training pool the states are standardized and reduced by whitened
principal component analysis to 64 components, and a logistic regression with the regularization of \LS{} is trained
on the safe-to-stop label. The layer is chosen by five-fold cross-validated AUC on the training pool, with the
checkpoints of a question kept in one fold; the evaluation set is not used. Table~\ref{tab:hidden_variants} lists the
AUC of each layer and the saving of each variant, each certified as a policy of its own. All variants save
\hidVarMinMath--\hidVarMaxMath\% on MATH-500, less than \LS, and \hidVarMinMmlu--\hidVarMaxMmlu\% on MMLU-Pro, less
than the static budget.

\begin{table}[!htbp]
\centering
\caption{Variants of \textsc{HiddenStop} at $\alpha=0.05$, $p=0$ (Qwen3-8B, 50 splits): cross-validated AUC of the
layer on the training pool, saving, realised lost-correct risk and share of splits with realised risk above
$\alpha$.}
\label{tab:hidden_variants}
\smallskip
\footnotesize\setlength{\tabcolsep}{3.6pt}
\begin{adjustbox}{max width=\textwidth}\begin{tabular}{llrrrr}
\toprule
Task & Variant & CV AUC & saving (\%) & risk (\%) & splits above $\alpha$ (\%) \\
\midrule
MATH-500 & selected layer &  & 14.1 & 2.2 & 2 \\
MATH-500 & layer 18 & 0.69 & 14.1 & 2.2 & 2 \\
MATH-500 & layer 24 & 0.68 & 13.9 & 2.3 & 10 \\
MATH-500 & layer 30 & 0.66 & 15.9 & 2.5 & 2 \\
MATH-500 & layer 36 & 0.65 & 16.2 & 2.7 & 0 \\
MATH-500 & with budget features &  & 16.6 & 2.8 & 14 \\
MATH-500 & out-of-fold thresholds &  & 12.6 & 2.3 & 2 \\
\midrule
MMLU-Pro & selected layer &  & 18.9 & 3.3 & 4 \\
MMLU-Pro & layer 18 & 0.73 & 18.9 & 3.3 & 4 \\
MMLU-Pro & layer 24 & 0.73 & 21.9 & 3.6 & 6 \\
MMLU-Pro & layer 30 & 0.71 & 18.0 & 3.2 & 0 \\
MMLU-Pro & layer 36 & 0.71 & 19.0 & 3.4 & 6 \\
MMLU-Pro & with budget features &  & 22.0 & 3.3 & 8 \\
MMLU-Pro & out-of-fold thresholds &  & 19.0 & 3.3 & 8 \\
\bottomrule
\end{tabular}\end{adjustbox}
\end{table}

\section{Reproducibility}
\label{app:repro}

Every table and figure of the offline analysis is generated from the logged probe records by the analysis code at \url{https://github.com/dongzhe1/llm-think-stop},
whose README lists the commands; the repository holds the code and the analysis outputs, and the probe records and
the question files are available from the authors on request. GPQA question texts are not redistributed, at the
request of the GPQA authors; the code rebuilds them from the gated dataset. For every question and setting the records hold the thinking
length, the free response, and, at each of the 14 checkpoints and at the end of the trace, the
probe continuation (line-stop and full), its token log-probabilities and top-20 entropies, its decoded lengths and the
backtracking-marker density of the prefix. The reasoning traces themselves and their token sequences are not included in the released records.
The per-run summaries of the online measurement are released with the analysis outputs; its request logs are
available with the probe records. The hidden states used by \textsc{HiddenStop} (about 0.7 GB) are not; the released extraction
script regenerates them from the reasoning traces, which are available from the authors on request.
The fine-grid probes of the static budget (answer text and decoded lengths, without log-probabilities) are included
for the evaluation splits. Any checkpoint schedule that is a subset of the 14 budgets, any of the three probe accountings, a different grader, and any
stopping policy computable from these records can therefore be evaluated on a CPU; policies that need the trace text or other checkpoints require new generation.

\bibliographystyle{plainnat}
\bibliography{references}

\end{document}